\documentclass[10pt,journal,compsoc]{IEEEtran}                                                     
\IEEEoverridecommandlockouts                              
\usepackage[utf8]{inputenc}
\usepackage[T1]{fontenc}                                            
\usepackage{graphicx} 
\usepackage{subfigure}
\usepackage{url}
\usepackage{float}
\usepackage{mathrsfs}
\usepackage{multirow}
\usepackage{microtype}
\usepackage{cite}
\usepackage{amssymb}
\usepackage{xcolor}
\usepackage{adjustbox}
\usepackage{savesym}
\usepackage[ruled,vlined]{algorithm2e}
\usepackage{pifont} 
\usepackage{algpseudocode}
\alglanguage{pseudocode}
\usepackage{mathrsfs}

\SetKwInput{KwInput}{Input}                
\SetKwInput{KwOutput}{Output}              
\usepackage{fvextra} 
\DefineVerbatimEnvironment{Highlighting}{Verbatim}{breaklines,commandchars=\\\{\}}

\usepackage{soul,color}
\definecolor{R}{RGB}{0,0,150}

\newtheorem{researchq}{RQ}
\usepackage[framemethod=TikZ]{mdframed}
\usepackage{enumitem}

\usepackage{paralist}
\usepackage[medium,compact]{titlesec}
\titlespacing*{\subsection}{0pt}{*0.95}{*0.95}
\titlespacing*{\subsubsection}{0pt}{*0.95}{*0.95}
\begin{document}

\title{\LARGE \bf
Evaluation of Federated Learning in Phishing Email Detection
}

\author{Chandra Thapa,
        Jun Wen Tang,
        Alsharif Abuadbba,
        Yansong Gao,
        Seyit Camtepe,
        Surya Nepal,
        Mahathir Almashor,
        and~Yifeng Zheng
\IEEEcompsocitemizethanks{\IEEEcompsocthanksitem C.~Thapa, and S.~Camtepe are with Data61, CSIRO, Sydney, Australia. E-mail: \{chandra.thapa; seyit.camtepe\}@data61.csiro.au.}
\IEEEcompsocitemizethanks{\IEEEcompsocthanksitem A.~Abuadbba, S.~Nepal, and M.~Almashor are with Data61, CSIRO, Sydney, Australia, and Cyber Security Cooperative Research Centre, Australia. E-mail: \{sharif.abuadbba; surya.nepal; mahathir.almashor\}@data61.csiro.au.}
\IEEEcompsocitemizethanks{\IEEEcompsocthanksitem J.W.~Tang, Y.~Gao, and Y.~Zheng were with Data61, CSIRO, Sydney, Australia during this work. E-mail: jun.tang@student.unsw.edu.au, yansong.gao@njust.edu.cn, and yifeng.zheng@hit.edu.cn}
}





  
  


\markboth{}%
{Shell \MakeLowercase{\textit{et al.}}: Evaluation of Federated Learning in Phishing Email Detection}

\IEEEtitleabstractindextext{%
\begin{abstract}

The use of Artificial Intelligence (AI) to detect phishing emails is primarily dependent on large-scale centralized datasets, which opens it up to a myriad of privacy, trust, and legal issues. 
Moreover, organizations are loathed to share emails, given the risk of leakage of commercially sensitive information. Consequently, it is uncommon to obtain sufficient emails to train a global AI model efficiently. Accordingly, privacy-preserving distributed and collaborative machine learning, particularly Federated Learning (FL), is a desideratum.
%
%
Already prevalent in the healthcare sector, questions remain regarding the effectiveness and efficacy of FL-based phishing detection within the context of multi-organization collaborations. To the best of our knowledge, the work herein is the first to investigate the use of FL in email anti-phishing.
This paper builds upon a deep neural network model, particularly Recurrent Convolutional Neural Network (RNN) and Bidirectional Encoder Representations from Transformers (BERT) for phishing email detection. It analyzes the FL-entangled learning performance under various settings, including (i) balanced and asymmetrical data distribution among organizations and (ii) scalability.
%
%
Our results corroborate comparable performance statistics of FL in phishing email detection to centralized learning for balanced datasets, and low organization counts.
%
Moreover, we observe a variation in performance when increasing organizational counts. For a fixed total email dataset, the global RNN based model suffers by a 1.8\% accuracy drop when increasing organizational counts from 2 to 10. In contrast, BERT accuracy rises by 0.6\% when going from 2 to 5 organizations.
However, if we allow increasing the overall email dataset with the
introduction of new organizations in the FL framework, the organizational level performance is improved by achieving a faster convergence speed. Besides, FL suffers in its overall global model performance due to highly unstable outputs if the email dataset distribution is highly asymmetric.

\end{abstract}

}
\maketitle

\IEEEdisplaynontitleabstractindextext

\IEEEpeerreviewmaketitle
\ifCLASSOPTIONcompsoc
\IEEEraisesectionheading{\section{Introduction}\label{sec:intro}}
\else
\section{Introduction}
\label{sec:introduction}
\fi

\IEEEPARstart{E}{mail} is the most usual means of formal communication. At the same time, it is exploited as a common attack vector for phishing attacks, where attackers disguise as a trustworthy entity and try to install malware or obtain sensitive information such as login credentials and bank details of the recipient. Based on the 2019 phishing and email fraud statistics~\cite{phishingreport1}, phishing accounts for 90\% of data breaches, which leads to an average financial loss of \$3.86M. Moreover, phishing attacks cost American businesses half a billion dollars a year \cite{mathews2017phishing}, and it is increasing. Recently, COVID-19 drives phishing emails up to an unprecedented level by over 600\%~\cite{covidphishing}.


To protect users from phishing attacks, various techniques have been devised. These techniques can generally be divided into two categories: traditional methods and artificial intelligence (AI) based methods. Traditional methods rely on known email formats, and they are inefficient because the email formats can be easily manipulated with time by the attackers. In comparison, the AI-based method is context-aware. It can continuously learn from the newly available emails and adapt to handle the new attack formats/cases efficiently on time. 


Among AI-based methods, deep learning (DL) feeds the email data directly to the system without requiring delicate feature engineering. Moreover, feature engineering is usually a time-consuming and laborious domain-specific task necessary for the conventional ML-based methods such as decision tree~\cite{review1}. This makes DL a suitable method to learn against evolving threats with time. Convolutional Neural Network~\cite{CNNemail}, Recurrent Convolutional Neural Network (RCNN)~\cite{fang2019phishing}, and transformers~\cite{bert} are typical examples of the DL-based method. Although DL-based methods are preferable over other methods considering its performance and automated feature engineering, it requires a considerable amount of email data as a trade-off.

Unfortunately, emails are sensitive to organizations\footnote{Organizations are commonly referred to as clients in the remainder of this paper.}, and disclosure to third parties is often avoided~\cite{phishingsurvey}.
Even anonymization of the email is problematic because it can be easily circumvented --- attackers can exploit various characteristics, e.g., social graphs, to re-identify the victim's entity~\cite{gao2018resisting}. As such, it is non-trivial to aggregate emails for centralized analysis. Besides, a recent work~\cite{ho2019detecting} emphasizes the strict ethical concerns when accessing and analyzing the emails of 92 organizations, even with access permission. 
%
%
%
For any purpose, improperly centralized data management could violate specific rules, such as reusing the data indiscriminately and risk-agnostic data processing~\cite{gdprviolations}, required by General Data Protection Regulation (GDPR) \cite{gdpr} and HIPAA \cite{hipaa2003}. Therefore, even with the users' permission to use their data for agreed tasks (e.g., DL), handling the email data under a centralized cloud is still risky under the set of privacy regulations. 
%
%
Thus there is an urgent need for methods that preserve data privacy in DL and break the email data silos. As such, DL can access abundant email datasets and improve its performance (e.g., detection accuracy). 


In this regard, federated learning (FL)~\cite{fedlearningMcMahan17,fed2,kairouz2019advances}, the most popular collaborative learning, is a suitable candidate method. It trains a joint DL model by harvesting the rich distributed data held by each client in a default privacy mode. The privacy of the raw data is enabled by two means; firstly, data never shared with other clients/participants. Secondly, data is always within the control of data custodians (i.e., clients). Consequently, the data custodians have an assurance of some level of privacy and control to their data, motivating them to participate in distributed machine learning for overall social good (e.g., an anti-phishing DL model with high detection). 


FL has been explored in various applications such as finance~\cite{yang2019ffd}, health~,\cite{rieke2020future,kairouz2019advances} and natural language processing (NLP)~\cite{fednlp}. However, it is still unclear how efficient and effective it would be in phishing email detection with regard to the relevant deep models such as (i) THEMIS~\cite{fang2019phishing}, which is the best performing RCNN based centralized model on phishing emails, and (ii) BERT~\cite{bert}, which is the best performing transformer-based centralized model on text data. 
FL over email data is similar to FL in NLP, but the challenge here is that the phishing emails are highly subjective, e.g., spear phishing, and its dataset size is smaller than the NLP corpus.
To the best of our knowledge, the applicability of FL for detecting phishing emails has not been explicitly investigated. Thus in this paper, we take the first studies of FL entangled phishing email detection and investigate its performance using the DL models. 

\subsection{Our contributions}
 This work considers both the balanced and asymmetrical data distribution among clients. The evaluations are carried out with the following six research questions in mind:
\begin{compactitem} 
    \item [\textbf{RQ1}] \textbf{(Balanced data distribution) Can FL be applied to learn from distributed email repositories to achieve comparable performance to centralized learning (CL)?}
    We develop deep-learning anti-phishing models based on FL and CL under balanced data distribution. Their performances show that FL achieves a comparable performance to CL. For example, (i) THEMIS model's performance at 45 epoch is 99.3\% test accuracy, and 97.9\% global test accuracy for the CL and FL with 5 clients, respectively, and (ii) BERT model's performance at 15 epoch is 96.2\% test accuracy and 96.1\% global test accuracy for the CL and FL with 5 clients, respectively. Details are provided in Section~\ref{sec:distributed learning}. 
    
    \item [\textbf{RQ2}]\textbf{(Scalability) How would the number of clients affect FL performance and convergence?}
    Our experiments under balanced data distribution suggest that while keeping the same total email dataset, the convergence of the accuracy curve and its maximum value is model dependent.
    We observe THEMIS model dropping by around 0.5\% in global test accuracy at 45 epoch going from 2 to 5 clients, but the BERT model improved by around 0.6\% in global test accuracy at 15 epoch going from 2 to 5 clients. 
    \textbf{RQ4} addresses some FL's benefits to the THEMIS model. Details are provided in Section~\ref{sec:distributed learning}.   
    
    \item [\textbf{RQ3}]\textbf{(Communication overhead) What is the communication overhead resulting from FL?}
    FL has a communication overhead as a trade-off to privacy, and it is only dependent on the model size. For example, we quantify the overhead per global epoch per client for THEMIS around 0.192 GB and BERT around 0.438 GB for all cases under our setting. We regard such overheads as not of particular concern for organizational level participants. Details are provided in Section~\ref{sec:distributed learning}.

    \item [\textbf{RQ4}]\textbf{(Client-level perspectives in FL) Can a client leverage FL to improve its performance?}
    We investigate client-level performances under both balanced and asymmetric data distribution in FL, considering the cases where clients are available with time in the training process, and the total email dataset increases with the addition of the new client. A fast convergence in the accuracy curve is observed with the THEMIS model. Details are in Section~\ref{sec:transfer learning}.
    
    \item [\textbf{RQ5}]\textbf{(Asymmetric data distribution) How would FL perform under asymmetric data distribution among clients due to the variations in the local dataset's size and the local phishing to legitimate sample ratio?}
   
   Our studies with the THEMIS model under 2, 5, and 10 clients suggest that FL performs well and similar over the asymmetric data distribution that is due to different local dataset sizes and the local phishing to legitimate sample ratio. Thus FL is resilient in these scenarios. Details are in Section~\ref{sec:imbalance data}.

    \item [\textbf{RQ6}]\textbf{(Asymmetric data distribution) How would FL perform under extreme dataset diversity among clients?}
    Data asymmetry, in this case, is due to the class skewness and different dataset size (small to large) across the clients. 
    Our studies suggest that forming a best performing global model for all clients under FL is not straightforward. In addition, the local and global performances are model depended. Details are in Section~\ref{sec:practical_setup}.  
\end{compactitem}

\section{Background}

\subsection{Centralized learning}
Centralized learning (CL) is normally performed by aggregating all available datasets (e.g., phishing and legitimate emails) at one central repository. Then it performs centralized machine learning on the aggregated dataset. 
During the learning process, a modeler can access the raw data shared by one or more clients, thus making it unsuitable if the data is private such as emails. Besides, in the era of big data and deep learning, it is non-trivial to maintain the required resources, including storage and computation, in CL. Thus there is a rise in distributed learning, in particular, FL. 


   
    

\subsection{Federated learning}

\begin{figure}[t]
	\centering
	\includegraphics[trim={0 1cm 0cm 1cm},clip, width=0.8\linewidth]{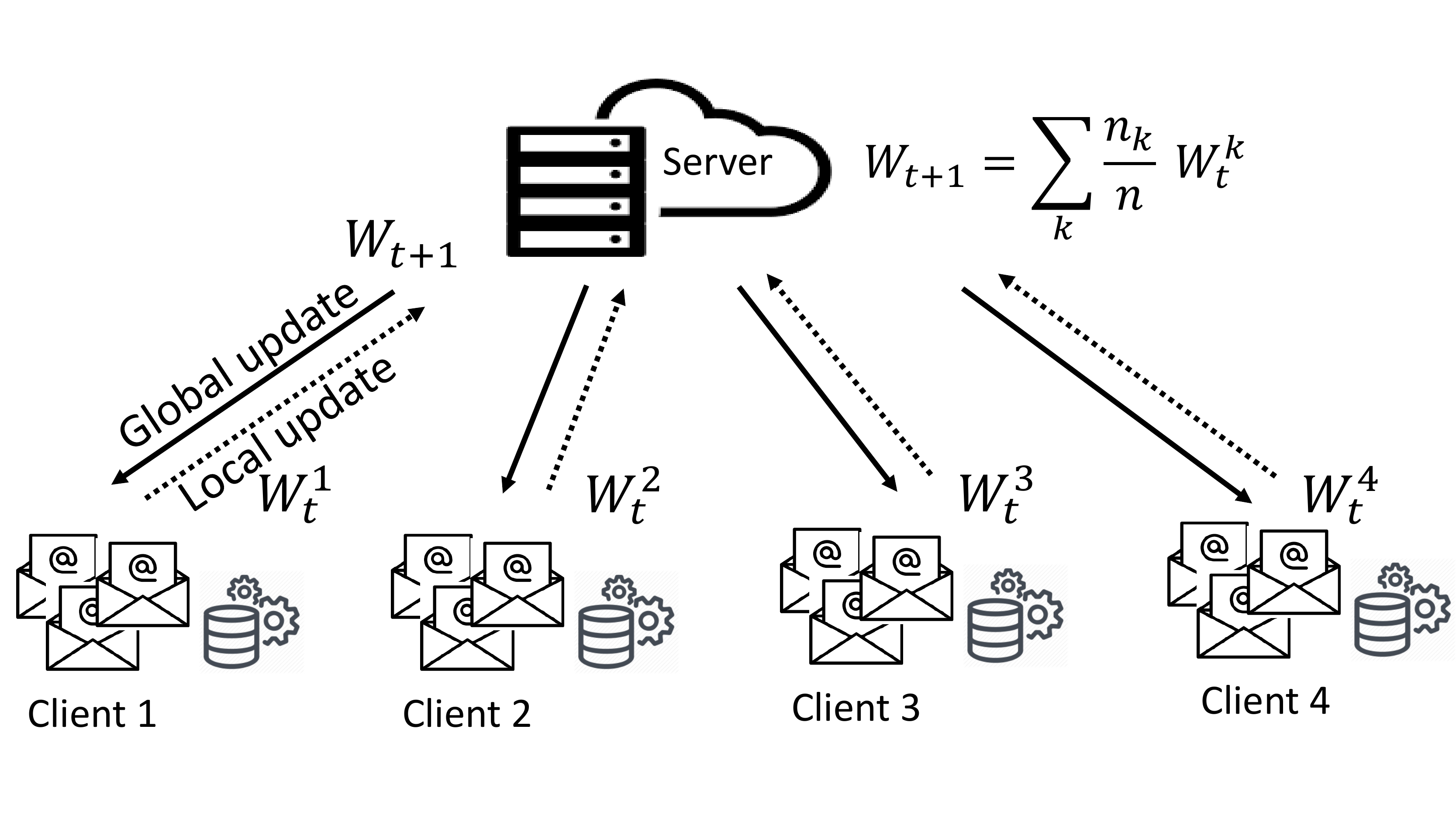}
	\caption{An overview of federated learning.}
	\label{fig:fedlearning}
	\vspace{-0.4cm}
\end{figure}

Federated learning (FL)~\cite{fedlearningMcMahan17} allows parallel deep learning training across distributed clients and pushes the computation to the edge devices (i.e., clients). Fig.~\ref{fig:fedlearning} illustrates an overview of FL. There are four exemplified clients with their local email datasets and one coordinating server. 
Firstly, each client $k$, $k\in \{1,2,3, 4\}$, trains the model on their local email dataset $D_k$ and produce the local model $W^k_t$ at time instance $t$. Secondly, all clients upload their local models to the server. Then the server performs the weighted averaging (i.e., aggregation) of the local models and updates the global model $W_{t+1}$. Finally, the global model is broadcast to all clients (model synchronization), and this completes the one round, known as one global epoch, of the FL process. This process continues until the model converges (see Algorithm~\ref{algo:fedtraining}). In FL, the server synchronizes the training process across the clients. Over the entire training process, only the models (i.e., model parameters) are transmitted between the clients and the server. Thus a client (e.g., financial institution) does not require to share their raw email data to the server (e.g., coordinated by an email analyzer) or any other clients during the training process. Thus the data are always local and kept confidential in FL.
\section{Experimental setup} 
\label{sec:experimentalsetup}

\subsection{Datasets}
In this work, primarily, phishing and legitimate email samples are collected from three popular sources, namely First Security and Privacy Analytics Anti-Phishing Shared Task (IWSPA-AP)~\cite{IWSPA-AP}, Nazario’s phishing corpora (Nazario)~\cite{nazario}, and Enron Email Dataset (Enron)~\cite{enron}. Besides, we consider phishing emails from CSIRO\footnote{https://www.csiro.au/} (private emails) and Phishbowl~\cite{phishbowl}. 
The dataset contains email samples with both header\footnote{Email header precedes the email body and contains information of the header fields, including \emph{To, Subject, Received, Content-Type, Return-Path,} and \emph{Authentication-Results.}} and without header: IWSPA-AP has both types, whereas all email samples in Nazario and Enron have the header accompanied by the body, no header for CSIRO and Phishbowl emails. Overall, the data sources include Wikileaks archives, SpamAssassin, IT departments of different universities, synthetic emails created by Data engine~\cite{dada}, Enron (emails generated by employees of Enron corporation), Nazario (personal collection), and private emails (CSIRO). 
CSIRO emails are phishing emails reported by CSIRO staff between 2017 to 2020, and we manually labeled them to remove the spam. 
Phishbowl emails are published by Cornell University, and we collected emails reported from April 2019 to January 2021. The emails on the website have a header but with partial fields or body only, so we consider the body only of these emails for our dataset.
To provide more insight into the email samples, we present some frequently appearing words in them as follows: 
\begin{compactitem} 
    \item IWSPA-AP phishing email (a) with header includes \emph{account, PayPal, please, eBay, link, security, update,\linebreak bank, online,} and \emph{information}, and (b) without header includes \emph{text, account, email, please, information, click, team, online,} and \emph{security}. IWSPA-AP legitimate email (a) with header includes \emph{email, please, new, sent, party, people, Donald, state,} and \emph{president}, and (b) without header includes \emph{text, link, national, US, Trump,} and \emph{democratic}.
    
    \item Nazario includes \emph{important, account, update, please, email, security, PayPal, eBay, bank, access, information, item, click, confirm,} and \emph{service.} 
    
    \item Enron includes \emph{text, plain, subject, please, email, power, image, time, know, this, message, information,} and \emph{energy.}
    
    \item CSIRO includes \emph{shopping, parcel, invitation, payment, employee, webinar, survey, newsletter, program} and \emph{workshop}.
    
    \item Phishbowl includes \emph{account, id, password, Cornell, upgrade, notice, administrator, message, job, server,} and \emph{verify}.
    
\end{compactitem}

\begin{algorithm}[t]
\footnotesize
\KwInput{Email dataset}
\KwOutput{Model performance (e.g., Accuracy, F1-score, and Precision)}
    \tcc{Server-side}
    \SetKwProg{Fn}{Server:}{}{}
    \Fn{}{
        Initialize and send global model $W_t$ to all $K$ clients\;
        
       \For{\textup{epoch} $e \in E$}{  
        \For{\textup{each client $k\in \{1,2,\dots, K\}$ in parallel}}{
            $W_t^k \leftarrow \textup{ClientUpdate}(W_t^k)$ \tcp*{local updates} 
        }
        Perform weighted averaging and update the global model: $ W_{t+1} \leftarrow \sum_{k= 1}^{K} \frac{n_k}{n} W_t^k, \quad n=\sum_{k} n_{k}$\;
        
        Send the updated global model $W_{t+1}$ to all clients\;
        }
    }
    \vspace{0.3cm}
    \tcc{Client-side at each client $k$}
    \SetKwProg{Fn}{ClientUpdate$(W_t^k)$:}{}{}
    \Fn{}{
        
        \tcc{Runs once at the beginning}
        \SetKwProg{Fn}{Email dataset preparation:}{}{}
          \Fn{}{
            Data extraction, setting up phishing to legit email ratios, 80:20 train-test split, and Tokenization (details in Section~\ref{sec:datasetpreparation})\;
        }
        
        \tcc{Runs repetitively during the training/testing}
        \While{\textup{global model $W_t$ is received from the server}}{
            Set $W_t^k = W_t$\; 
            \tcc{Training/testing on the local email dataset $X$ having $n_k$ samples}
            \For{\textup{batch} $b \in \mathcal{B}$}{
                
                $ W^k_{t} \leftarrow W^k_{t} -\eta \triangledown f(W^k_t;X)\ \textup{for}\ X\sim \mathcal{P}_k$
                \tcp*{$\mathcal{B}$ is batch size, $\eta$ is learning rate, and $\triangledown f(W^k_t;X)$ is gradients with respect to the cost function}
            }
            Send locally trained $W_t^k$ to Server\;
        }

}
 \caption{Federated learning}
 \label{algo:fedtraining}
 \end{algorithm}

We have considered the {\it updated email dataset} from the sources, e.g., Nazario's phishing corpus 2019. In total, the experimental dataset has $23916$ email samples, and Table~\ref{data_info} shows the number of emails extracted from each source.     
\begin{table}[H]
    \centering
    \footnotesize
    \caption{The number of email samples.}
    \begin{tabular}{|c|c|c|c|}
    \hline
    Source & Phishing (P) & Legitimate (L) & P+L\\ \hline
    \multicolumn{1}{|c|}{IWSPA-AP}   & 1132     & 9174 &      10306 \\ \hline
    \multicolumn{1}{|c|}{Nazario} & 8890     & 0     &    8890  \\ \hline
    \multicolumn{1}{|c|}{Enron}   & 0        & 4279   &    4279 \\ \hline
    \multicolumn{1}{|c|}{CSIRO}  &  309 & 0  & 309\\\hline
    \multicolumn{1}{|c|}{Phisbowl}  & 132   & 0 & 132\\\hline
    \multicolumn{1}{|c|}{\emph{Total}}  &10463 & 13453 & 23916\\
    \hline
    \end{tabular}
    \label{data_info}
\end{table}
 
\subsection{Deep learning model selection:}
\label{sec:modelSelection}


The selected models are described in the following:

\subsubsection{THEMIS model}
\label{sec:themis_model}
THEMIS is one of the recent models, which has been demonstrated to be highly effective for phishing email detection. It employs Recurrent Convolutional Neural Network (RCNNs) and models emails at multiple levels, including char-level email header, word-level email header, char-level email body, and word-level email body~\cite{fang2019phishing}. This way, it captures the deep underlying semantics of the phishing emails efficiently and consequently making THEMIS better than existing DL-based methods that are limited to natural language processing and deep learning~\cite{deeplearingemail}. 

\begin{figure}[t]
	\centering
	\includegraphics[scale=0.55]{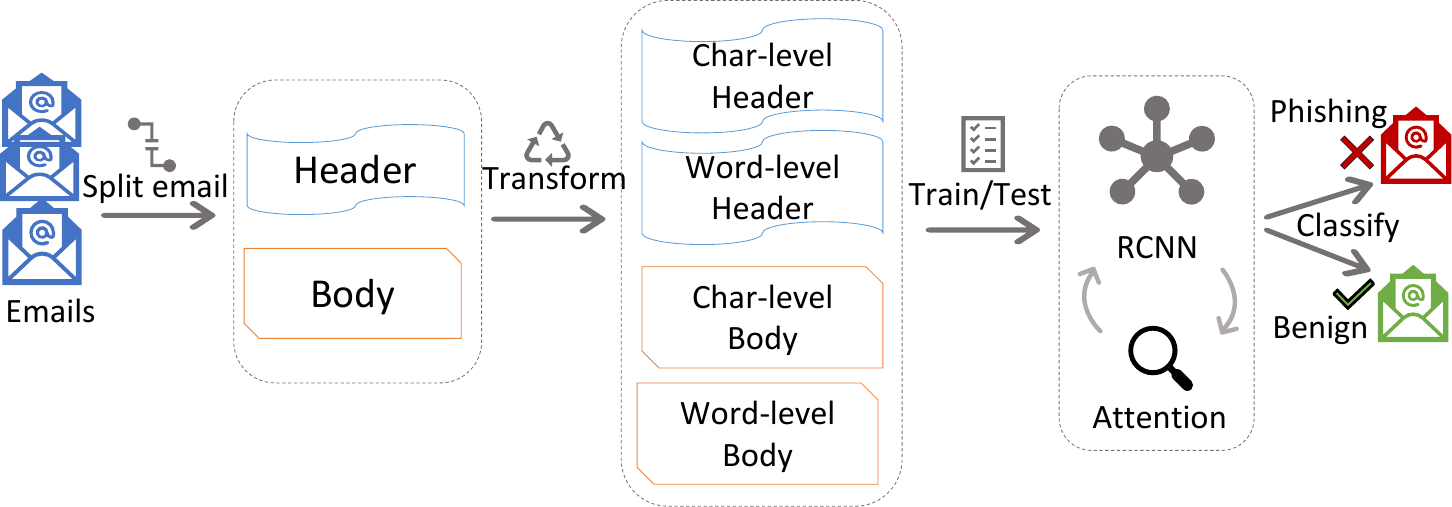}
	\caption{An overview of THEMIS model.}
	\label{fig:themis}
\end{figure}

\textbf{Overview of THEMIS model}: Fig. \ref{fig:themis} illustrates a system overview of the THEMIS model. Firstly, THEMIS extracts the char-level and word-level of the email header and body, and then an embedding layer converts all these levels to the respective vector representation. Afterward, it feeds each vector representation into the RCNN model \cite{lai2015recurrent} and learns a representation for the email header and email body, respectively.    
%
%
%
%
THEMIS RCNN consists of four Bidirectional-Long Short-Term Memory (Bi-LSTM) that obtain the left and right semantic information of a specific location with its embedding information from the above four vectors, thus forming something called a triple. Next, these triples are mapped into specified dimensions using a \texttt{tanh} activation function.  The longitudinal max polling is then applied to obtain four different representations, which will be paired to form only two representations for the header and the body.
As the email header representation and body representation have varying degrees of impact on phishing detection, an attention mechanism is applied to compute a weighted sum of the two representations. This produces an ultimate representation of the whole email, which is further processed to produce the classification result. For more details of the THEMIS model, we refer readers to \cite{fang2019phishing}.

The THEMIS original paper considered only emails with headers (8780 samples) and the THEMIS model was trained on those data~\cite{fang2019phishing}. In contrast, this paper considers (i) around 2.7x more email samples, (ii) emails with header and without header information, (ii) THEMIS under emails with both headers and without headers (body only), and we analyze their performances with CL and FL. 

\subsubsection{Bidirectional Encoder Representations from Transformers}
Bidirectional Encoder Representations from Transformers (BERT)~\cite{bert} is a language model initially developed by Google. Transformer encoders are basic blocks of BERT. Transformers learn the contextual information in the input sequence by an attention mechanism that enables them to relate different parts of the input sequence to find their relationship, for example, the contextual information of the words/sub-words in a sentence. BERT reads the entire input sequence to get bidirectionally trained. This enables BERT to learn the contextual information better than the techniques looking at the sequence from one direction (e.g., each word conditioned on its previous or next words). 

\begin{figure}[t]
	\centering
	\includegraphics[trim={3cm 0.5cm 3cm 2cm},clip, width=0.8\linewidth]{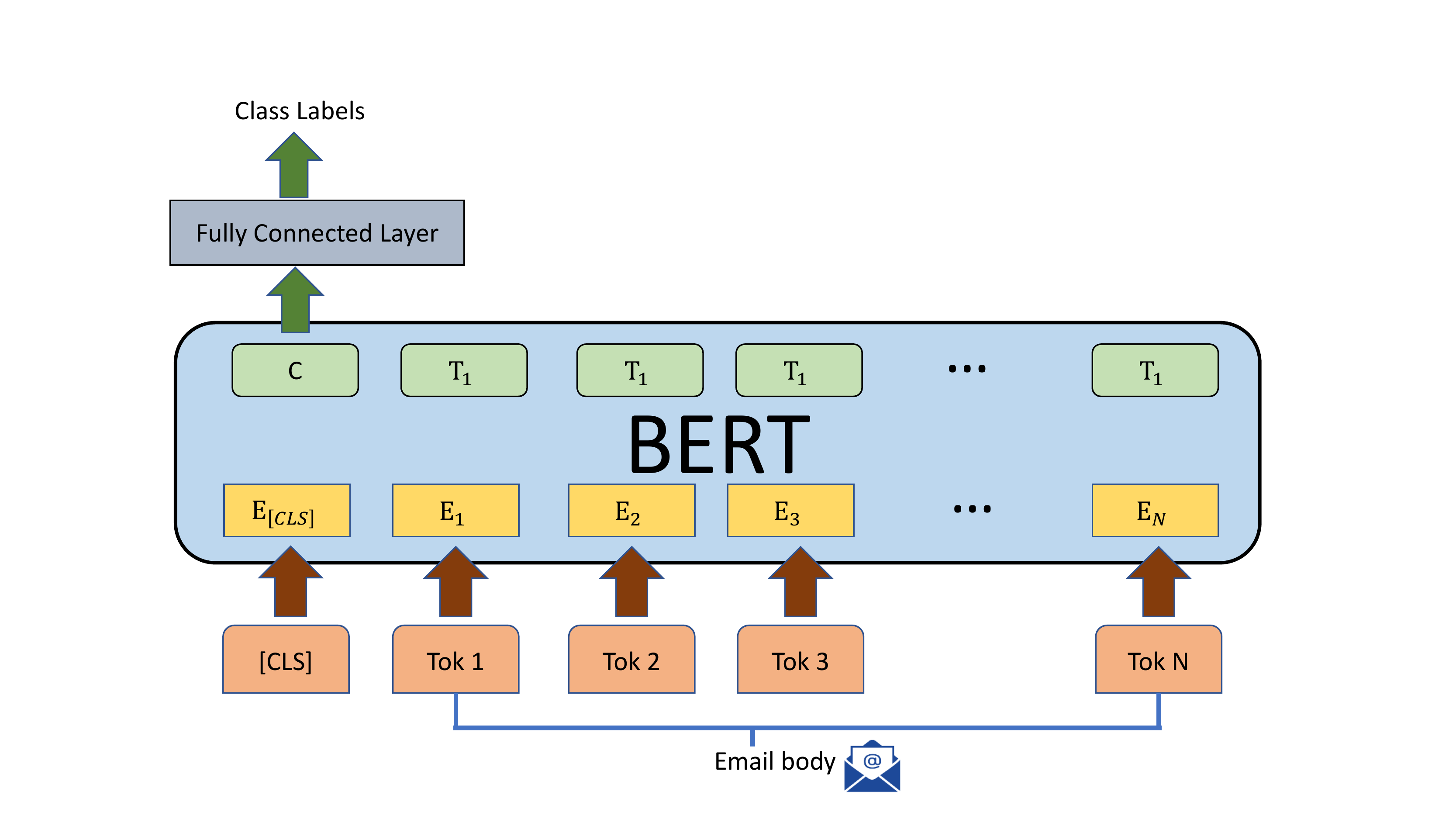}
	\caption{An overview of BERT model for phishing detection, where $\textup{E}$ is input embeddings, $\textup{C}$ and $\textup{T}_i$ are the final hidden vectors of token [CLS] and $i^{\textup{th}}$ token, respectively.}
	\label{fig:bert}
\end{figure}

\textbf{Overview of BERT model}: 
In this paper, we use Huggingface's library called \emph{transformers} to use the bert-base-uncased pretrained on a large corpus of English texts~\cite{bert_base} such as English Wikipedia and BookCorpus. The model has 12 layers of transformers, 768 hidden sizes, 12 self-attention heads, and altogether 110 million parameters. 
BERT model is used for various tasks other than natural language processing by performing its fine-tuning. In this process, few layers are added to the end of the model (e.g., classification layer) and train/test the whole model with a small learning rate over the available dataset. 
The model is pre-trained for two tasks: masked language modeling (MLM) and next sentence prediction (NSP). MLM predicts the masked words in a sentence whose 15\% words are randomly masked by the model at the beginning. NSP predicts whether two sentences, which have their words masked to some percentage, follow each other or not. Thus the embedding has special tokens [CLS] at the beginning of each sentence, [SEP] to separate two sentences in a sequence and the end of the sentence, and [MASK] to mask any word in the sentence. An overview of the BERT model for a classification task is depicted in Fig.~\ref{fig:bert}.

BERT has been used in phishing email detection~\cite{catbert}. The authors designed a smaller BERT, called CatBERT, by pruning odd-numbered transformers from it and replacing those with adapters. CatBERT is reported to achieve 87\% of detection rate on their own data collected at Sophos, and it is best compared to the DistilBERT (compressed BERT model)~\cite{distilbert} and LSTM based models on their dataset.  
Our focus in this paper is to demonstrate the feasibility of FL on the detection side. We proceed by exploring the standard BERT rather than distilled BERT models (which only benefit computation). To our best knowledge, only CL has been observed and analyzed for the BERT in phishing email detection. 
Moreover, in centralized learning, the performance of the standard BERT in phishing email detection is still not clear.

In our experiments, BERT considers only the body of email samples as its input because, unlike THEMIS, it has no dedicated architectural part to get all header fields separately. However, we can concatenate the header information to the body and feed it to the BERT model. 

\subsection{Data preparation}
\label{sec:datasetpreparation}

We have different data sources and data distribution among the clients in our FL setups. For \textbf{RQ1} to \textbf{RQ5}, we consider three email sources, namely IWSPA-AP, Nazario, and Enron, with a total of 23,475 email samples. We consider the other two sources, CSIRO and Phishbowl emails, for \textbf{RQ6}. We did this division as there are only 441 phishing email samples from CSIRO and Phishbowl emails, and they should be analyzed only in the extreme dataset diversity under \textbf{RQ6}.

Under our balanced dataset setup for \textbf{RQ1} to \textbf{RQ4}, including the asymmetric dataset for \textbf{RQ4}, we consider equal phishing and legitimate email samples out of 23,475 data samples. So, we prepare the experimental dataset of size 20,044 (i.e., $2 \times 10,022$) --- to be precise, 10,022 is aligned with the number of phishing emails while the number of legitimate emails is 13,453. Moreover, the new dataset has four parts - phishing header, phishing body, legitimate header, and legitimate body - each part with 10,022 samples.  
The experimental dataset is equally and uniformly distributed in all our distributed setups with multiple clients except for the cases with the asymmetric dataset (\textbf{RQ4}, \textbf{RQ5}, and \textbf{RQ6}). For example, cases with five clients have a dataset of size 4008 (i.e., around 20044 divided by 5) in each client. 
Under the distributed setup for \textbf{RQ5}, we perform two experiments (i) each client has different sizes of email samples with 50:50 phishing to legitimate email ratio, and (ii) each client has the same sizes of email samples but with different phishing to legitimate email ratio. 
Under \textbf{RQ6}, we consider five clients, each uniquely corresponding to one of our five data sources.


We perform all the above data distributions for experimental setups, which simulate cases with multiple organizations having their own local data (distributed data) in different geo-locations and remaining in silos. 
For all experiments, the training-to-testing data split ratio is 80:20. 

Our email data sources have two types of file formats, viz., text file (.txt) and mbox file (.mbox). Each email is a single text file if the email sample is in text format. In the mbox format, all messages are concatenated and stored as plain text in a single file. Moreover, each message starts with the four characters "From" followed by a space. Both types of email files are firstly parsed into two parts, namely email header and email body, and then subjected to further processing, including cleaning and tokenization.


\subsubsection{Extraction of Header and Body} The class \texttt{Header} of the python module, called \texttt{email.header}~\cite{emailheader}, is used to extract the email header, and this separates the header and body part of the email samples. In the header section, we consider only the \emph{Subject} and the \emph{Content-Type} field, which are deemed essential for phishing detection. This separation is done by using the python library called the regular expression (RE) module~\cite{re}. 

\subsubsection{Cleaning of the Extracted Header and Body} 
The python library \texttt{Beautiful Soup 4}~\cite{soup} and \texttt{HTML parser}~\cite{htmlparser} are used to clean the text information in HTML format. Besides, we use RE for the plain text (both in header and body) cleaning by removing punctuation and non-alphabetic characters. To filter out the stop words from the header and body, we use \texttt{stopwords} of the nltk packages (\texttt{nltk.corpus})~\cite{nltk} of python.

\subsubsection{Tokenization} 

Our two models under investigation require different tokenization methods. 

For THEMIS, we performed tokenization in the following way:
To get the char-level and word-level sequences of the tokens for both header and body parts, the Tokenizer class provided by \texttt{Keras} library~\cite{tokenizer} is used.
Basically, this is to encode each character/word as a unique integer as required by the input format of the embedding layer.
Two main functions are used for tokenization; these are ‘fit\_on\_texts,’ which updates internal vocabulary based on a list of texts, and ‘texts\_to\_sequences,’ which transforms each text in texts to a sequence of integers by considering only words known by the tokenizer. In all our measurements, we keep 50, 100, 150, and 300 as the length of the four sequences of tokens, which are word-level header, char-level header, word-level body, and char-level body, respectively.

For BERT, we consider only the email body and performed its tokenization in the following way: We used BertTokenizer~\cite{berttokenizer} provided by Huggingface library to encode the email's body to tokens. Besides, the tokenizer inserts additional special tokens such as [CLS] and [SEP] in the process. BERT allows only 512 tokens to be inputted at a time, and it is considered during tokenization. Also, the tokenizer returns original input ids, attention masks, and token type ids required during learning.

\subsection{Experimental steps}
For performance analysis, we use a High-performance Computing (HPC) platform that is built on \emph{Dell EMC's PowerEdge} platform. It has the \emph{Tesla P100-SXM2-16GB} GPU model. All code is written in Python 3.6.1. The THEMIS model, which has an RCNN, is implemented by using TensorFlow 2.2.5~\cite{tensorflow} and Keras 2.2.5~\cite{keras} framework, and the BERT model, a pretrained transformer model, is downloaded from the Huggingface library. In all measurements, we keep the same random seed, i.e., \texttt{random.seed(123)}. We run centralized model training and federated model training under various settings in our experiments, but with the same (i) learning rate of 0.0001 and batch size of 256 for THEMIS model, and (ii) learning rate of 0.00001 and batch size of 4 for BERT model. The specific batch size is chosen based on the available resources (e.g., GPU has 16GB internal memory). 
%


\section{Results}
To ease the presentation, we divide this section into four parts under which the six research questions are analyzed with the empirical results. We perform experiments with the THEMIS model in this work, considering both with and without email header information. For convenience, we refer to \emph{THEMIS} model if it considers both email's header and body information, and \emph{THEMISb} model if it considers only email's body information in the remainder of this paper.

\subsection{Distributed email learning under balanced data distribution}~\label{sec:distributed learning}

\begin{figure}[t!]
	\centering
		\subfigure[]{
			\includegraphics[width=0.65\columnwidth, trim=2 2 2 2,clip]{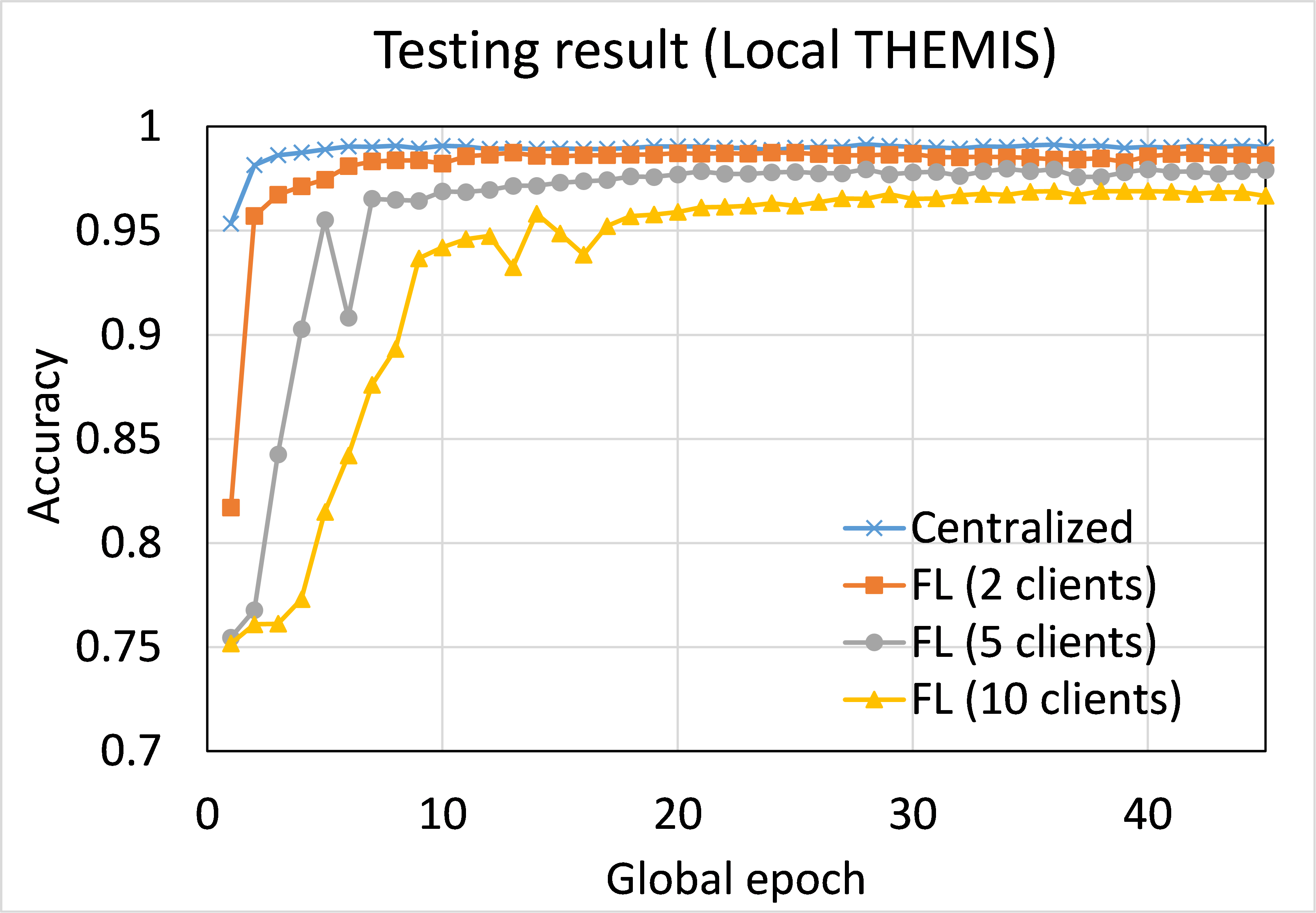}
			\label{fig:themis1}
		}
	\subfigure[]{
			\includegraphics[width=0.65\columnwidth, trim=2 2 2 2,clip]{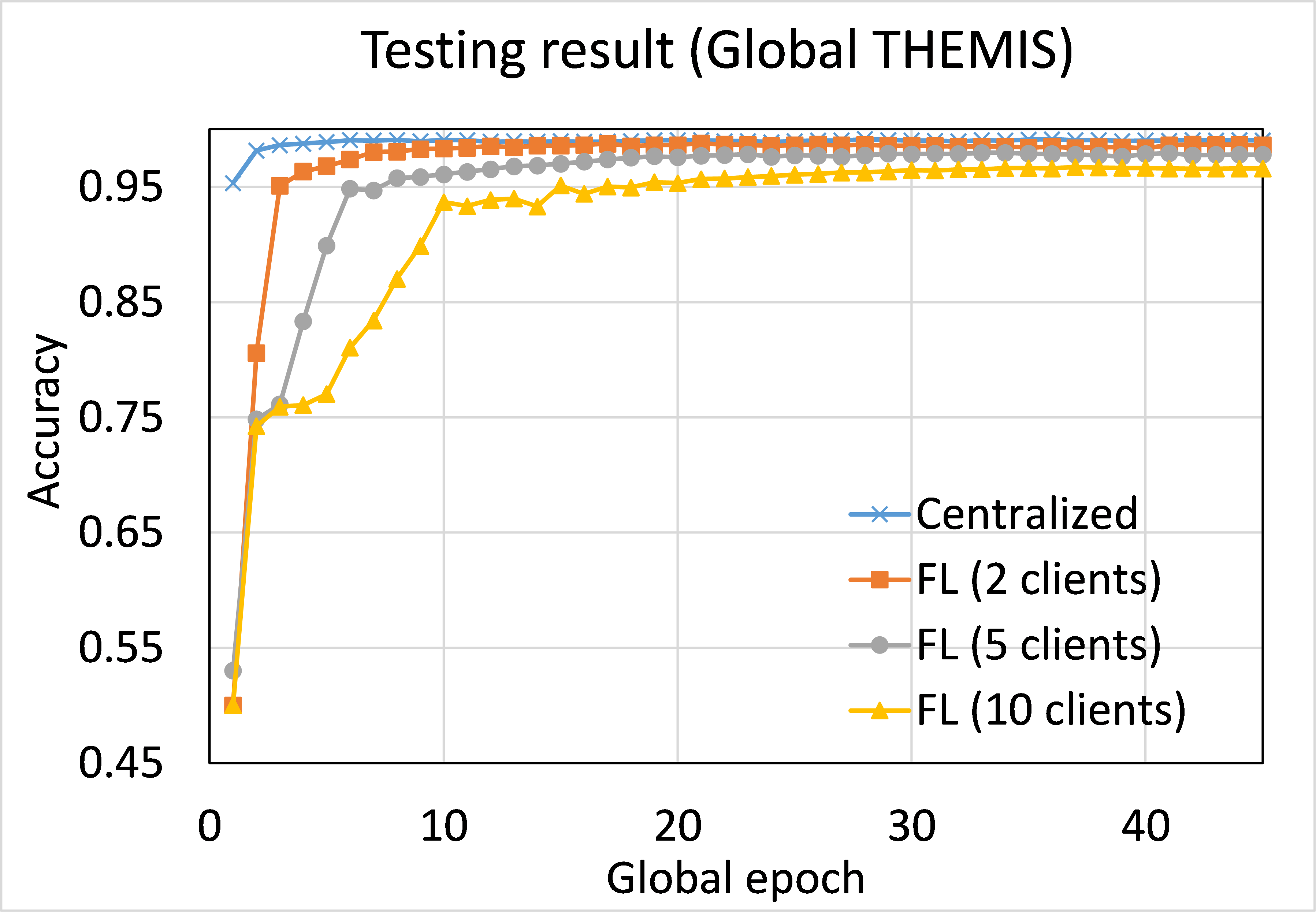}
			\label{fig:themis2}
		}
		\subfigure[]{
		    \includegraphics[width=0.9\columnwidth, trim=2 1 1 1,clip]{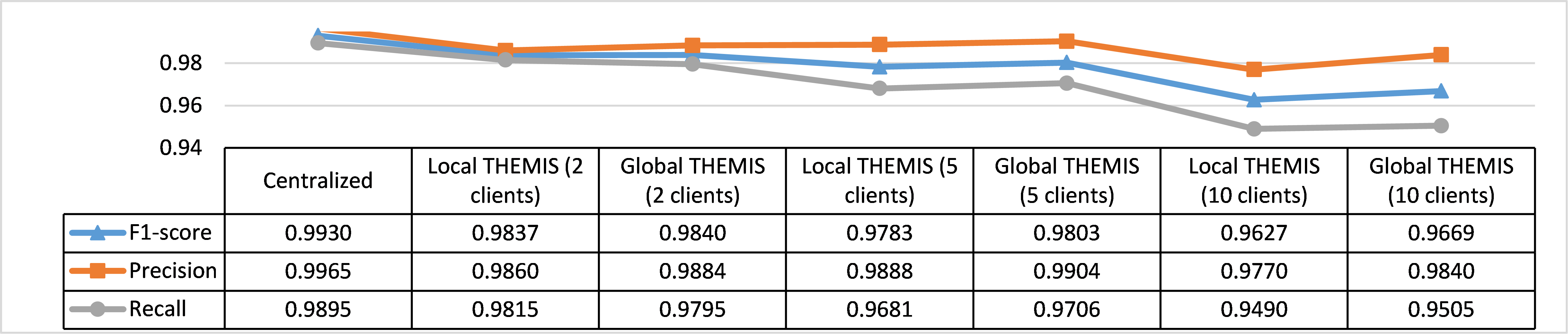}
		    \label{fig:themis3}
		}
		\subfigure[]{
			\includegraphics[width=0.9\columnwidth, trim=2 2 2 2,clip]{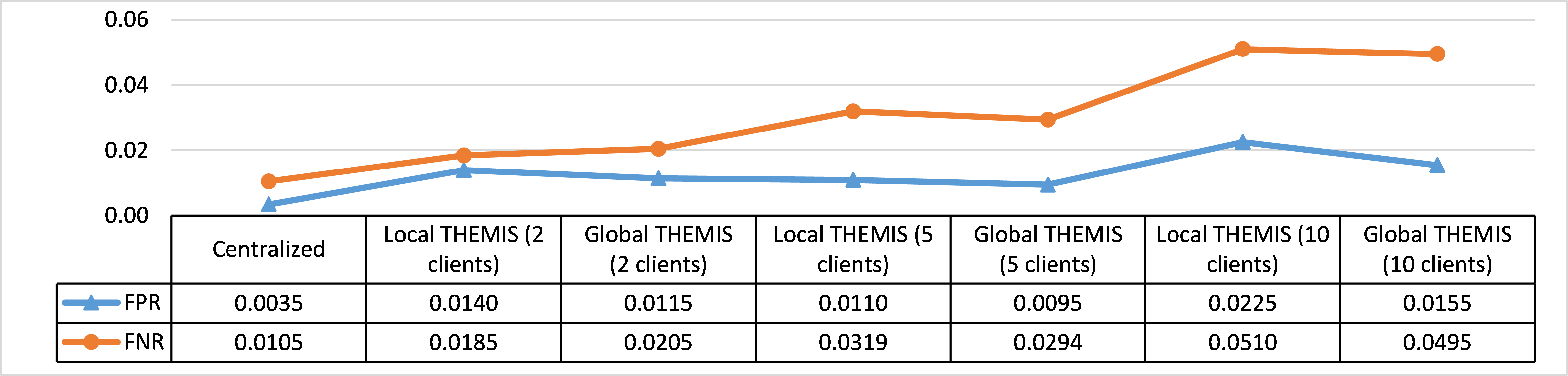}
			\label{fig:themis4}
		}
		\caption{Results for the THEMIS model: Convergence curves of average testing accuracy for the (a) local models and (b) global model. Performance metrics of the testing results at the 45 global epoch in centralized and federated learning (FL) with two, five, and ten clients are depicted in (c) and (d). 
		}
		\label{fig:themis_exp1}
\end{figure}

\begin{figure}[H]
	\centering
		\subfigure[]{
			\includegraphics[width=0.7\columnwidth, trim=2 2 2 2,clip]{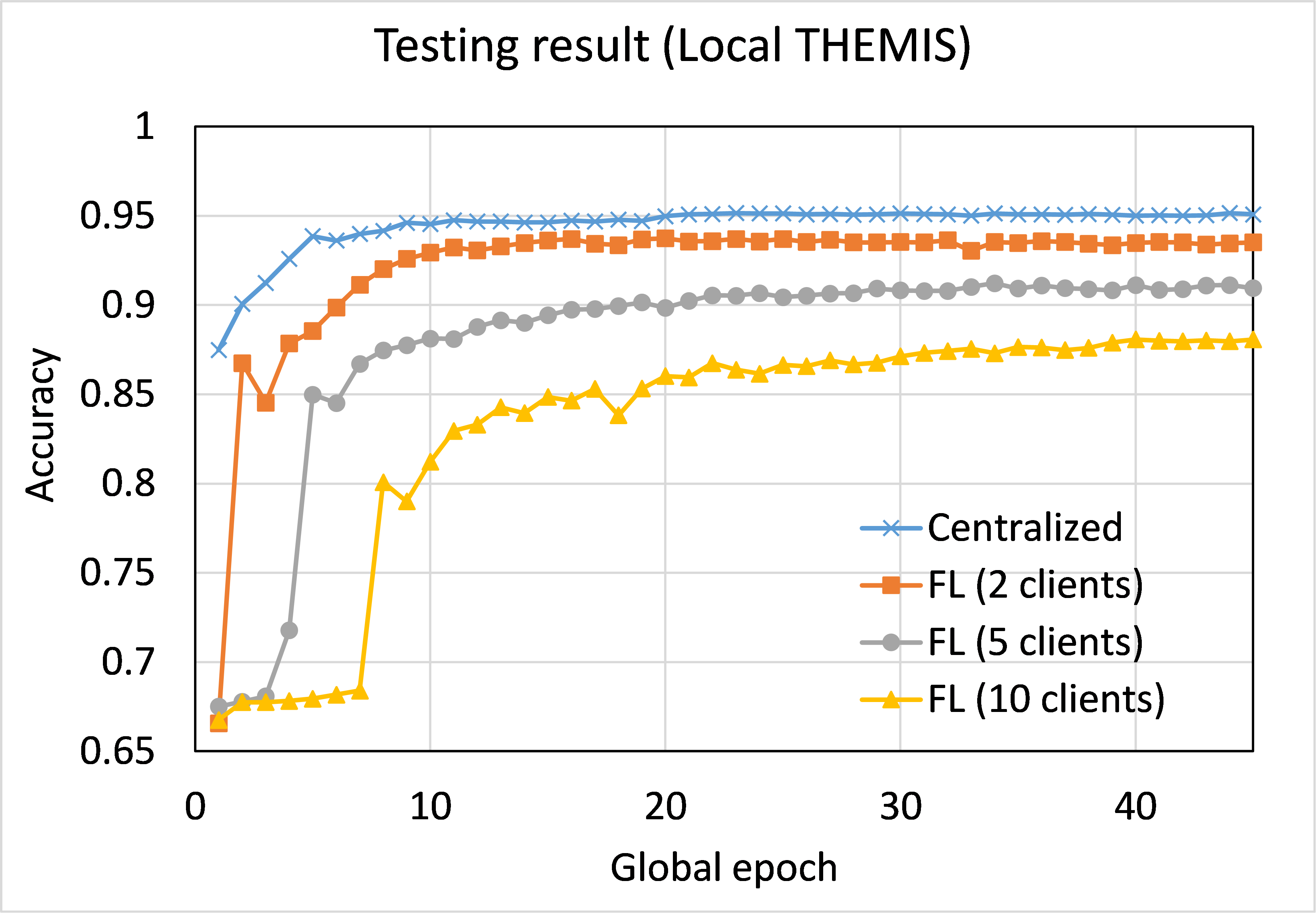}
			\label{fig:b_themis1}
		}
		
    \vskip5pt
 	\subfigure[]{
			\includegraphics[width=0.7\columnwidth, trim=2 2 2 2,clip]{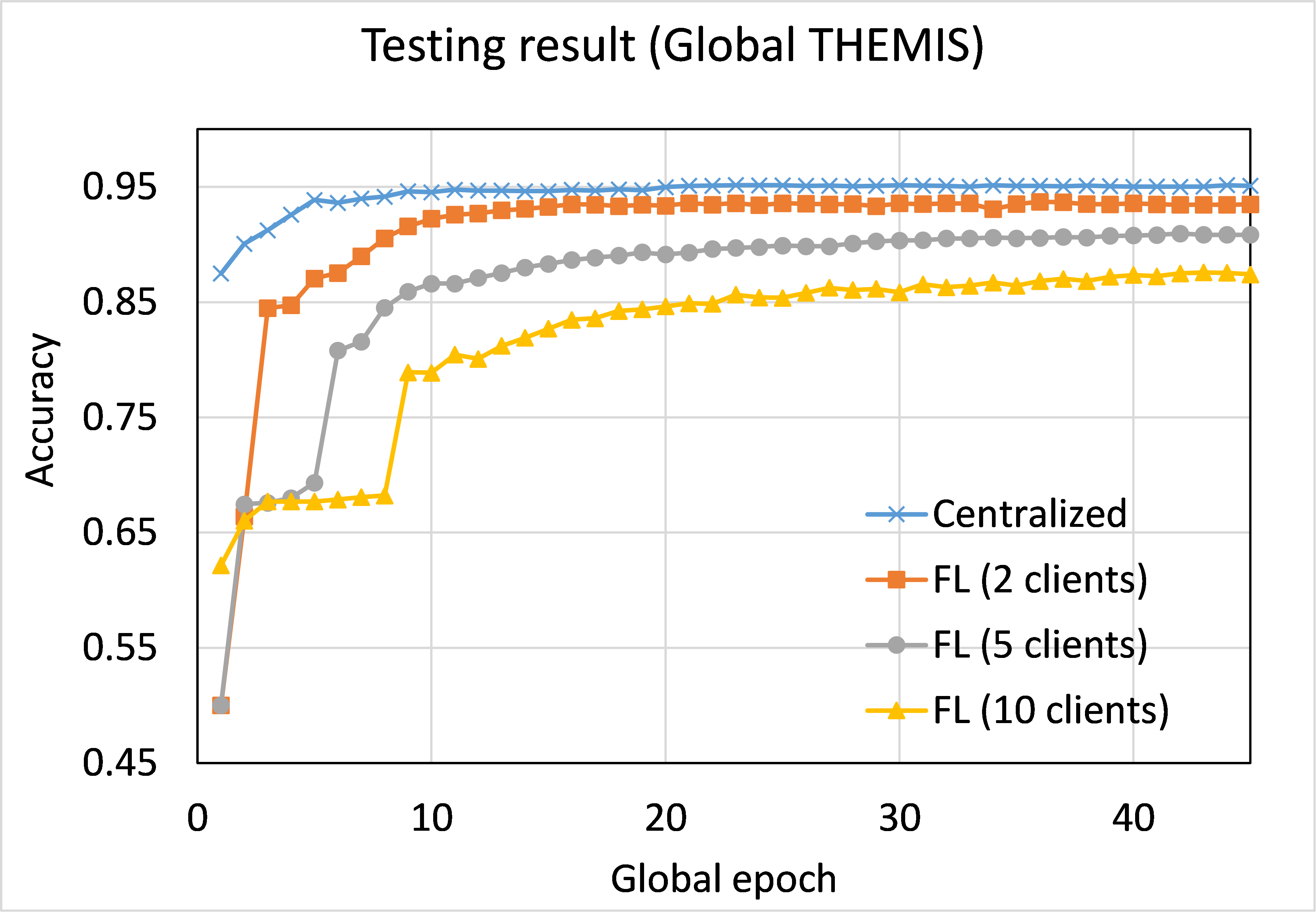}
			\label{fig:b_themis2}
		}
		\vskip5pt
		\subfigure[]{
		    \includegraphics[width=0.9\columnwidth, trim=2 1 1 1,clip]{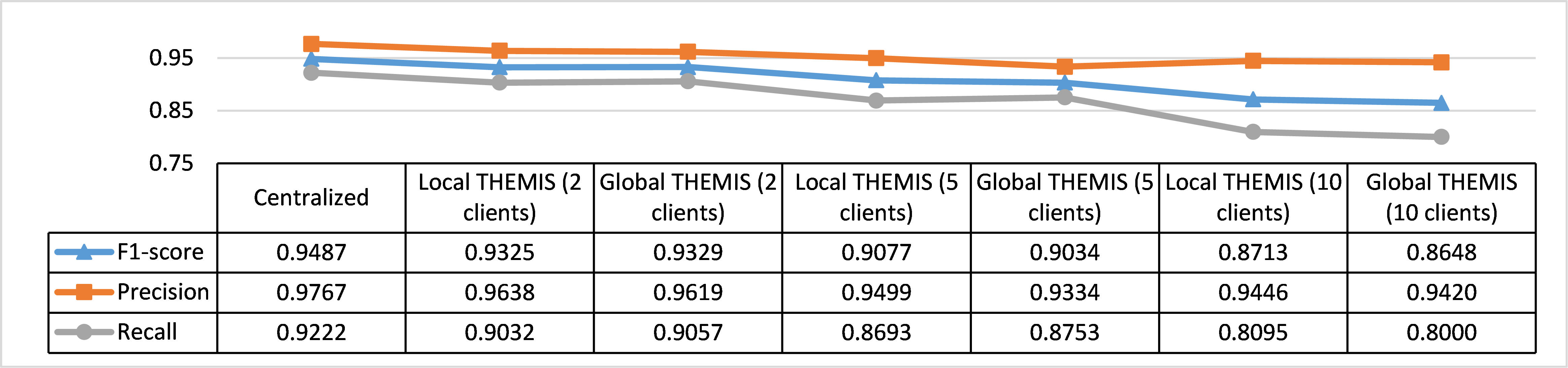}
		    \label{fig:b_themis3}
		}
		\vskip5pt
		\subfigure[]{
			\includegraphics[width=0.9\columnwidth, trim=2 2 2 2,clip]{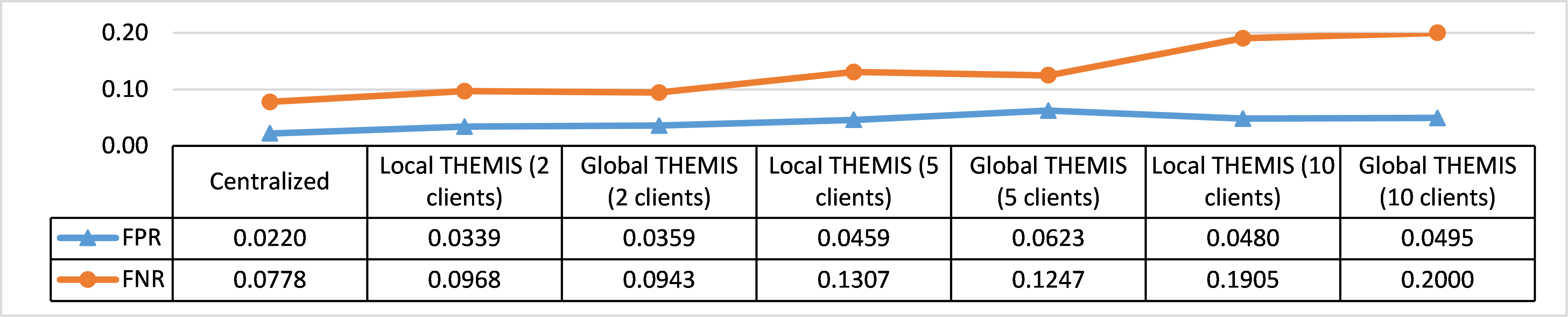}
			\label{fig:b_themis4}
		}
		\caption{Results of the THEMISb model: Convergence curves of average testing accuracy for the (a) local models and (b) global model considering body only of email samples. Corresponding performance metrics of the testing results at the global epoch of 45 in centralized and federated learning (FL) with two, five, and ten clients are depicted in (c) and (d). 
		}
		\label{fig:body_themis_exp1_appendix}
\end{figure}

Considering the CL's accuracy as the baseline, for {\bf RQ1}, {\bf RQ2}, and {\bf RQ3}, we perform experiments under a balanced data distribution among the clients where the total dataset remains the same. In other words, for the total dataset $D$, and for any number of clients $K\geq 1$, $\bigcup_k D_k = D$, where $D_k$ is the dataset of the client $k$, $k\in \{1,2,\dots, K\}$. 
We keep the same size of the total dataset despite the change in the number of clients. This is done to see the effect of the change in the number of clients (datasets distribution) within the same total dataset.
In our setups, we reasonably assume that the clients are with resourceful computation to jointly training the FL model to preserve the privacy of emails. \\
%
\textbf{How THEMIS and THEMISb perform?}\\
THEMIS outperforms THEMISb in our experiments. In CL at the 45 global epoch, for THEMIS, we observe an accuracy of 99.301\%, FPR of 0.0035, and FNR of 0.0105 (see Fig.~\ref{fig:themis_exp1}), whereas THEMISb only provides an accuracy of 95.085\% (drop by around 4\%), FPR of 0.022 and FNR of 0.0778 (see Fig.~\ref{fig:body_themis_exp1_appendix}). This indicates that header information is critical for the THEMIS model, and it is leveraging them well for phishing detection.
The accuracy and FPR stated in the THEMIS paper~\cite{fang2019phishing} are 99.848\% and 0.043\%, respectively. These values are nominally different than our case. This can be due to various reasons, including email data samples, sample size (see Section~\ref{sec:themis_model}), and model hyper-parameters.

\begin{researchq}
\textbf{Can FL be applied to learn from distributed email repositories to achieve comparable performance to centralized learning?}
\end{researchq} 

\begin{figure}[t!]
	\centering
		\subfigure[]{
			\includegraphics[width=0.65\columnwidth, trim=2 2 2 2,clip]{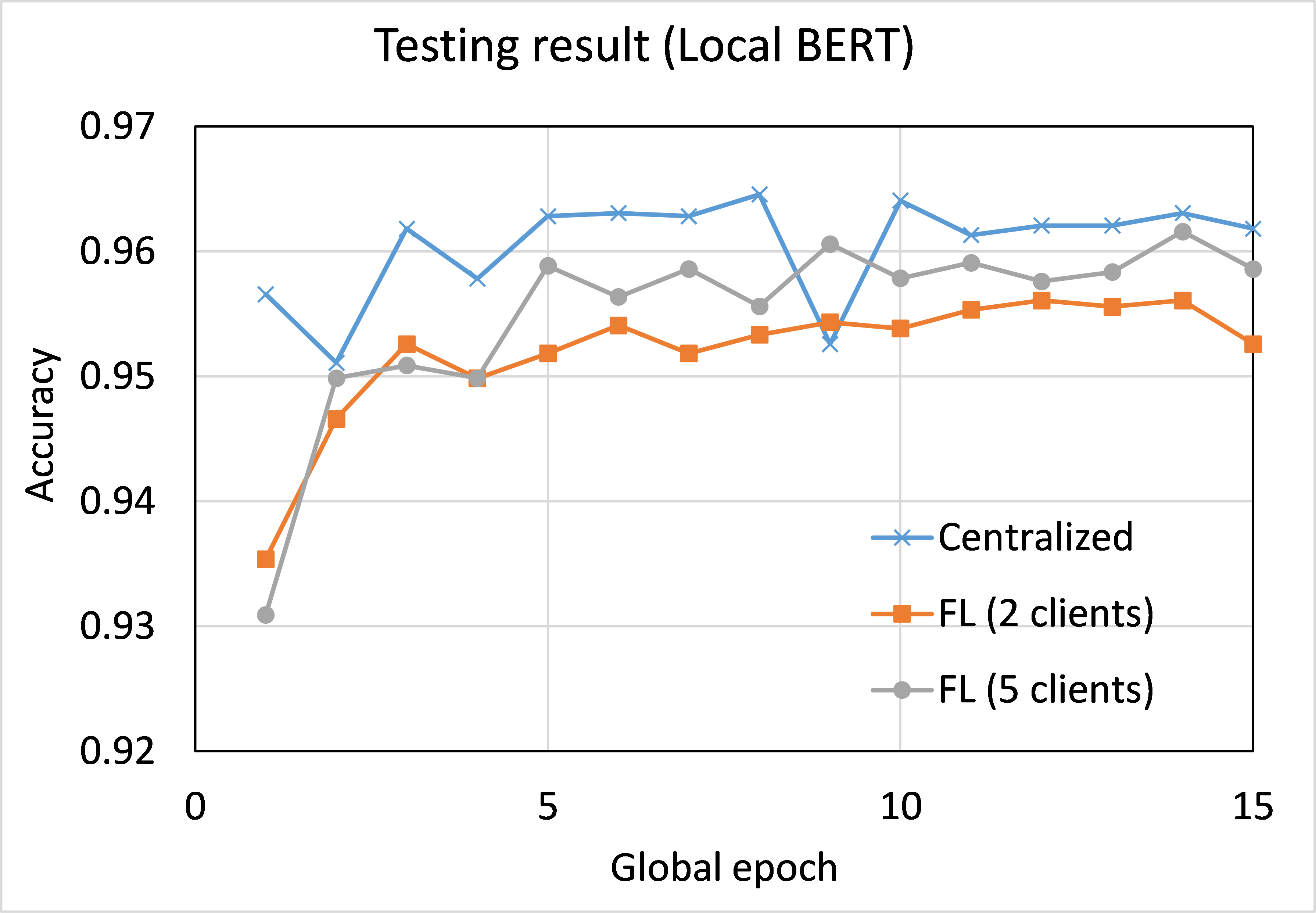}
			\label{fig:bert1}
		}
	\subfigure[]{
			\includegraphics[width=0.65\columnwidth, trim=2 2 2 2,clip]{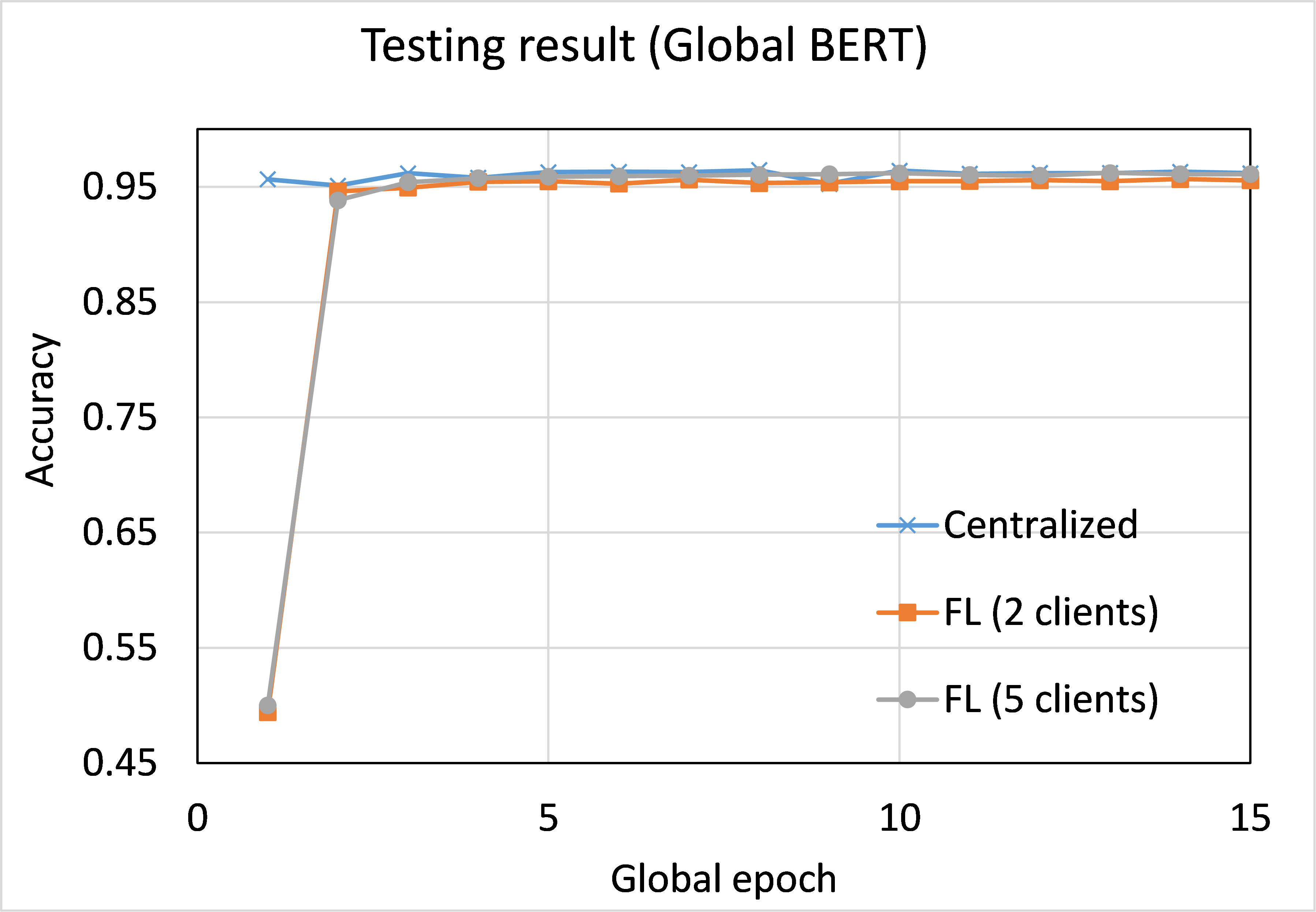}
			\label{fig:bert2}
		}
		\subfigure[]{
		    \includegraphics[width=0.9\columnwidth, trim=2 1 1 1,clip]{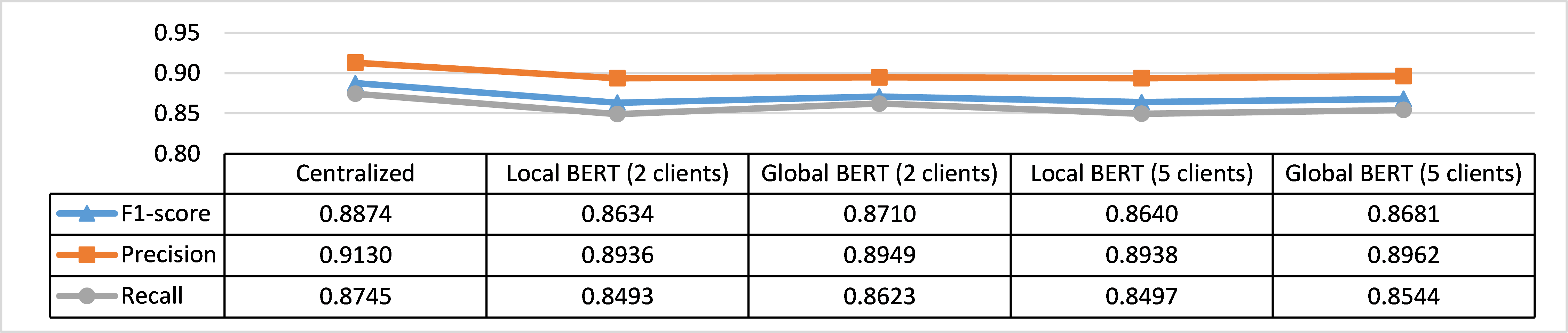}
		    \label{fig:bert3}
		}
		\subfigure[]{
			\includegraphics[width=0.9\columnwidth, trim=2 2 2 2,clip]{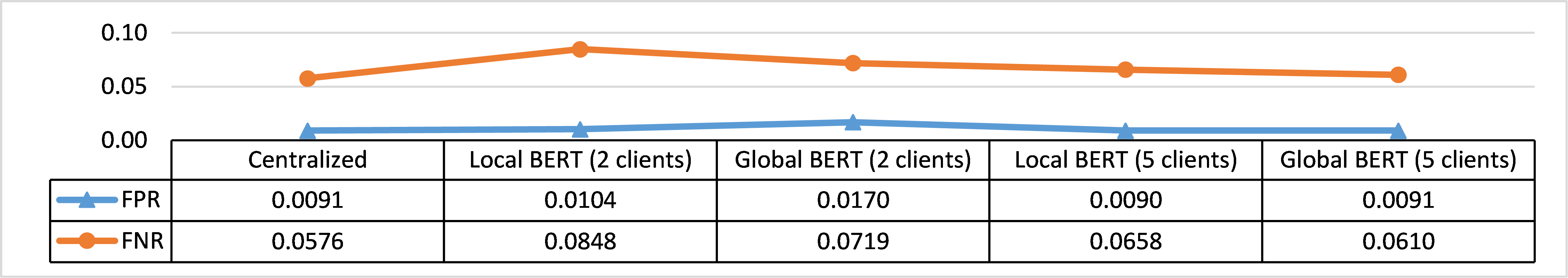}
			\label{fig:bert4}
		}
		\caption{Results for the BERT model: Convergence curves of average testing accuracy for the (a) local models and (b) global model. Corresponding performance metrics of the testing results at the global epoch of 15 in centralized and federated learning (FL) with two and five clients are depicted in (c) and (d).  
		}
		\label{fig:bert_exp1}
\end{figure}

Under FL with 2, 5, and 10 clients, the THEMIS model converges to get local and global model test accuracy greater than around 96\% in the observation window of 45 global epochs. However, none of them achieves the CL performance of 99.3\% accuracy within our observation window. Similar performance is observed with the THEMISb model.
%
For the BERT model (which only considers emails without header information while training/testing), at 15 global epoch, we have (i) in CL, testing accuracy of 96.183\%, FPR of 0.0091 and FNR of 0.0576, (ii) in FL with two clients, global testing accuracy of 95.559\%, FPR 0.017 of and FNR of 0.0719, and (iii) in FL with five clients, global testing accuracy of 96.11\%, FPR 0.0091 of and FNR of 0.0610 (see Fig.~\ref{fig:bert_exp1}).
We find that BERT performance is not good as THEMIS, but it is better than THEMISb in our observations.

\begin{mdframed}[backgroundcolor=black!10,rightline=false,leftline=false,topline=false,bottomline=false,roundcorner=2mm]
	\textbf{Summary:} FL is feasible with comparable performance to the CL for phishing email detection. It enables privacy benefits to the system, but it could not achieve the CL performance as a trade-off in our experiments. 
\end{mdframed}

\begin{researchq}
\textbf{How would the number of clients affect FL performance and convergence?}
\end{researchq}

For FL with THEMIS model, the convergence patterns for both local and global models are similar. However, in both cases, the average performances degraded with the increase in the number of clients. For example, we observe the global testing accuracy at the 45 global epoch drops by 1.8\% going from two clients setup to ten clients setup with the THEMIS model (see Fig.~\ref{fig:themis_exp1}). The potential reason for this drop might be the effect on convergence rate due to local shuffling in distributed setup; the convergence rate is dominated by local training size and more the better~\cite{shuffling}. For THEMISb, the drop is about 6\% going from two clients to ten clients. This drop in THEMISb is significant in comparison to THEMIS.  

For FL with BERT, the local and global BERT model performance drop is negligible compared with the BERT in CL, e.g., only 0.6\% for 2 clients (see Fig.~\ref{fig:bert_exp1}). In contrast to THEMIS/THEMISb performance, the BERT performance does not degrade with the increase in clients, as the BERT with 5 clients is showing better performance than BERT with 2 clients by 0.6\% in accuracy at the 15 global epoch.

\begin{mdframed}[backgroundcolor=black!10,rightline=false,leftline=false,topline=false,bottomline=false,roundcorner=2mm]
	\textbf{Summary:} The convergence and performance with the increase in the number of clients in FL are model-dependent. For THEMIS, there are slower convergences and performance degradation with the increase in clients, whereas BERT has the opposite in our experiments.
\end{mdframed}

\begin{researchq}
\textbf{What is communication overhead resulting from FL?}
\end{researchq}
As the main server is assumed to have sufficient resources to handle any communication overhead, our concern is with clients who have relatively low resources than the server. Thus the quantification of communication overhead in FL is limited to the client-side. 
We measure the data uploaded (i.e., a sum of the data packet size of $W_t^k$ and $n_k$) and download (i.e., data packet size of $W_{t+1}$) to and from the server, respectively, and it is averaged by the total number of the clients. In CL, we do not consider a client-server setup; thus, the communication overhead is zero. 

As the sample size information is negligible comparing the model size, the download and upload are almost the size of the global model and the local model, respectively while training, at each client in FL.
Thus the communication overhead solely depends on the model size and thus not on the number of clients or epochs. This is verified from our experiments with the various number of clients. For THEMIS and BERT models, we observe a consistent average communication overhead of around 0.192GB and 0.438GB per global epoch per client, respectively, for all cases. 
The overhead can be easily addressed by a well-connected setup with wired or wireless connections between the server and clients who participated in the anti-phishing framework. Thus this is not a concern for organizational-level participation in distributed email learning as these organizations are usually resourceful clients.

\begin{mdframed}[backgroundcolor=black!10,rightline=false,leftline=false,topline=false,bottomline=false,roundcorner=2mm]
	\textbf{Summary:} Communication overhead per client per global epoch is the model's size-dependent.  
\end{mdframed}

\subsection{Client-level  perspectives  in  FL}~\label{sec:transfer learning}

To demonstrate the client-level performance in phishing email detection in distributed setups, we perform three experiments under a balanced and asymmetric data distribution among the clients, where the total dataset changes with the number of clients. In other words, for any number of clients $K\geq 1$, if $\bigcup_k D_k = D'$, where $D_k$ is the dataset of the client $k$, $k\in \{1,2,\dots, K\}$, then for $K+1$ clients, $\bigcup_k D_k = D$ such that $D'\subset D$. 

For this section, asymmetric data distribution is only due to the different sample sizes among clients but with the equal number of phishing and legitimate emails. The variation in the local data sizes is based on the maximum percentage of the variation provided by the term ``$\mathsf{var}$.'' For example, if $\mathsf{var} = 10\%$, then the variation of the data across the five clients is given by $[-10\%, -5\%, 0\%, +5\%, +10\%]$, where -10\% referred to the 10\% less local data, and +10\% referred to the 10\% more data in the respective clients. This means, 3606, 3806, 4008, 4208, and 4408 local data samples are resided in clients 1, 2, 3, 4 and 5, respectively, if $\mathsf{var} = 10\%$. This way, we create a variation of the sizes of the local data by maintaining the total size of the datasets. Besides, the balanced data distribution is the case where $\mathsf{var} = 0$.  
%

\begin{researchq}
 \textbf{Can a client leverage FL to improve its performance?}
\end{researchq}
For this research question, we limit our experiments to the THEMIS model (considering both email's header and body information) because BERT has high training/testing overhead if we go up to 50 global epochs. 

\subsubsection{Experiment 1: A client-level and overall effects of adding one new client in FL}
In this experiment, we consider five clients in total, where the first four clients (C1--C4) participate in the FL until 15 global epochs and train the model collaboratively. Afterward, the learning is carried out only with the fifth client, and the training proceeds for the next 15 global epochs (i.e., until 30 global epochs). 
Besides, the testing results are computed for all five clients throughout the process for the performance evaluation. This experiment examines how a newly joining client member is benefited in FL by improving its performance in phishing detection.  

\begin{figure}[t]
	\centering
		\subfigure[]{
			\includegraphics[width=0.5\columnwidth, trim=2 2 2 2,clip]{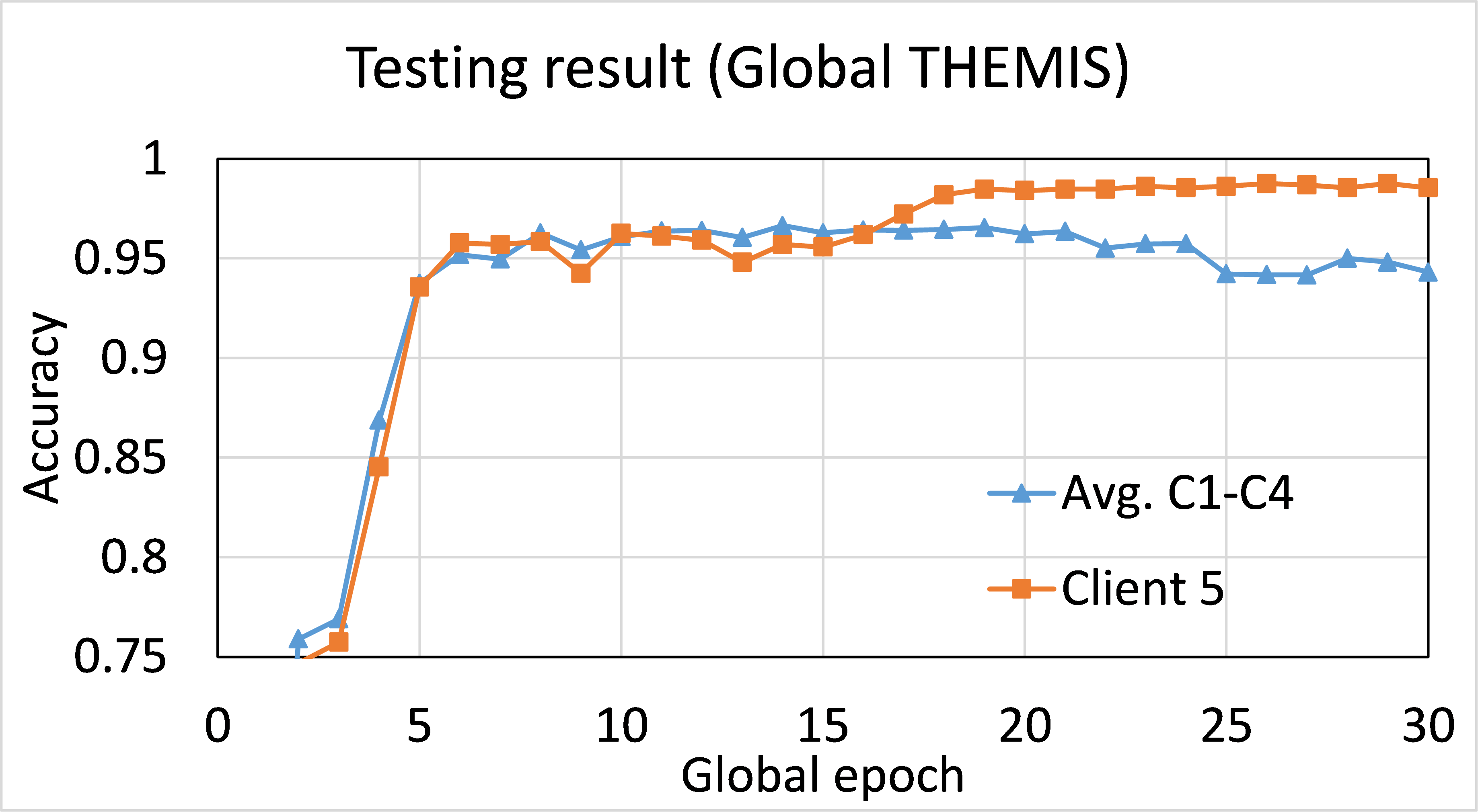}
		}
		\hskip-10pt
		\subfigure[]{
			\includegraphics[width=0.47\columnwidth, trim=2 2 2 2,clip]{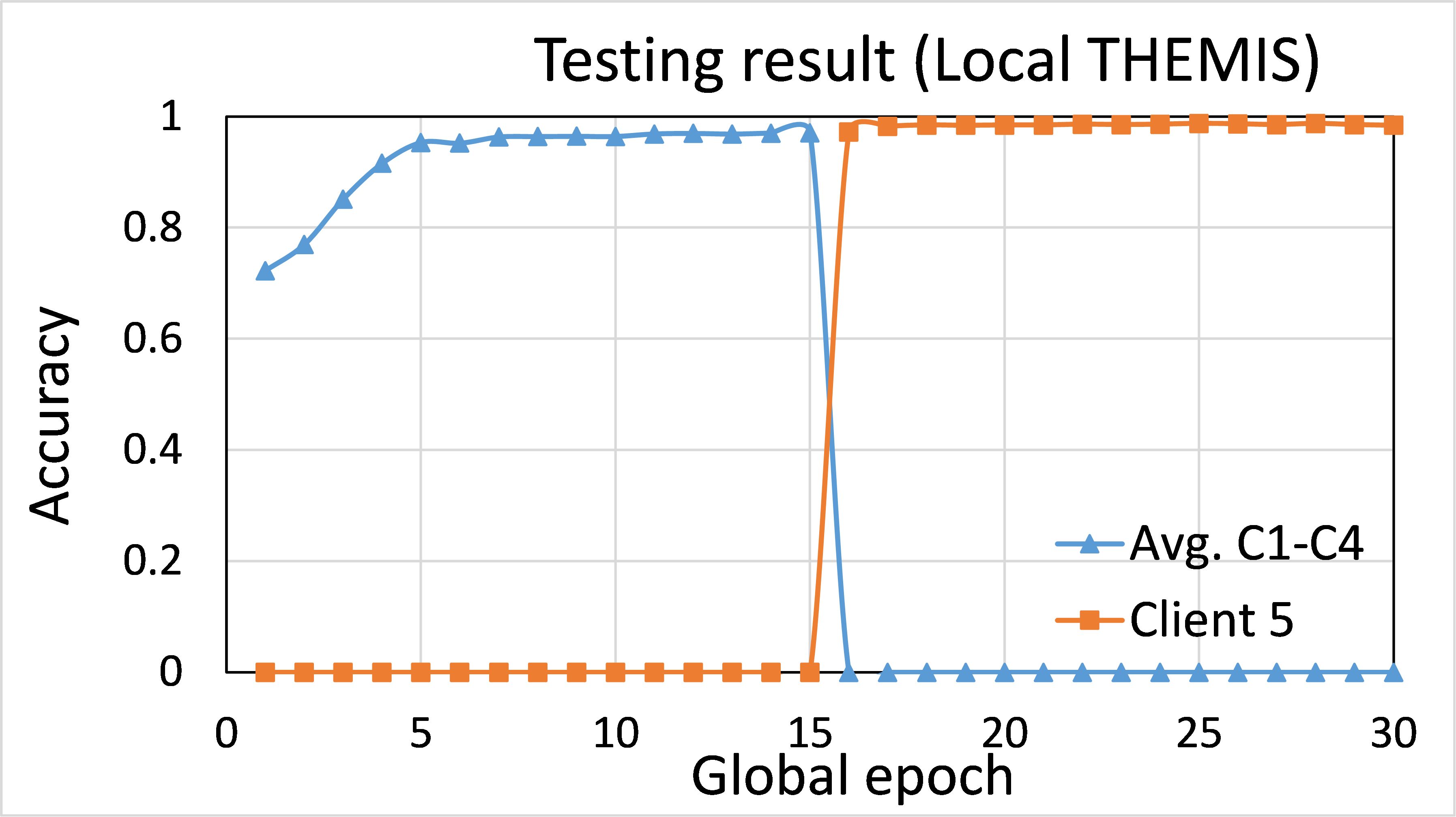}
		}
		\caption{Convergence curves of (a) global testing accuracy and (b) local testing accuracy from the Experiment~1 with five clients and $\mathsf{var} = 80$. The first four clients train the model until 15 global epochs, and then (only) the fifth client trains the model.}
		\label{fig:asyn_var80}
\end{figure}

The experimental result depicted in Fig.~\ref{fig:asyn_var80} is for the case with $\mathsf{var} = 80$, which provides the variations in the sizes of the local dataset (i.e., [-80\%, -40\%, 0\%, +40\%, +80\%]) to capture a practical setting among the five clients. The figure shows that the average global test accuracy of the first four clients is slightly higher than the fifth client (not participated in the learning process) until 15 global epochs. Afterward, the fifth client trains the model, so its average global testing accuracy improves by 2.98\%, and FPR and FNR improve by 3.5\% and 2.2\%, respectively. This performance decreases with the lesser variation in the sizes of the local dataset; the improvements in average global test accuracy are 2.91\%, 2.87\%, and 2.24\% with $\mathsf{var}$ equal to 50, 30, and 0, respectively. Refer to Fig.~\ref{fig:asyn_var0} in Appendix for the results with the balanced email distribution, i.e., $\mathsf{var} = 0$. 
%
%
Overall results show that the evolved model (after training by client 5) is still relevant to the first four clients (C1--C4) as their average testing results with and without client 5 differ only nominally. Nonetheless, the fifth client boosts the accuracy of phishing detection in its local dataset under the FL setup.

\subsubsection{Experiment 2: A client-level and overall effects of continuously adding new clients in FL}
In this experiment, the learning process is started with the first client, and then one new client is joined continuously at an interval of 10 global epochs as the training proceeds. Refer to Table ~\ref{tab:explain} for details. This experiment simulates the practical cases where more than one client (different than the Experiment~1) is available with time during model training and demonstrates how the newly available clients can continue to perform FL to contribute accuracy improvements for phishing detection.

\begin{figure}[t]
	\centering
		\subfigure[]{
			\includegraphics[width=0.65\columnwidth, trim=2 2 2 2,clip]{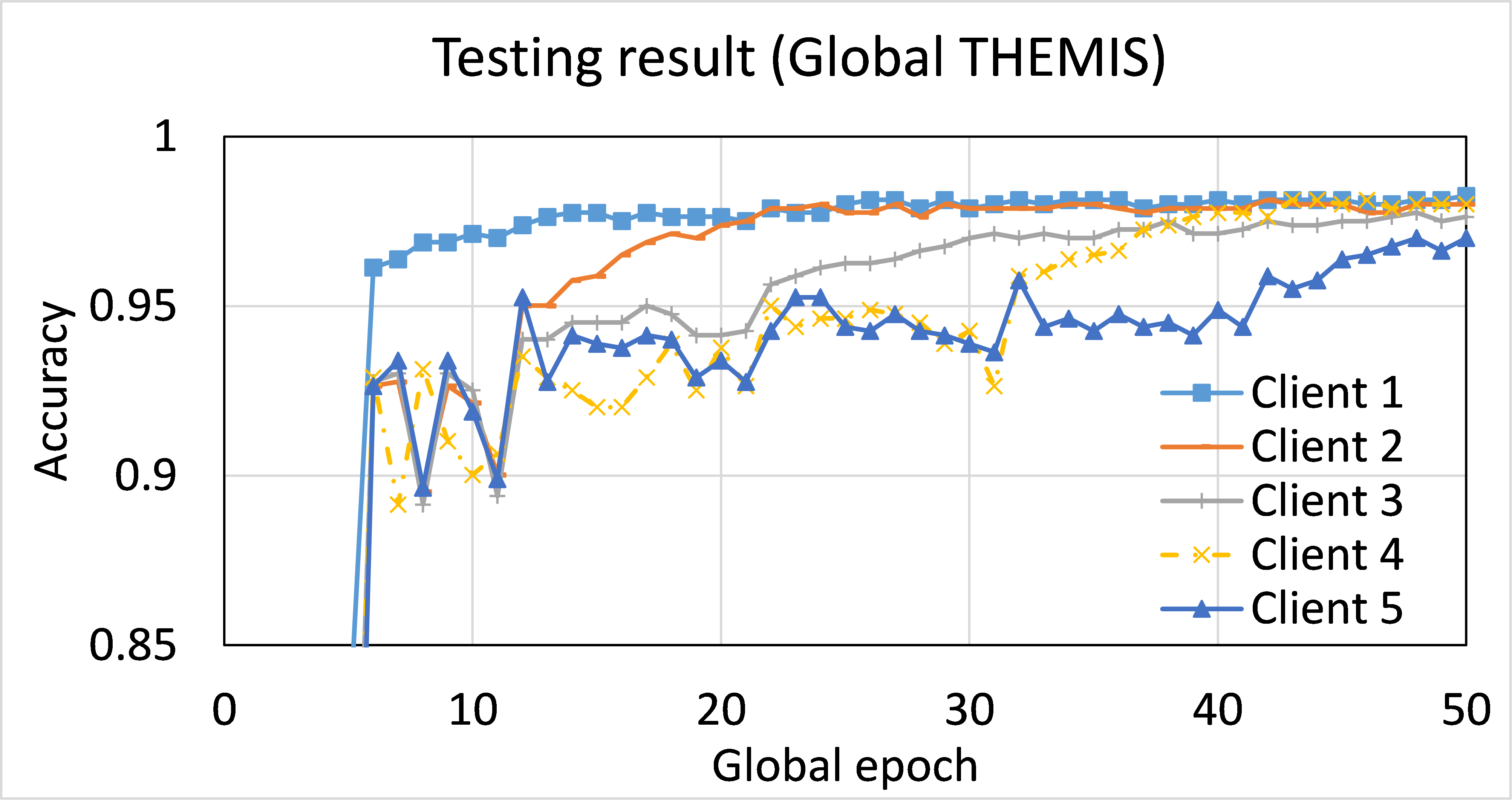}
		}
		\vskip10pt
		\subfigure[]{
			\includegraphics[width=0.65\columnwidth, trim=2 2 2 2,clip]{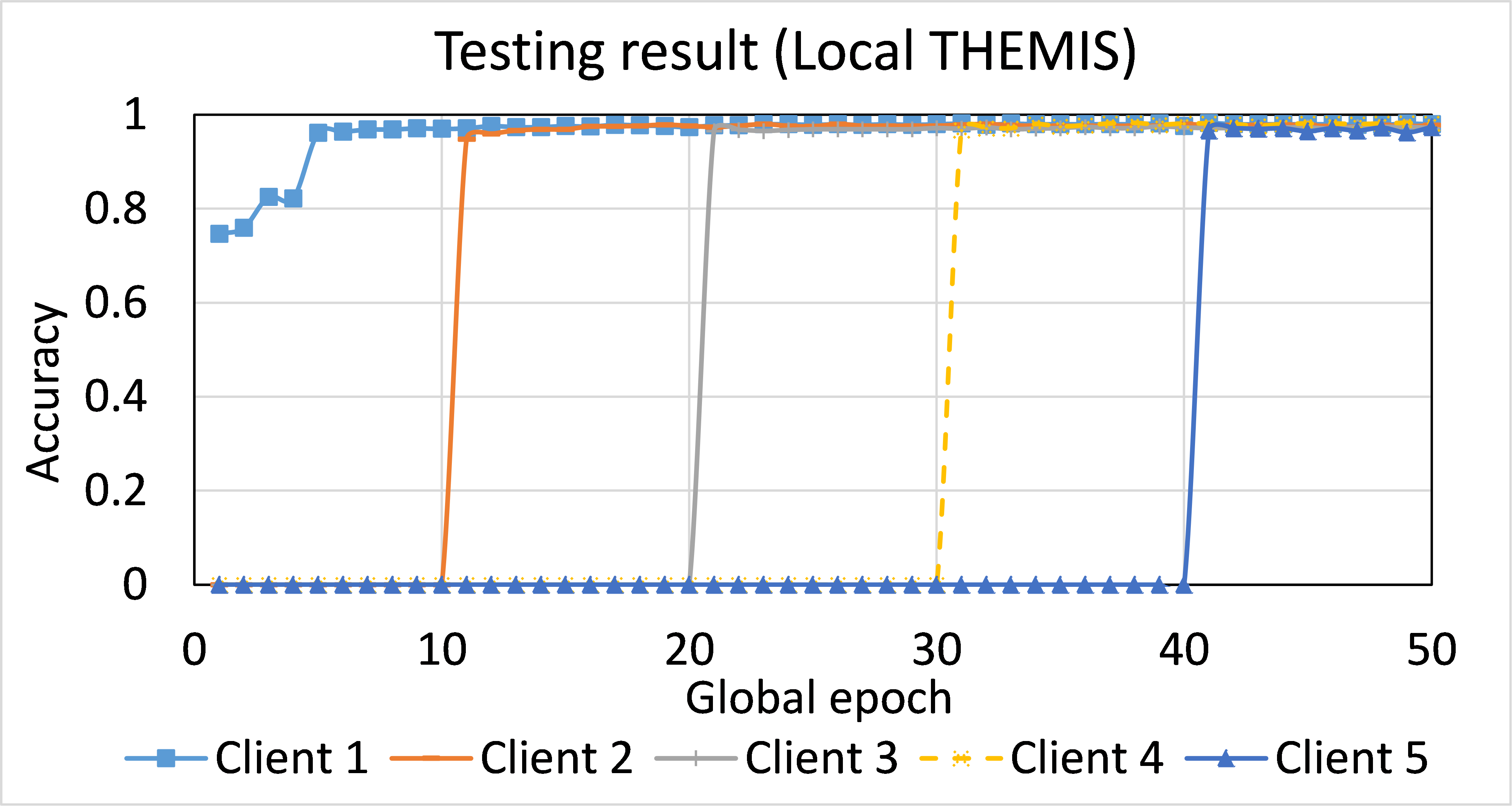}
			\label{figb:asyn2_var0}
		}
		\caption{Convergence curves of (a) global testing accuracy and (b) local testing accuracy from the Experiment 2 with five clients and $\mathsf{var} = 0$. The FL training starts with one client, i.e., client~1, and a new client joins the training at every 10-th global epochs in a sequence from client~2 to client~5.}
		\label{fig:asyn2_var0}
\end{figure}

\begin{table}[t]
\centering
\footnotesize
\caption{Experimental steps for the Experiment~2}
\label{tab:explain}
\begin{tabular}{|c|c|}
\hline
\textbf{Round}                                                                                                               & \textbf{Involvement of clients}                                                                                                                       \\ \hline
0 to 9                                                                                                                       & Only the first client.                                                                                                                                   \\ \hline
10 to 19                                                                                                                     & Only the first and second client.                                                                                                                        \\ \hline
20 to 29                                                                                                                     & Only the first, second, and third client.                                                                                                                 \\ \hline
30 to 39                                                                                                                     & First, second, third, and fourth client.                                                                                                              \\ \hline
40 to 50                                                                                                                     & All five clients.                                                                                                                                    \\ \hline
\multicolumn{2}{|c|}{\begin{tabular}[c]{@{}c@{}} The local and global test accuracy are measured\\ for all clients throughout the process.\end{tabular}} \\ \hline
\end{tabular}
\end{table}

The result depicted in Fig.~\ref{fig:asyn2_var0} is for the case with the same size of the local dataset among the five clients (i.e., $\mathsf{var} = 0$), which are gradually added to the learning process, as stated in Table~\ref{tab:explain}. As per expectation, it shows that the global testing accuracy improves for each client when it is added to FL. For example, the average global testing accuracy jumps by around 4.9\% (corresponding to the accuracy at the 10 and 19 global epoch) for client~2 when it joins client 1 in training the model at global epoch 10. 
The local testing results are carried only when the client is involved in the model training. Thus the local testing accuracy before a client joins the training is zero in Fig.~\ref{figb:asyn2_var0}. 
The overall performance pattern for the case with $\mathsf{var} \in \{30, 50, 80\}$ is similar to the case with $\mathsf{var}=0$. However, we observe the dominance of late joining clients (e.g., client 4 and 5) in their performance since the initial clients (e.g., client 1) have a fewer number of samples than the late joining clients if $\mathsf{var}\neq 0$. Moreover, the initial client, such as client 1, could not catch up with the performance of the late joining client, such as client 5 (for details, refer to Fig.~\ref{fig:asyn2_var80} in Appendix).


\subsubsection{Experiment 3: Benefits to the newly participated client in the FL learning process}
In this experiment, we analyze the performance of client~1, which we assume a newly participating client, with and without leveraging FL. In the FL setup, the model is first trained by the four clients (client~2 to client~5) for 20 global epochs, and then the resulting model (pre-trained model) is further trained by client~1 on its local email data samples. The dataset distribution of the clients is defined by $\mathsf{var}$. On the other hand, for the case without FL, client~1 performs CL only on its local email dataset.    

\begin{figure}[htb]
	\centering
		
		\includegraphics[width=0.65\columnwidth,trim=2 2 2 2,clip]{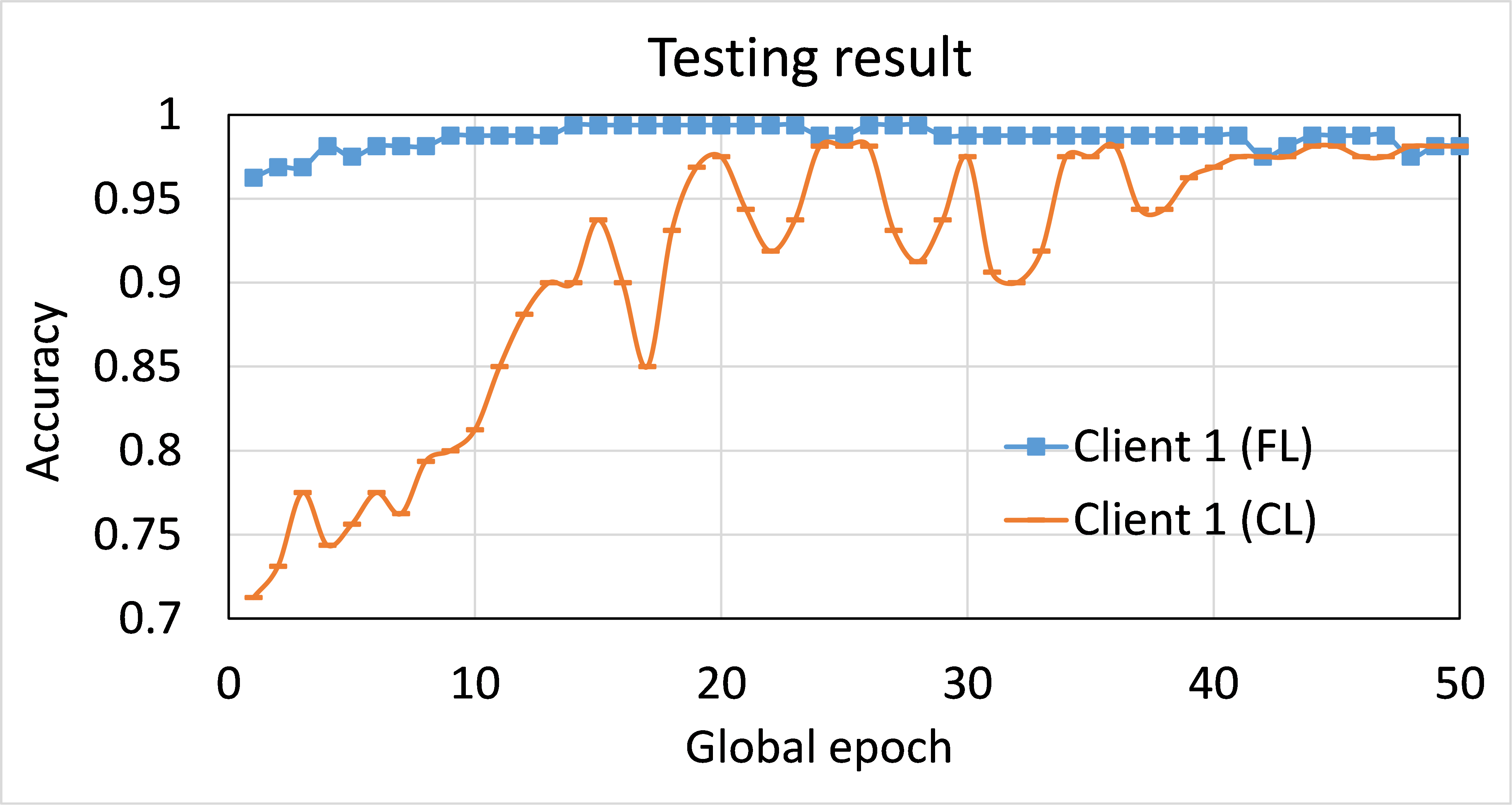}
			
		\caption{Convergence curves of (local) testing accuracy for the client~1 with and without leveraging FL. The dataset of client~1 is based on $\mathsf{var} = 80$ among five clients.}
		\label{fig:asyn2_var80_e3}
\end{figure}
The result depicted in Fig.~\ref{fig:asyn2_var80_e3} is for the client~1, and its dataset is defined under five clients setup with $\mathsf{var} = 80$ (this means that client~1 has 80\% fewer data samples than client~3). It shows that the client~1 achieves a fast convergence and stable output by leveraging FL compared to its training under CL over its local dataset. However, the same final accuracy of around 98\% is observed for both cases at the 50 global epoch.  
If the client~1's dataset is assigned based on $\mathsf{var} = 0$ (balanced case) with five clients, then the convergence curves for the client~1 are more stable and flat after 10 global epoch; however, a fast convergence is consistently observed for this case when leveraging FL as well. Refer to Fig.~\ref{fig:asyn_var0} in the  Appendix for the results.


\begin{mdframed}[backgroundcolor=black!10,rightline=false,leftline=false,topline=false,bottomline=false,roundcorner=2mm]
	\textbf{Summary:} Based on the results in this section, the answer to the {\bf RQ4} is affirmative. The results demonstrate that FL is useful for performance-boosting, including fast convergence, in phishing detection training at the client level under the distributed setups.
\end{mdframed}
\vskip-10pt

\subsection{Distributed email learning under asymmetric data distribution} \label{sec:imbalance data}
We examine the performance of FL under an asymmetric dataset distribution mainly in two forms; (1) different sample size among clients (defined by $\mathsf{var}$) but with an equal number of phishing and legitimate emails, and (2) same sample size but the different number of phishing and legitimate emails. This setup is not precisely a non-IID distribution, which is due to the high skewness both in the number of samples and their classes present in each client's dataset.

\begin{researchq}
\textbf{How would FL perform under asymmetric data distribution among clients due to the variations in the local dataset’s size and the local phishing to legitimate sample ratio?}
\end{researchq}

%
%
We perform two experiments based on the phishing to legitimate email samples (P/L) ratio among clients.

\begin{figure}[tbh]
	\centering
		\subfigure[]{
			\includegraphics[width=0.65\columnwidth, trim=2 2 2 2,clip]{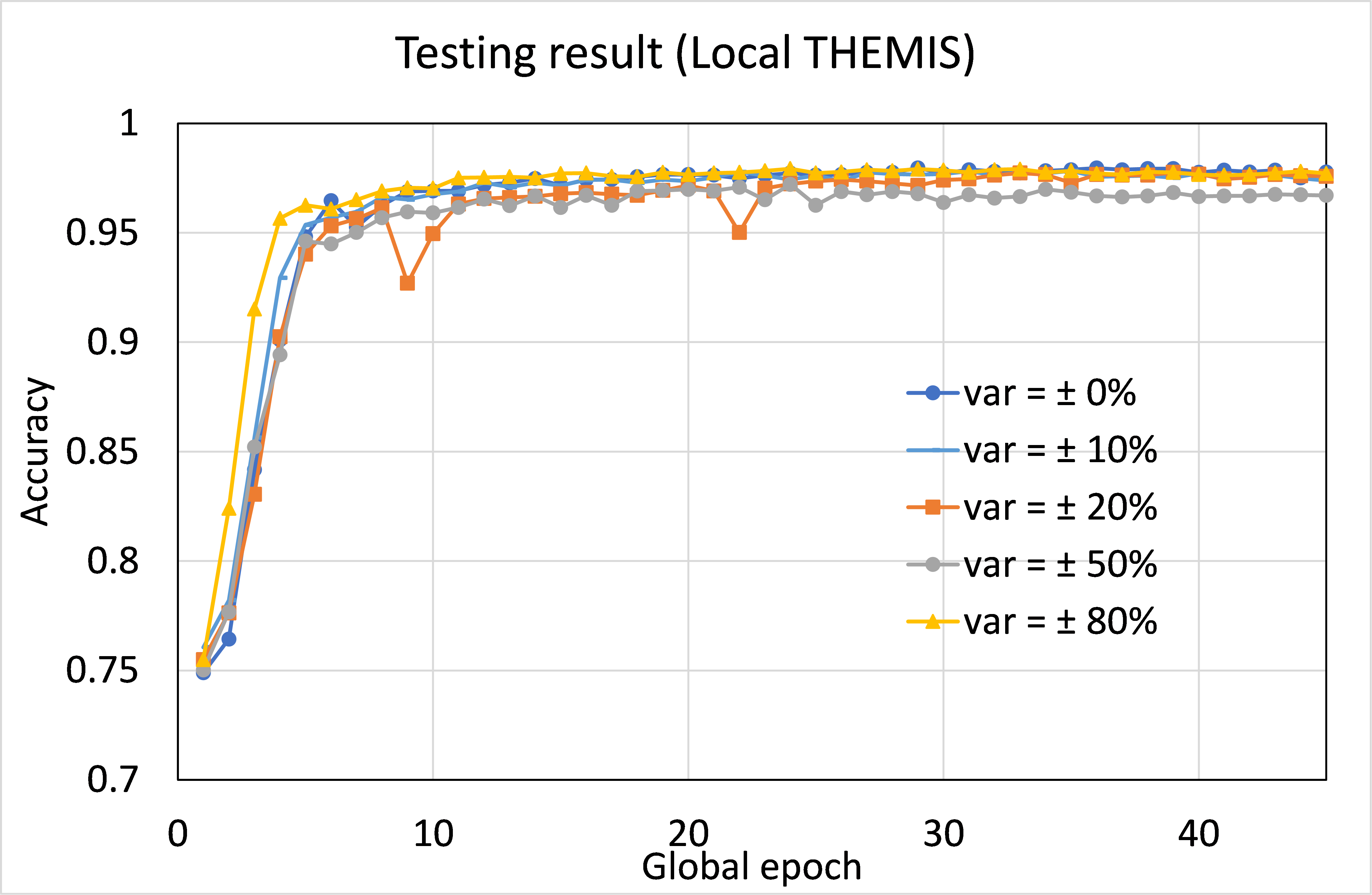}
			\label{fig:themis1b}
		}
	\subfigure[]{
			\includegraphics[width=0.65\columnwidth, trim=2 2 2 2,clip]{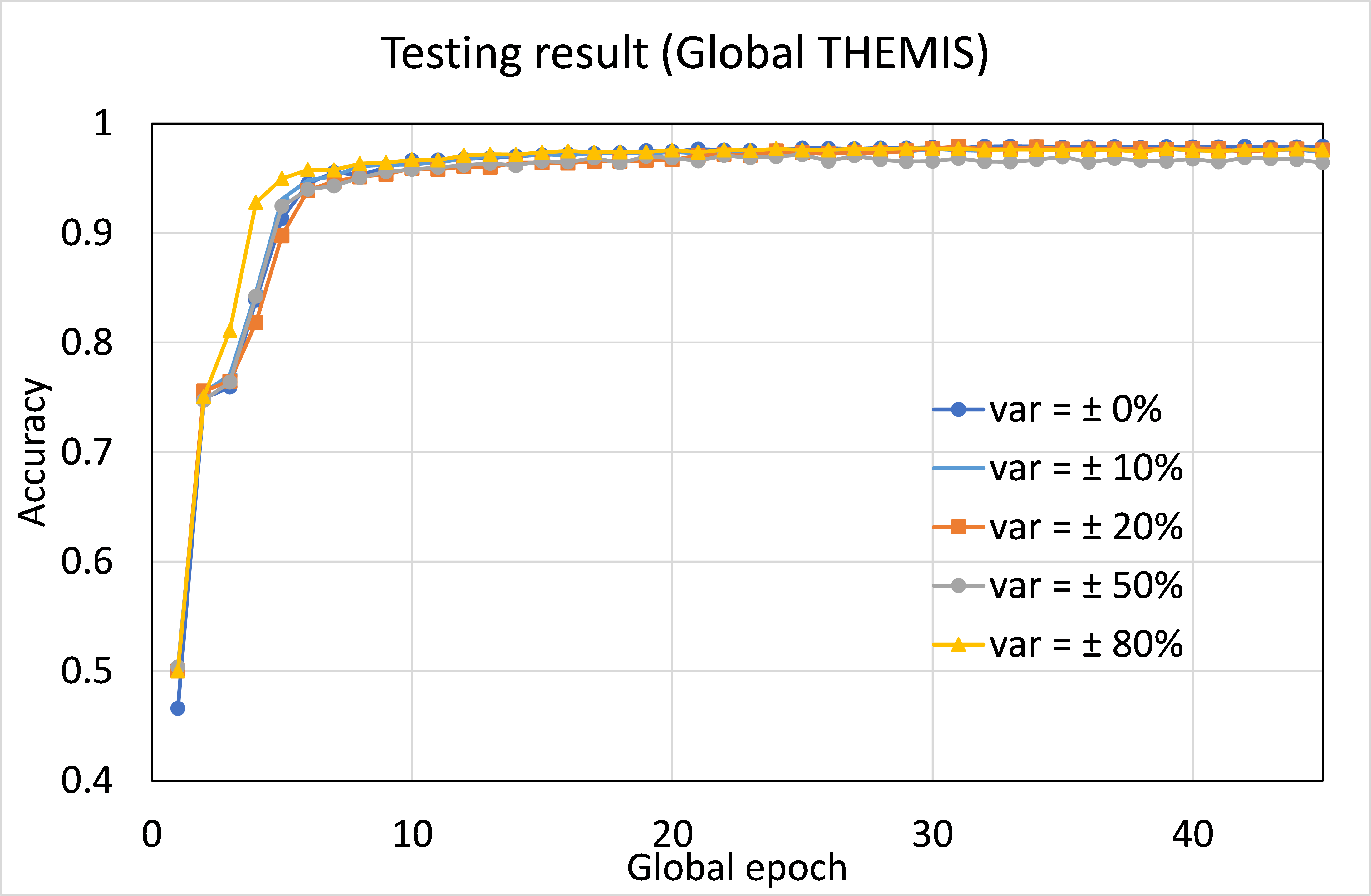}
			\label{fig:themis2b}
		}
		\subfigure[]{
		    \includegraphics[width=0.95\columnwidth, trim=2 1 1 1,clip]{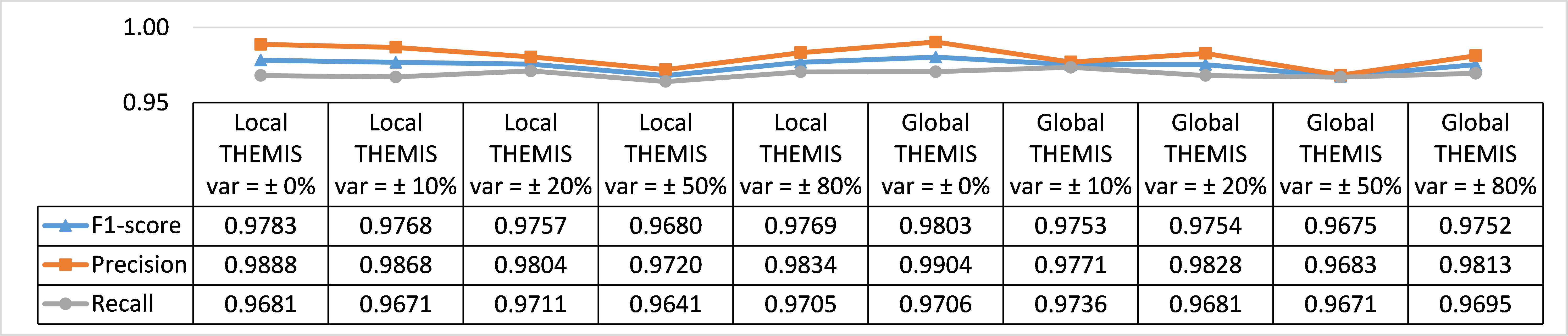}
		    \label{fig:themis3b}
		}
		\subfigure[]{
			\includegraphics[width=0.95\columnwidth, trim=2 2 2 2,clip]{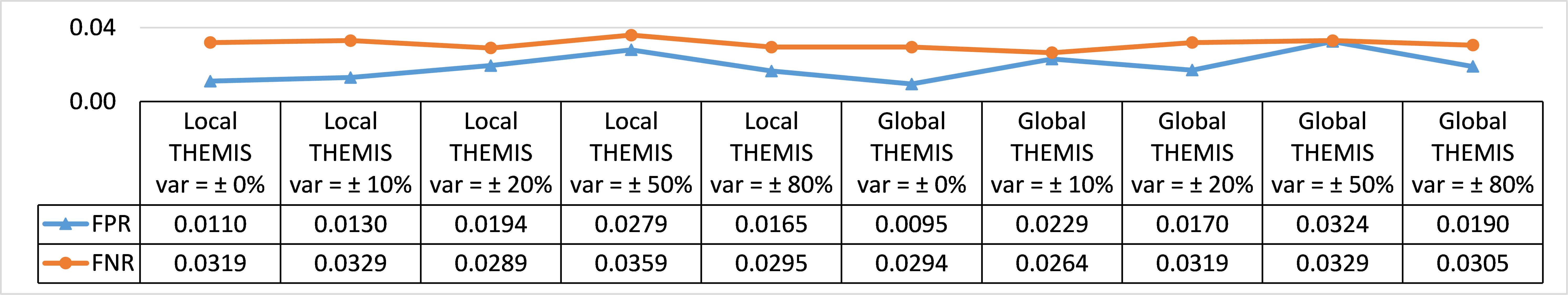}
			\label{fig:themis4b}
		}
		\caption{Showing the impact of different local data sizes provided by different $\mathsf{var}$ among clients to their convergence in FL with five clients on testing accuracy curves of (a) local model, and (b) global model. Corresponding performance metrics of the testing results at 45 global epoch in centralized and federated learning (FL) with five clients are depicted in (c) and (d).}
	\label{fig:data_size}
\end{figure}

\subsubsection{Same P/L ratio across clients but having different sizes of the local dataset}

\begin{figure}[H]
	\centering
		\subfigure[]{
			\includegraphics[width=0.7\columnwidth, trim=2 2 2 2,clip]{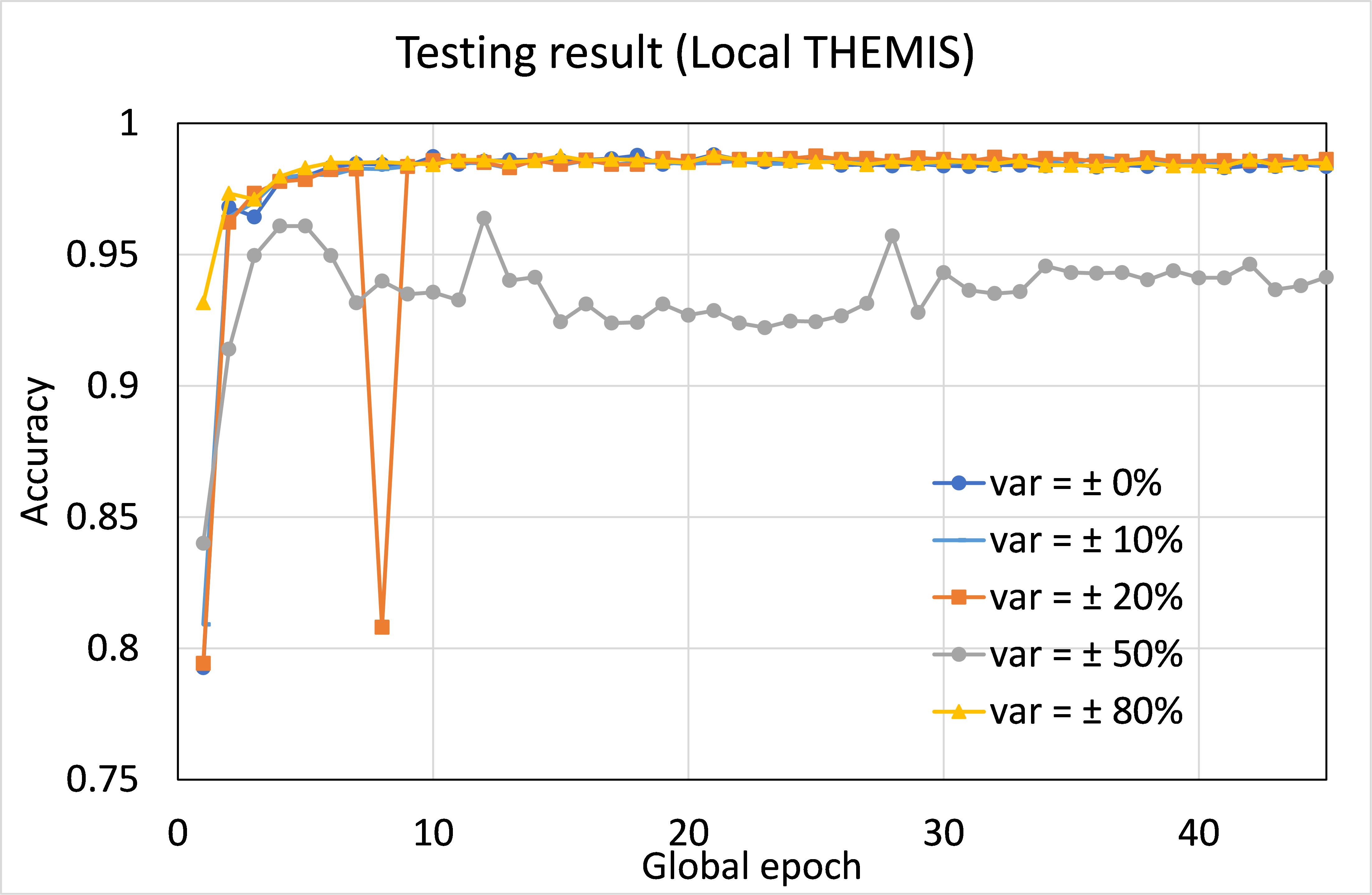}
			\label{fig:themis1d}
		}

		
	    \subfigure[]{
			\includegraphics[width=0.7\columnwidth, trim=2 2 2 2,clip]{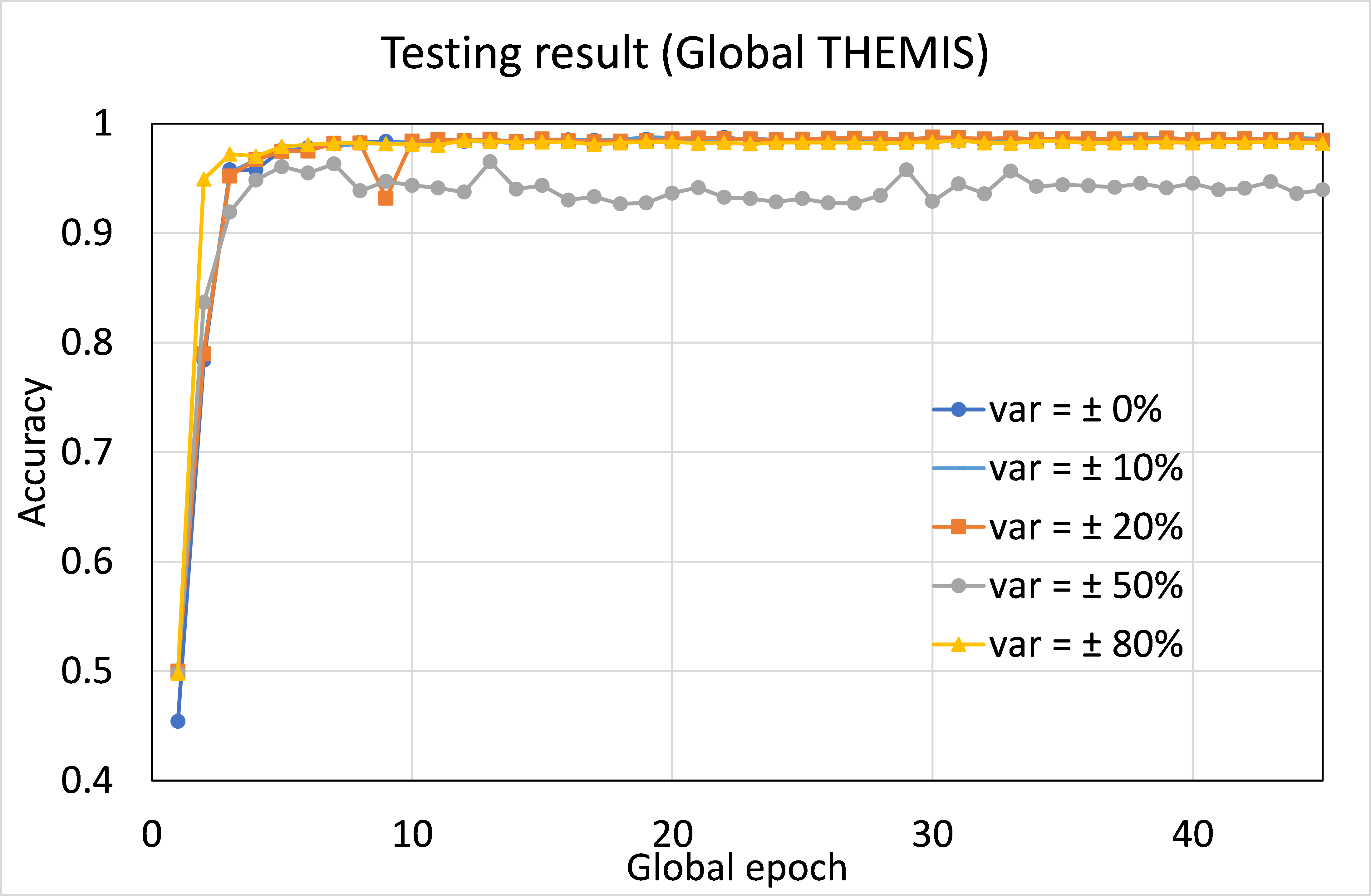}
			\label{fig:themis2d}
		}
		\subfigure[]{
		    \includegraphics[width=0.9\columnwidth, trim=2 1 1 1,clip]{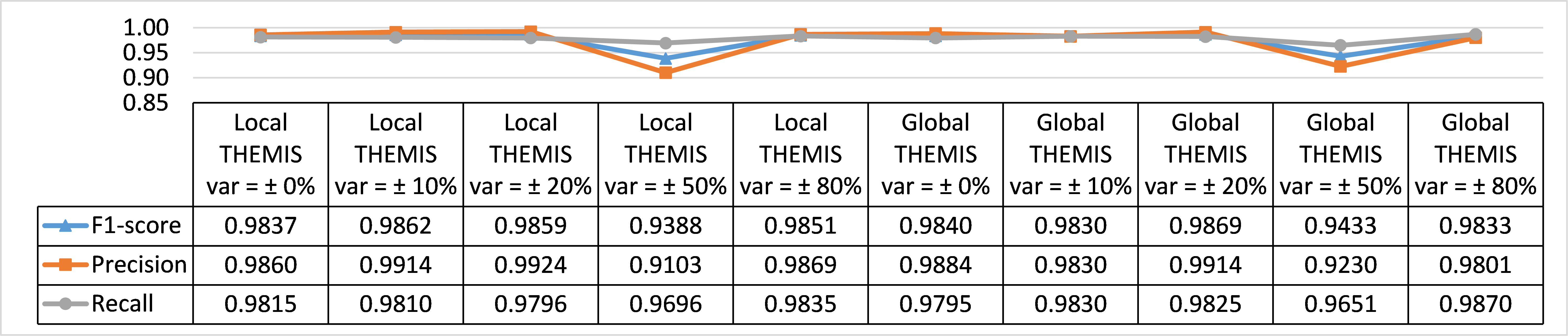}
		    \label{fig:themis3d}
		}
		\subfigure[]{
			\includegraphics[width=0.9\columnwidth, trim=2 2 2 2,clip]{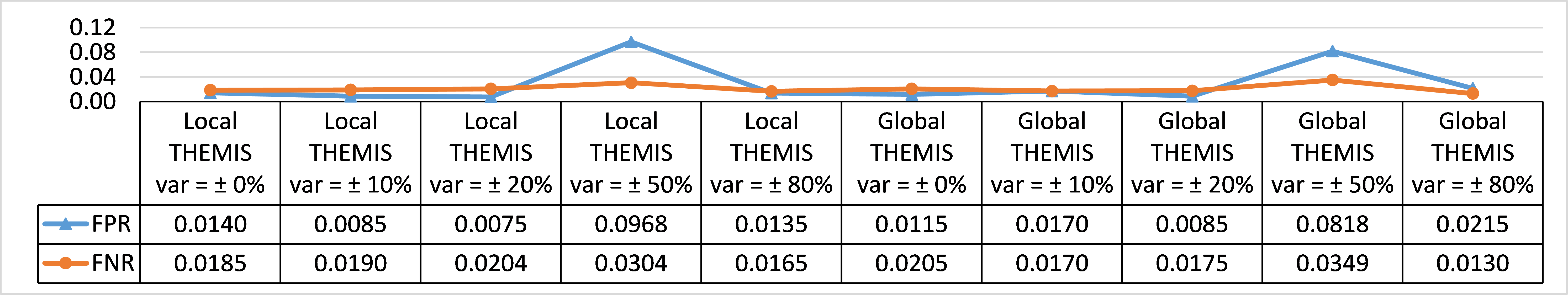}
			\label{fig:themis4d}
		}
		\caption{Showing the impact of different local data sizes provided by different $\mathsf{var}$ among clients to their convergence in FL with two clients on testing accuracy curves of the (a) local model, and (b) global model. Corresponding performance metrics of the testing results at the global epoch of 45 in centralized and federated learning (FL) with ten clients are depicted in (c) and (d).}
	\label{fig:data_size_appendix}
\end{figure}

The result for 0\% (balanced), 10\%, 20\%, 50\%, and 80\% variations in the sizes of the local dataset in FL among 5 clients is depicted in Fig.~\ref{fig:data_size}. The result shows that the convergence of the test accuracy curves rises until the global epoch of 10, then remains almost flat afterward. 
All cases with different $\mathsf{var}$ maintain an overall testing accuracy of around 97\% and similar FPR and FNR at the 45 global epoch (Fig.~\ref{fig:data_size}). We observe similar performance patterns for 10 clients. However, for 2 clients, the test performances are improved relative to 5 or 10 clients for all cases except with $\mathsf{var} = 50$ (refer to Fig.\ref{fig:data_size_appendix} for details).
The results show that the closeness of the performance for $\mathsf{var} = 50$ to other cases of $\mathsf{var}$ increases with the increase in the number of clients; it achieves the same performance as others for ten clients. 

For most cases, the similarity in performance despite variations in the local data sizes amongst clients indicates the FL's resilience (mostly enabled by weighted averaging) to the data size variations.

\subsubsection{Different legit email to phishing email sample ratio across clients but having the same sizes of the local dataset}
\begin{figure}[t]
	\centering
		\subfigure[]{
			\includegraphics[width=0.65\columnwidth, trim=2 2 2 2,clip]{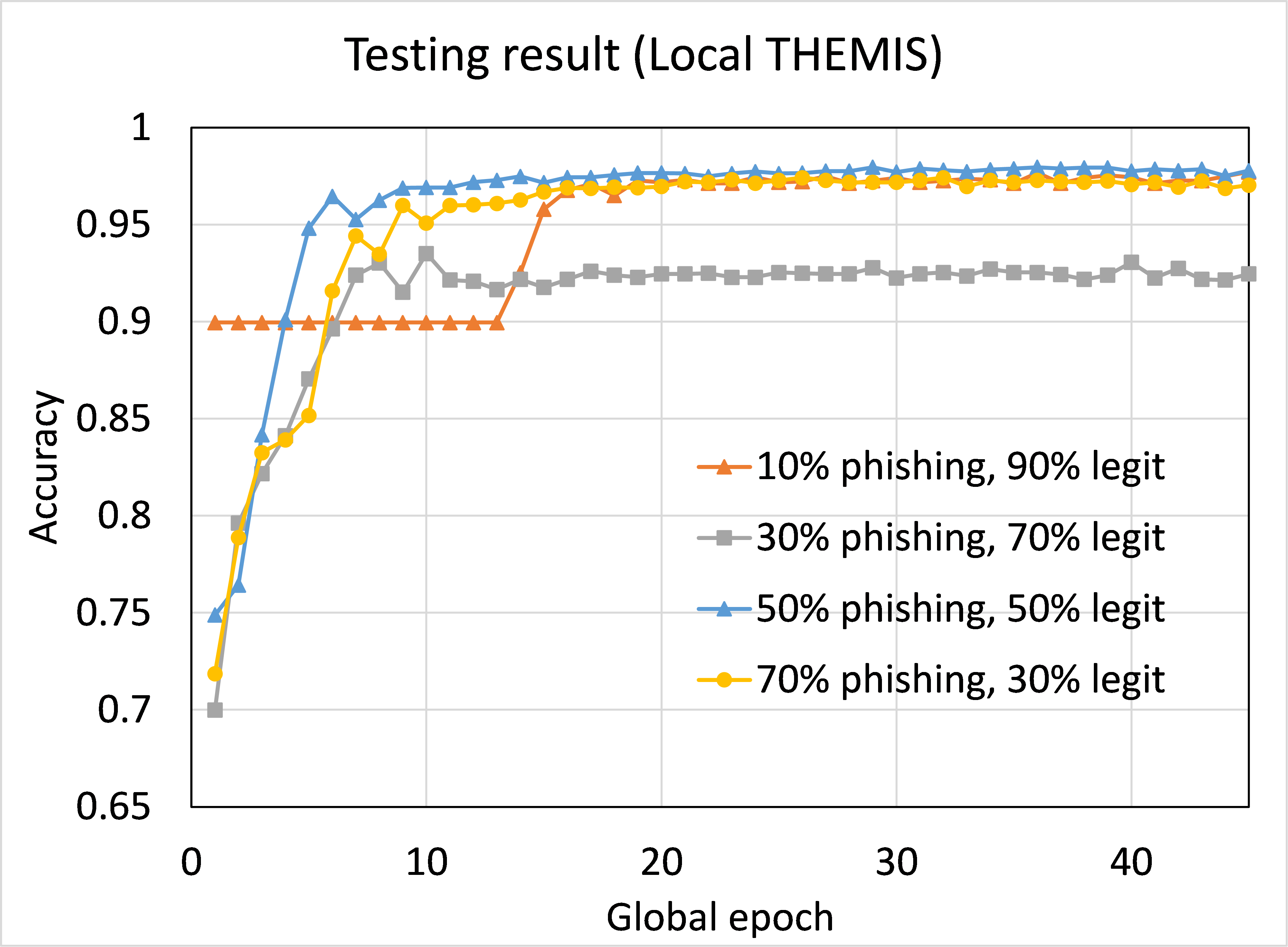}
			\label{fig:themis1c}
		}
	\subfigure[]{
			\includegraphics[width=0.65\columnwidth, trim=2 2 2 2,clip]{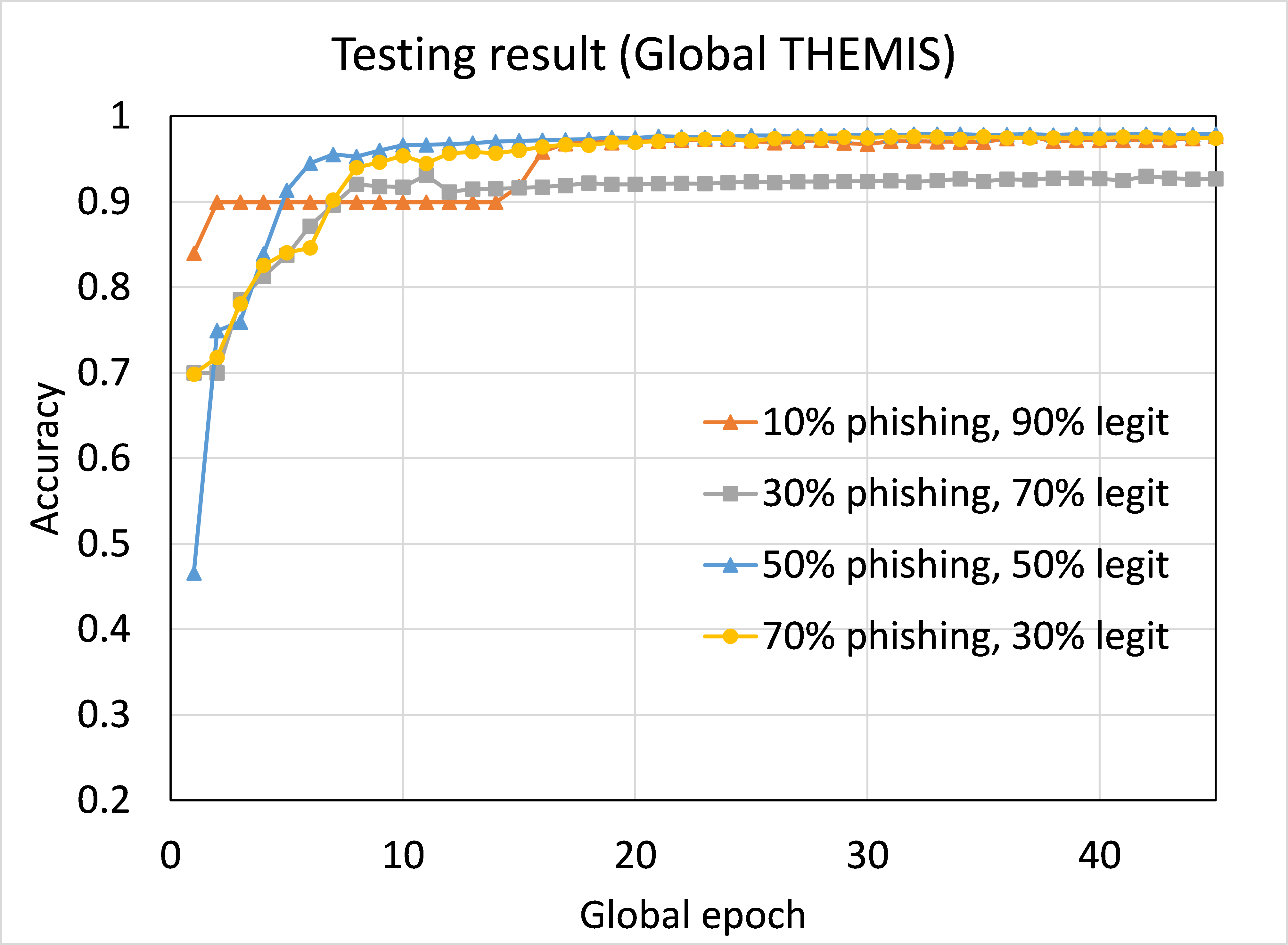}
			\label{fig:themis2c}
		}
		\subfigure[]{
		    \includegraphics[width=0.95\columnwidth, trim=2 1 1 1,clip]{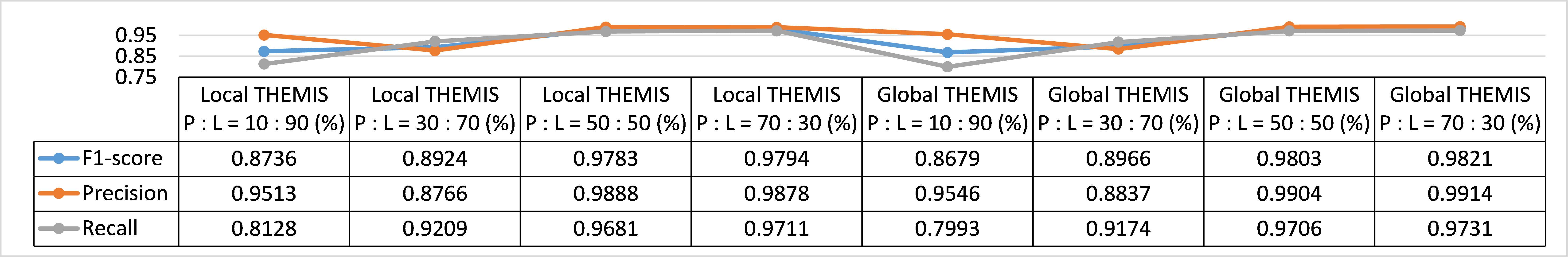}
		    \label{fig:themis3c}
		}
		\subfigure[]{
			\includegraphics[width=0.95\columnwidth, trim=2 2 2 2,clip]{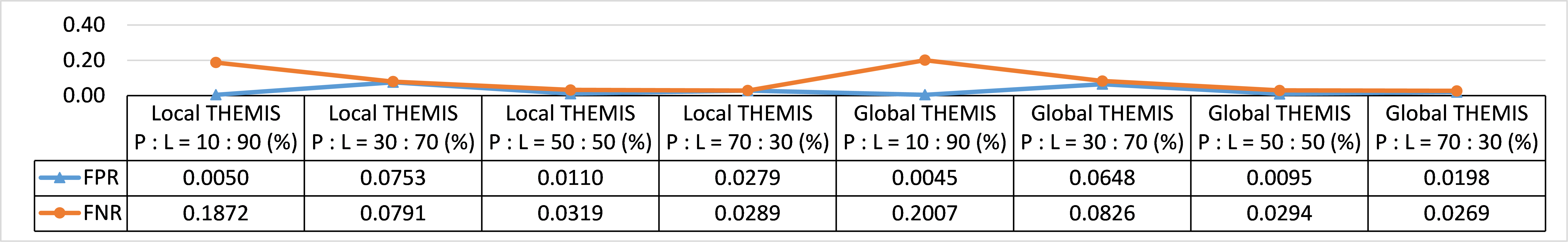}
			\label{fig:themis4c}
		}
			\caption{Showing the impact of different legit email to phishing email sample ratios in the local dataset to their convergence in FL with five clients on testing accuracy curves of (a) local model and (b) global model. Corresponding performance metrics of the testing results at the global epoch of 45 in centralized and federated learning (FL) with five clients are depicted in (c) and (d).}
	\label{fig:p_l_ratio}
\end{figure}

Fig.~\ref{fig:p_l_ratio} depicts the results for FL among 5 clients having the same size of the local dataset but all with P/L ratios of (i) 10:90 (first case), (ii) 30:70 (second case), (iii) 50:50 (third case), and (iv) 70:30 (fourth case). We choose the specific ratios for the test purpose so that the P/L ratio remains distinct. This setup is more practical than the setup with the same P/L ratio, as this has a bias in the samples. The experiments for this section have $\mathsf{var}=0$. The figure shows that until the global epoch of 15, there is a difference in the performance, where the first case (i.e., 10:90 P/L ratio) with the lower phishing email samples was not performing well compared with other cases with higher phishing email samples. However, after the 15 epoch, all cases converge similarly to provide an overall testing accuracy of around 97\% except for the second (30:70 P/L ratio) case, where the testing accuracy is around 93\%. Other performance metrics are provided in Fig.~\ref{fig:themis3c} and \ref{fig:themis4c}.  

Comparing the results for two, five, and ten clients (see Fig.~\ref{fig:p_l_ratio_appendix} in Appendix for the results with ten clients), they show a jump in the testing accuracy for the first case (10:90 P/L ratio) at different global epoch; jumps at 5, 15, and 30 global epochs for two, five and ten clients, respectively. The reasons behind these jumps are unclear.      
Overall, observing at the 45 global epoch for the setups with two, five, and ten clients, the global performance for various P/L ratios slightly decreases with the increase in phishing samples (going from 10\% to 70\%). In practice, organizations have fewer phishing emails than legitimate; thus, this concern can be contained.

\begin{mdframed}[backgroundcolor=black!10,rightline=false,leftline=false,topline=false,bottomline=false,roundcorner=2mm]
	\textbf{Summary:} Based on the empirical results in this section, in most cases, FL demonstrates highly resilient performances against the asymmetric data distribution due to the size and P/L ratio.  
\end{mdframed}

\subsection{Distributed email learning under an extreme asymmetric data distribution}
\label{sec:practical_setup}

In this section, we consider an extreme dataset diversity among the clients. We keep our different email sources as different clients; Client~1 has IWSPA dataset, Client~2 has Enron dataset, Client~3 has Nazario dataset,  Client~4 has CSIRO emails, and Client~5 has Phishbowl emails (refer to Table~\ref{data_info} for the size of local datasets). This setup captures both the variations in the sample sizes and class type across clients.

\begin{researchq}
 \textbf{How would FL perform under extreme dataset diversity among clients?}
\end{researchq}

\begin{figure}[htb]
	\centering
		\subfigure[]{
			\includegraphics[width=0.7\columnwidth, trim=2 2 2 2,clip]{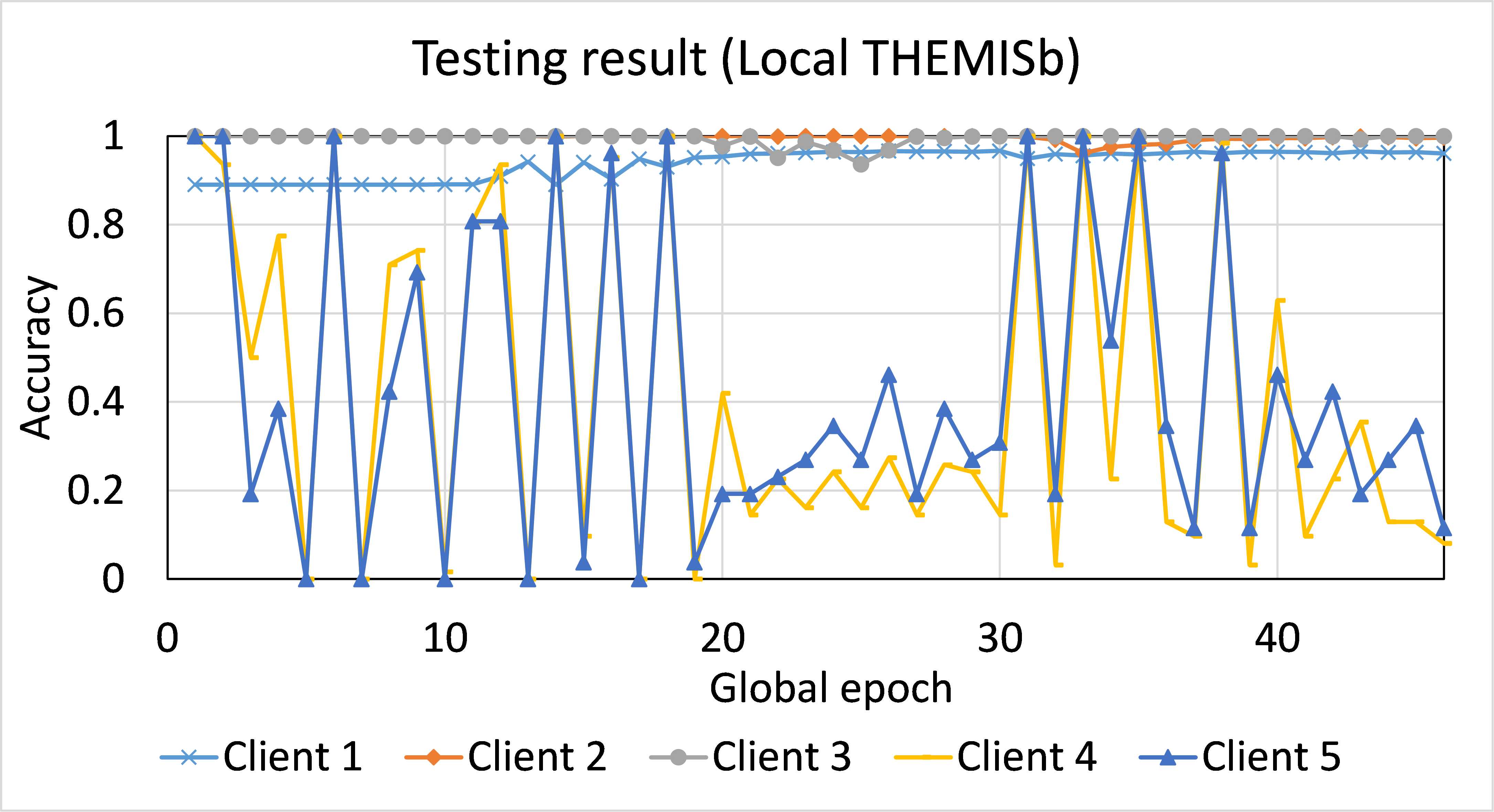}
		}
	    \subfigure[]{
			\includegraphics[width=0.7\columnwidth, trim=2 2 2 2,clip]{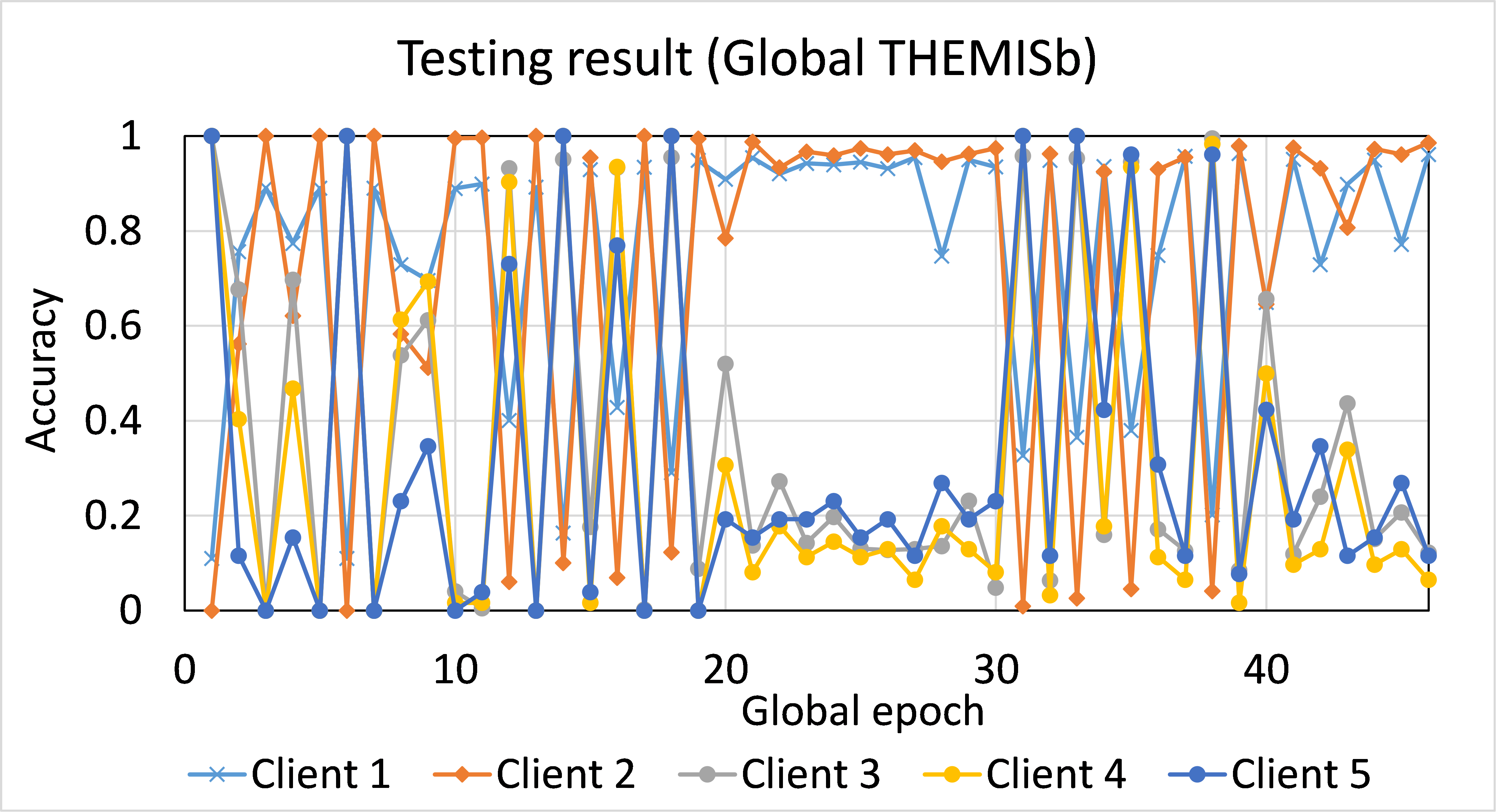}
		}
		\subfigure[]{
			\includegraphics[width=0.7\columnwidth, trim=2 2 2 2,clip]{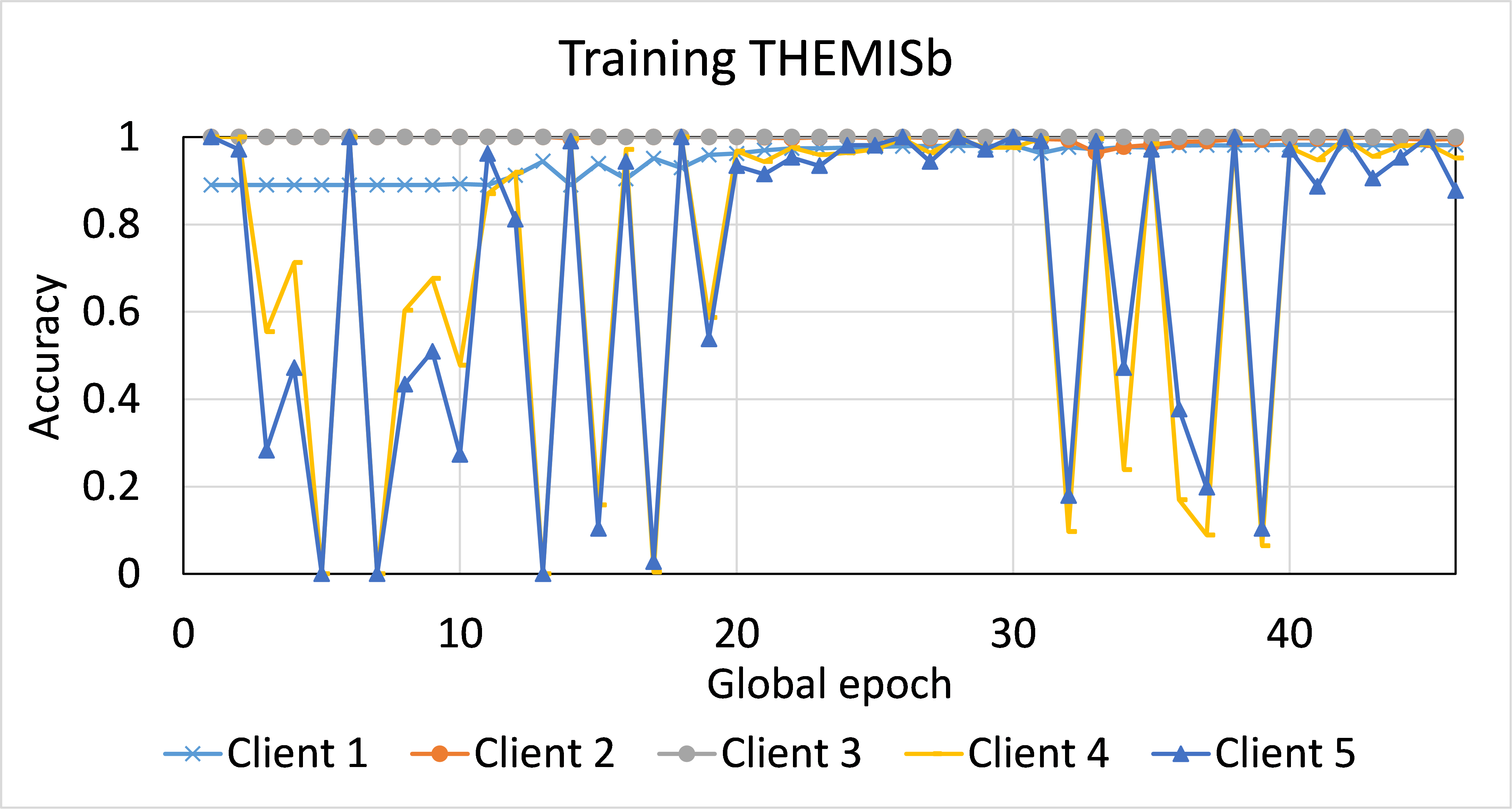}
			\label{figb:practical1}
		}
		\caption{THEMISb model with five clients: Client-level convergence curves of (a) testing accuracy of the local model, (b) testing accuracy of the global model, and (c) training accuracy.}
		\label{fig:practical_themis}
\end{figure}

\begin{figure}[htb]
	\centering
		\subfigure[]{
			\includegraphics[width=0.7\columnwidth, trim=2 2 2 2,clip]{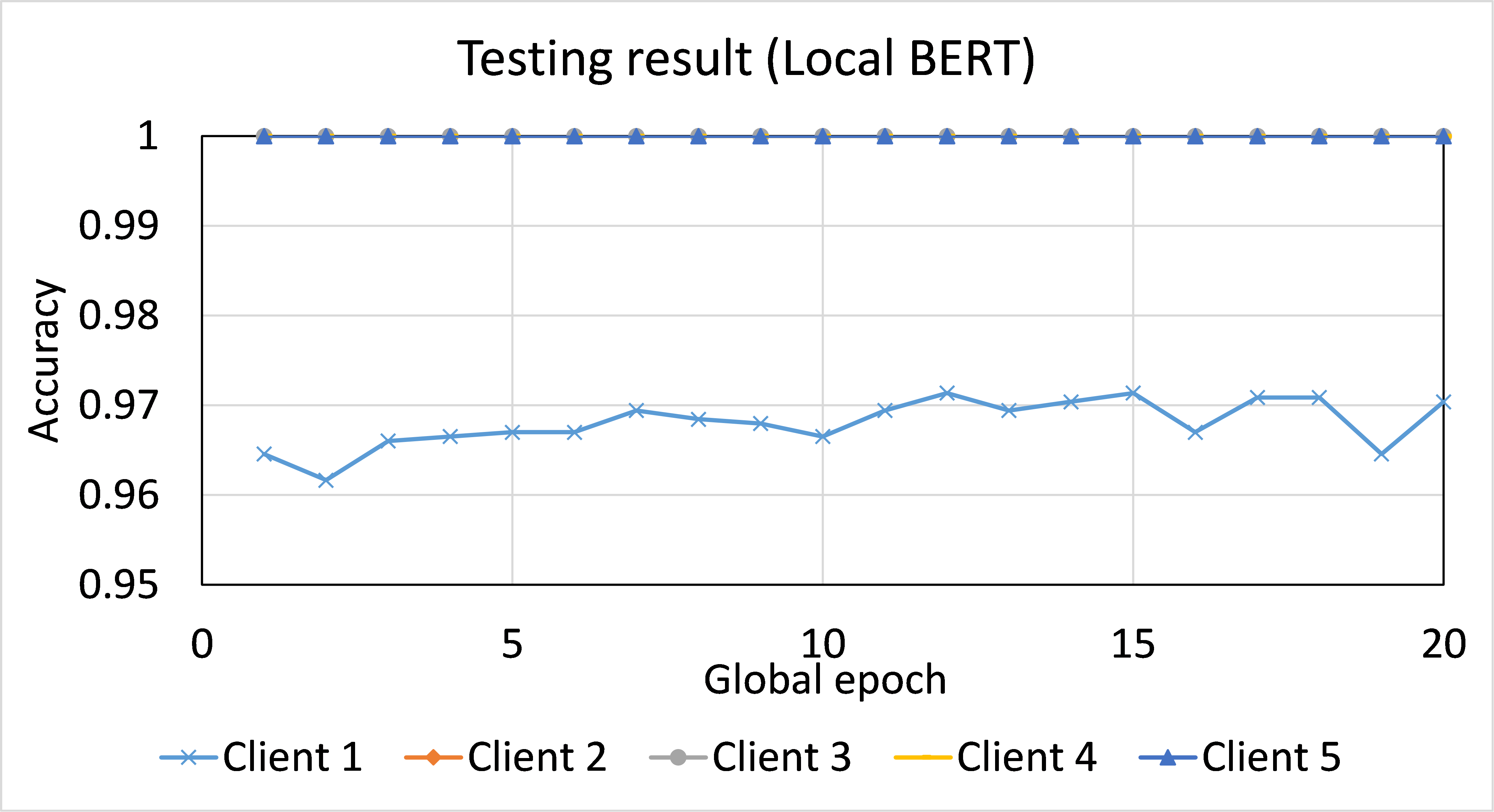}
		}
	    \subfigure[]{
			\includegraphics[width=0.7\columnwidth, trim=2 2 2 2,clip]{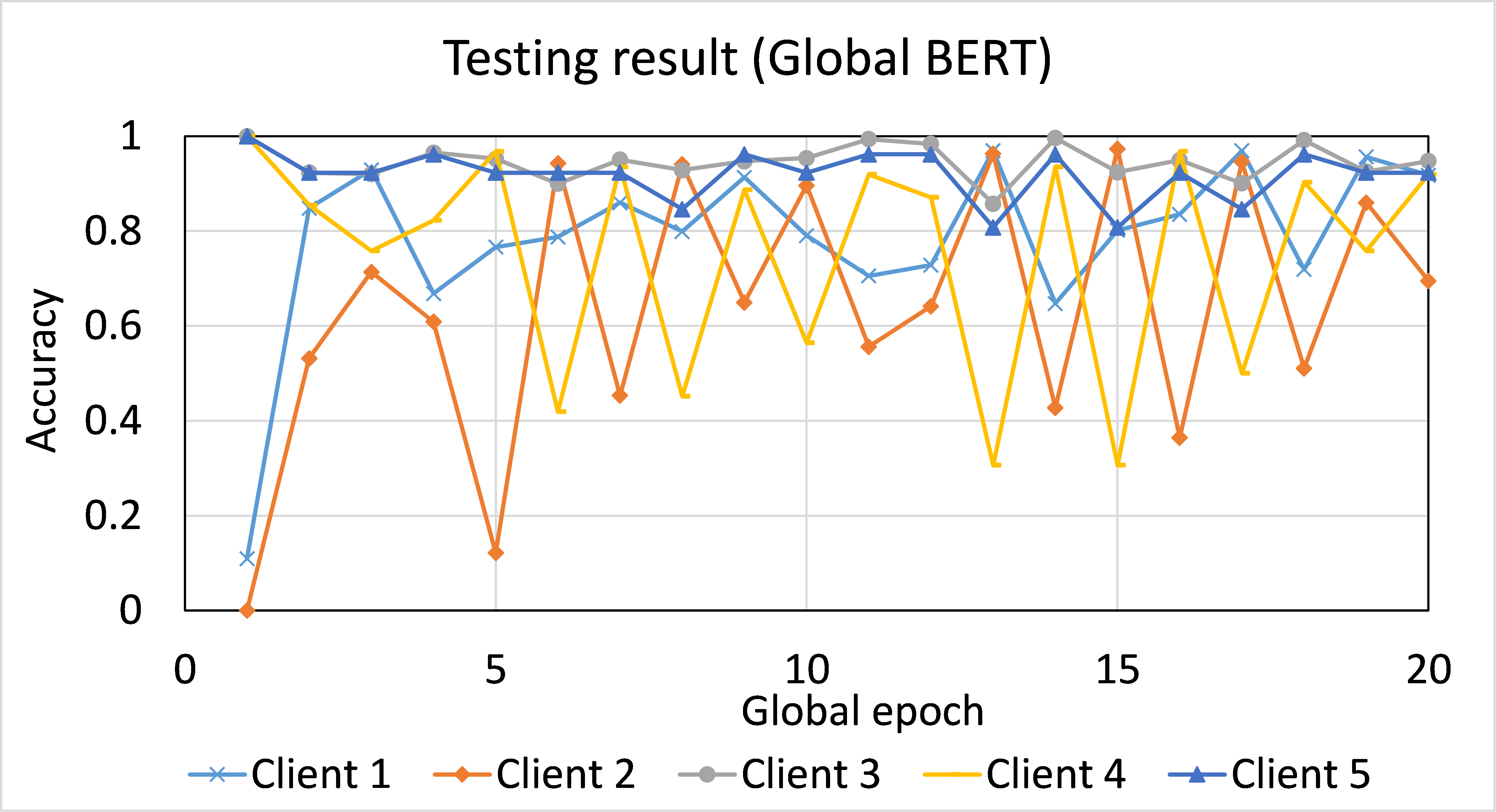}
		}
		\subfigure[]{
			\includegraphics[width=0.7\columnwidth, trim=2 2 2 2,clip]{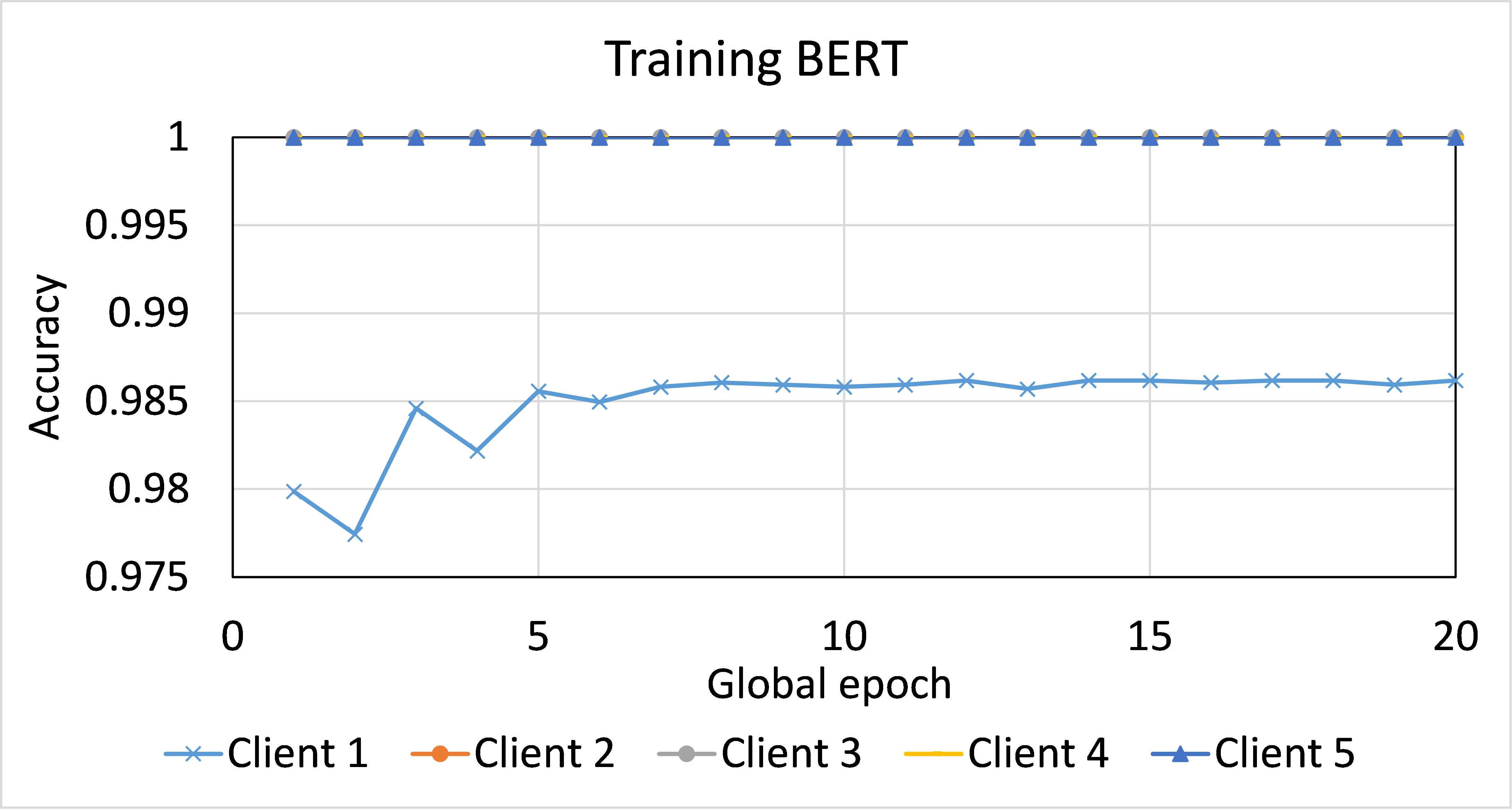}
			\label{figb:practical2}
		}
		\caption{BERT model with five clients: Client-level convergence curves of (a) testing accuracy of the local model, (b) testing accuracy of the global model, and (c) training accuracy.}
		\label{fig:practical_bert}
\end{figure}

The results for THEMISb and BERT, both considering only the body of emails while learning, are depicted in Fig.~\ref{fig:practical_themis}, and ~\ref{fig:practical_bert}, respectively. 
For the THEMIS and THEMISb model, clients having few samples and the latest phishing samples, such as client~4 and client~5, could not collaborate effectively while training and testing and suffer from high fluctuations in its results. Moreover, client~3, which has only phishing emails but in large numbers, also shows similar performance to client~4 and 5 in the global model testing. However, its local model testing result is excellent (around 99.99\% accuracy). Both the THEMIS models (THEMIS and THEMISb) show similar results. Refer to Fig.~\ref{fig:practical_themis_appendix} in the Appendix for the results of the THEMIS model (considering the email headers).  

Unlike the results of THEMISb and THEMIS, all clients converge well during training with the BERT model and perform well in the local model testing. Moreover, client~1 is under-performing among all with 97\% accuracy in the local model testing; others maintain accuracy of 99.99\%.
However, the global testing results show high fluctuations, specifically for clients 1, 2, and 4. Clients 4 and 5 have relatively stable results in global model testing than other clients. This result is in sharp contrast to that with the THEMIS/THEMISb model.
The problem with the THEMIS/THEMISb model in FL can be contained by allowing the clients to train their local models for a longer time and keep either the local or global model based on their performance for the deployment.

\begin{mdframed}[backgroundcolor=black!10,rightline=false,leftline=false,topline=false,bottomline=false,roundcorner=2mm]
	\textbf{Summary: Under extreme dataset diversity among clients, models suffer high fluctuation in their performances. However, BERT produces relatively stable results even for clients with few samples. This indicates the FL performance is model-dependent.} 
\end{mdframed}


\section{Related Works}
\subsection{Centralized learning in phishing detection}

A centralized email analysis based on AI-based methods for phishing detection has been explored for a long time. Conventional ML-based techniques such as decision trees, logistic regression, random forests, AdaBoost, and support vector machines are analyzed in phishing detection~\cite{Abu-NimehNWN07,BergholzBGMPS10,VermaSH12,vazhayil2018ped,GutierrezKCAGCB18,adaboostML,phishingsurvey}. These techniques are based on feature engineering, which requires in-depth domain knowledge and trials.
On the other hand, DL-based methods include deep neural networks \cite{SmadiAZ18}, convolutional neural networks (CNNs) \cite{CNNemail}, deep belief networks \cite{zhang2017phishing}, bidirectional LSTM with supervised attention~\cite{sys8}, and recurrent convolutional neural networks \cite{fang2019phishing}. These works are mostly based on natural language processing techniques for phishing detection.
While most existing works have focused on the effective detection of general phishing emails, few works consider specialized phishing attacks, including spear phishing attacks \cite{GasconUSR18} and business email compromise attacks \cite{cidon2019high} in specific contexts.
Despite the usefulness, all the above works operate under a setting where emails must be centralized for analysis and thus do not provide privacy protection of email datasets.

\subsection{Cryptographic Deep Learning Training}

There have been attempts on cryptographic approaches for supporting DL model training over encrypted data, which can be applicable for phishing email detection while preserving privacy.
For privacy-preserving neural network training, the first system design is SecureML~\cite{MohasselZ17}. 
%
%
In this system, multiple data providers can secretly share their data among two cloud servers, which will then conduct the training procedure over the secret-shared data.
This work relies on the secure computation techniques, e.g., secret sharing and garbled circuits, to design a secure two-party computation protocol, allowing two cloud servers to compute in the ciphertext domain the linear operations (addition and multiplication) as well as the non-linear activation functions.
Later, a design that works in the three-server model was proposed~\cite{WaghGC19}. It is based on the lightweight secret sharing technique with better performance than SecureML. 
This work assumes an adversary model where none of the three cloud servers will deviate from the protocol.
The work in \cite{MohasselR18} also operates under a similar three-server setting yet achieves more robust security against malicious adversaries who deviate arbitrarily.
This line of work presents valuable research endeavors in enabling {deep neural network} training over encrypted data. Yet, it has to rely on additional architectural assumptions (i.e., non-colluding cloud servers) and incur substantial performance overheads (up to orders of magnitude slower) compared to the plain text baseline.


%

\subsection{Federated Learning}

FL is attractive, especially when the data is sensitive, like in the financial sector (banks) and the medical sector (hospitals)~\cite{rieke2020future}. There have been several works in FL though none of them specifically address phishing email detection. Some works include the following: Google has used FL for next-word prediction in
a virtual keyboard for smartphones words~\cite{hard2018federated}, Leroy {\it et al.} applied FL for speech keyword spotting ~\cite{leroy2019federated}, Gao {\it et al.}~\cite{gao2019hhhfl} propose to use FL to train a joint model over heterogeneous ECG medical data to preserve the data privacy of each party, and 
Yang {\it et al.}~\cite{yang2019ffd} applied FL to detect credit card fraud. 


\section{Discussion and future work}
\label{sec:discussion}
This paper is the first step in federated email learning for phishing detection. Though our results demonstrated its benefits and performances, FL needs more studies, specifically for its robustness under aspects such as security attacks.

\subsection{Improving federated learning performance in phishing email detection}
This paper considered the federated averaging (FedAvg) algorithm for the model aggregation in FL.
But in literature, it is reported that FedAvg can have a hugely detrimental effect on the model's performance because there can be many variants of the model weights that only differ in the ordering of parameters, such as in neural networks~\cite{movingavg}. Besides, the personalization of the model may cause the deterioration of the model performance due to the FedAvg ~\cite{jiang2019improving}. Personalization means that the local model may fit well for some but not all devices. This means some devices will have outstanding performance and otherwise for other devices (the case for clients 4 and 5 with THEMIS model under an extreme asymmetric data distribution). With more users, the effect of personalization is more significant and resulting a low-performance global model. This effect can be mitigated using model agnostic meta learning ~\cite{jiang2019improving} to improve the global model so that it can fit well for the majority of the users and has a faster convergence. As a side effect, the local test accuracy, due to the less personalized effect, it is less likely for some devices to have excellent local model performance. Thus to improve the results further, studying other aggregating methods such as FedCurv and FedProx~\cite{fedcurv} may be required in phishing detection. 
Besides, studies with more models, a large corpus of email data, and their comparative analysis in FL are left as the next work.

\subsection{Federated learning, privacy, and security attacks}
The main challenge to deploy FL in phishing detection would be the possibility of security and privacy attacks in FL, which is observed under different datasets other than phishing emails. Moreover, privacy could be leaked due to the inference attack~\cite{nasr2019comprehensive}, while security attacks can occur due to backdoor attacks via data poisoning or parameter tampering~\cite{bagdasaryan2018backdoor}. Corresponding countermeasures, such as input filters, unlearning, strong intentional perturbation-based Trojan attack detection, and dynamic client allocation mechanism~\cite{wang2019neural,gao2019strip,kairouz2019advances,zhao2020shielding}, can be deployed to mitigate such attacks. However, future works are required to analyze their implications in phishing detection.

\subsection{Federated learning and data privacy}
FL is a privacy-by-design approach, however, FL alone can not guarantee data privacy. So, there are various other techniques such as homomorphic encryption~\cite{gentry} (a cryptographic approach) and differential privacy~\cite{diffprivacybook} used {\it along with} FL for guaranteed and provable privacy, respectively. However, homomorphic encryption increases computational overhead, and differential privacy degrades the performance as a trade-off. The integration of these techniques to FL in phishing detection remains as other research avenues.

\section{Conclusion}

This work took the first step to implement federated learning (FL) for email phishing detection in collaborative distributed frameworks. FL enables multiple organizations to collaborate to train a deep anti-phishing model without sharing their email data. Deep anti-phishing models usually have high detection but require huge data. Thus more organizations can contribute to improving the model performance, including the client-level performance, by harnessing data integration in FL, as illustrated in this paper.
Built upon the best deep learning model relevant to email phishing detection, namely BERT and THEMIS/THEMISb, our analysis under FL demonstrated promising results while preserving email privacy. More specifically, the deep learning model's performance under FL was comparable to centralized learning, and for most of the cases, it performed well under various scenarios, including asymmetric data distribution among clients. However, for an extreme asymmetry in data distribution, the performance is subjective to the model and dataset, where BERT was relatively stable than THEMIS/THEMISb. 

\section*{Acknowledgements}
The work is partially supported by the Cyber Security Cooperative Research Centre, Australia. 
The authors acknowledge Professor Rakesh Verma from the University of Houston, USA, for the IWSPA-AP corpus.

\ifCLASSOPTIONcaptionsoff
  \newpage
\fi

\bibliographystyle{IEEEtran}  
\bibliography{references}

\newpage
\appendices
\section{Supplemental results}
\subsection{Performance of THEMISb (considering email's body only) in the centralized and federated learning with two, five, and ten clients.}
\label{appendix:1}


\subsection{Client-level performance of the THEMIS model}
\label{appendix:3}

\begin{figure}[H]
	\centering
		\subfigure[]{
			\includegraphics[width=0.7\columnwidth, trim=2 2 2 2,clip]{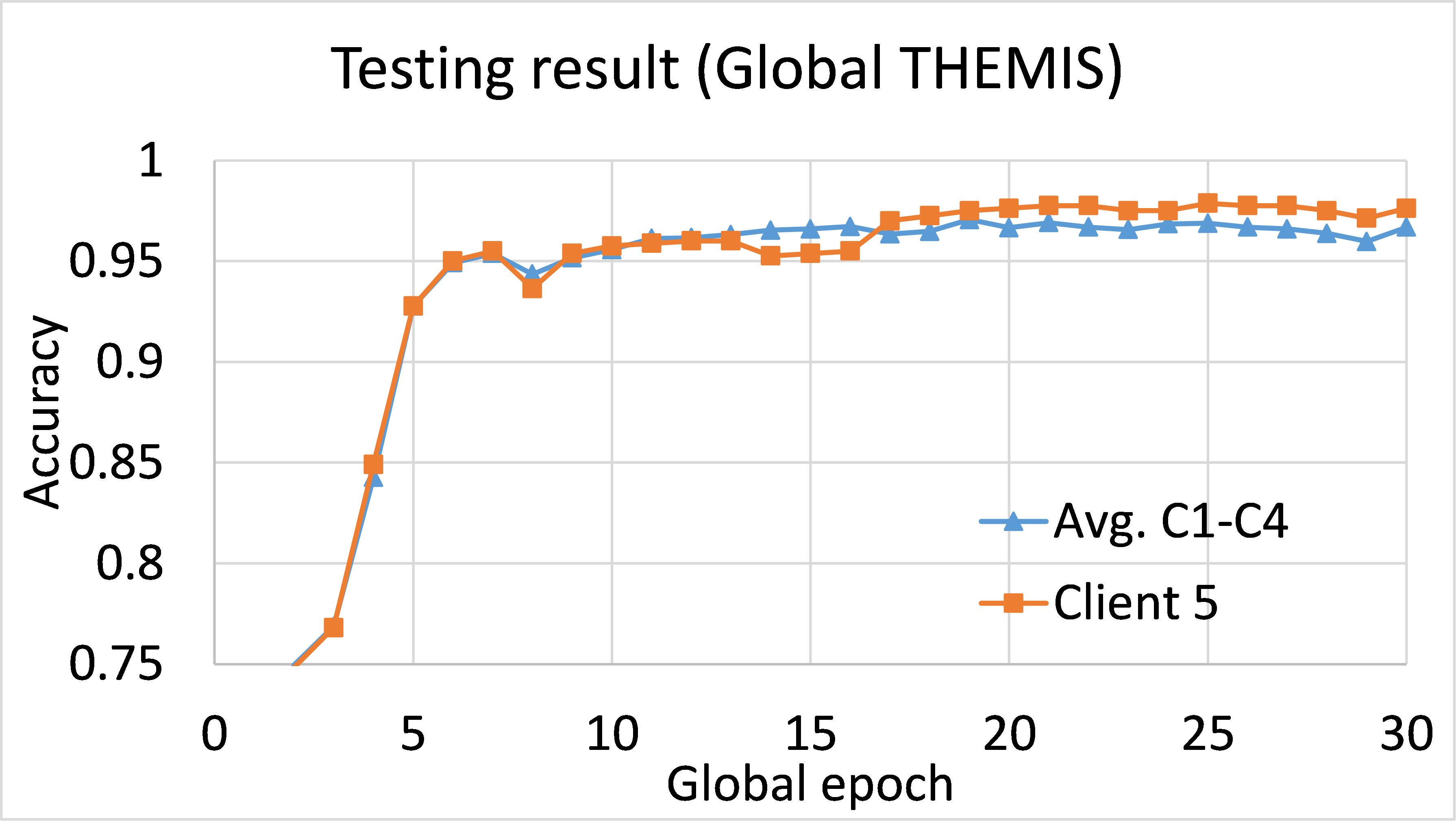}
		}
	
		\subfigure[]{
			\includegraphics[width=0.7\columnwidth, trim=2 2 2 2,clip]{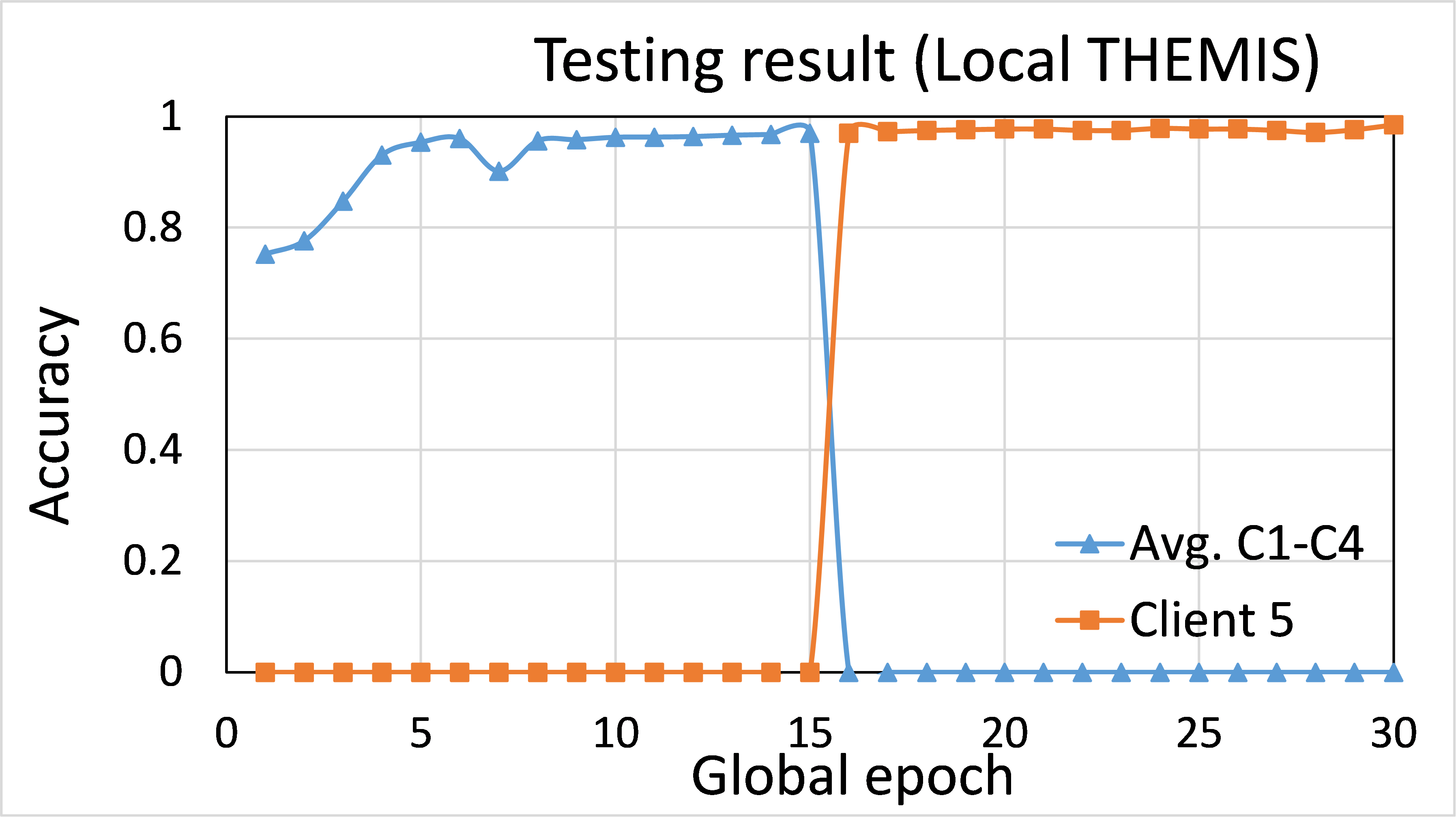}
		}
		\caption{Experiment 1: Convergence curves of (a) global testing accuracy and (b) local testing accuracy from the Experiment 1 with five clients and $\mathsf{var} = 0$. The first four clients train the model until 15 global epochs, and then (only) the fifth client trains the model.}
		\label{fig:asyn_var0}
\end{figure}

\begin{figure}[H]
	\centering
		\subfigure[]{
			\includegraphics[width=0.7\columnwidth, trim=2 2 2 2,clip]{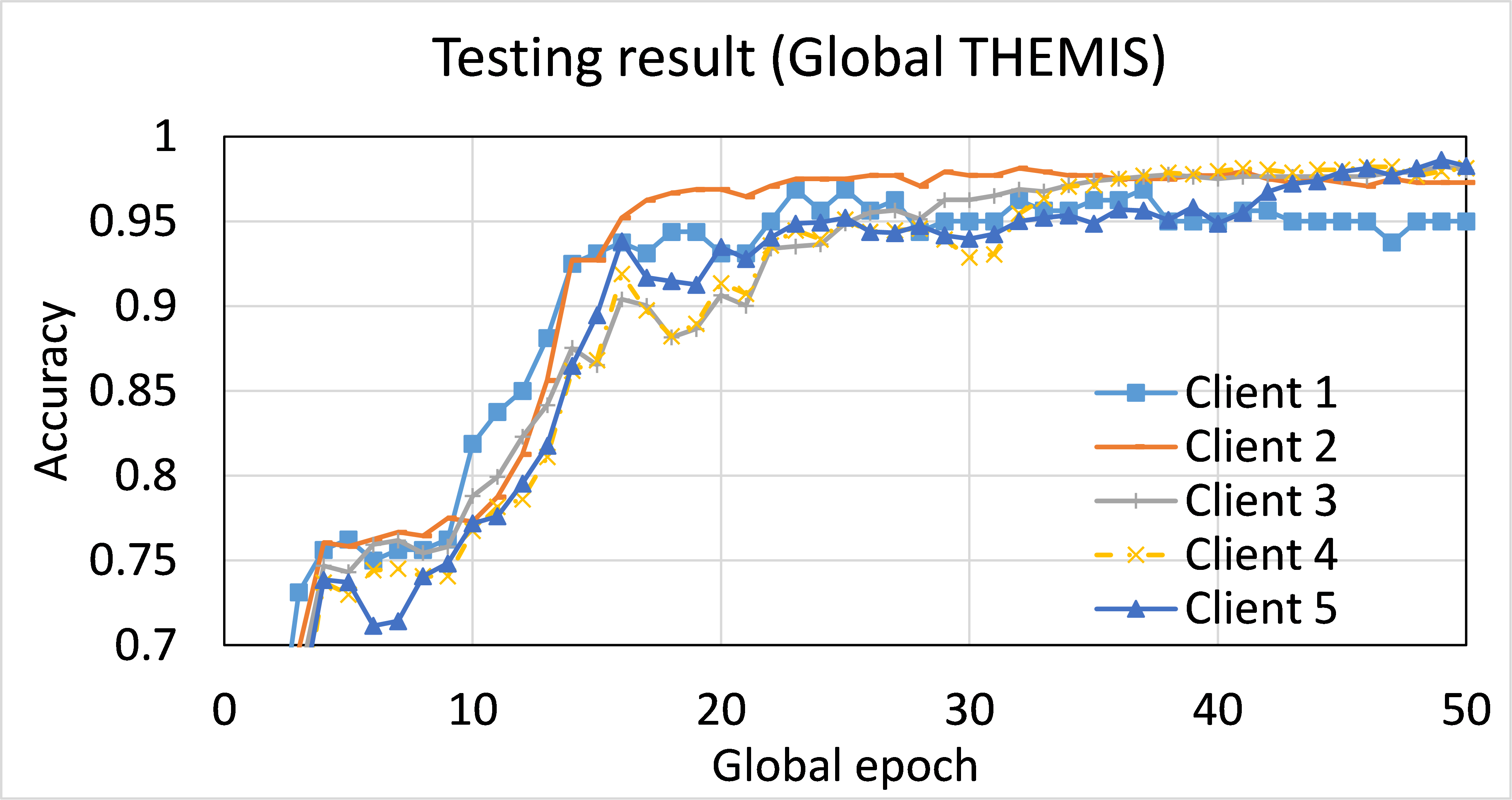}
		}
		\vskip10pt
		\subfigure[]{
			\includegraphics[width=0.7\columnwidth, trim=2 2 2 2,clip]{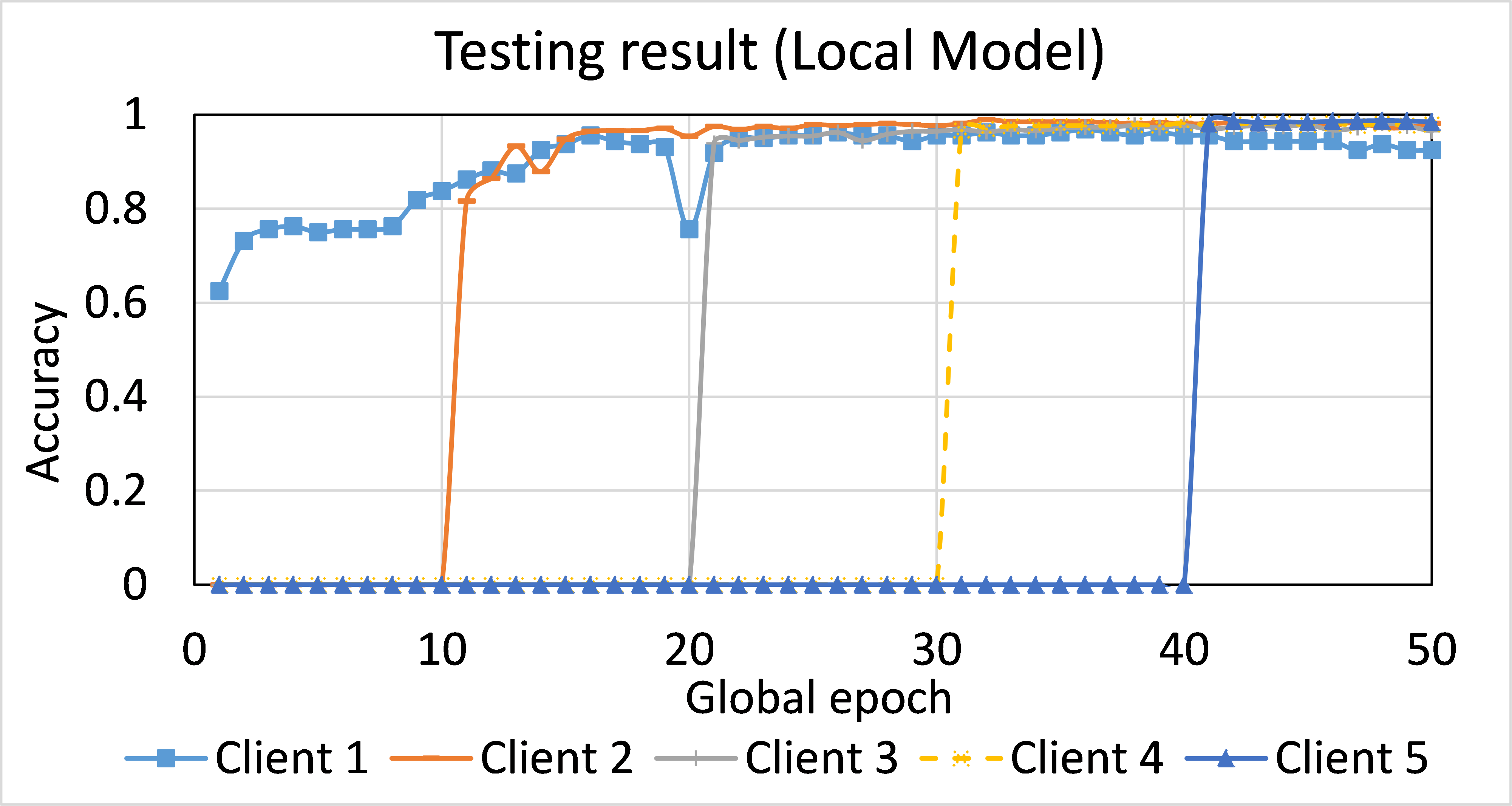}
			\label{figb:asyn2_var80}
		}
		\caption{Experiment 2: Convergence curves of (a) global testing accuracy and (b) local testing accuracy from the Experiment 2 with five clients and $\mathsf{var} = 80$. The FL training starts with one client, i.e., client~1, and a new client joins the training at every 10-th global epochs in a sequence from client~2 to client~5.}
		\label{fig:asyn2_var80}
\end{figure}


\begin{figure}[H]
	\centering
		
		\includegraphics[width=0.7\columnwidth,trim=2 2 2 2,clip]{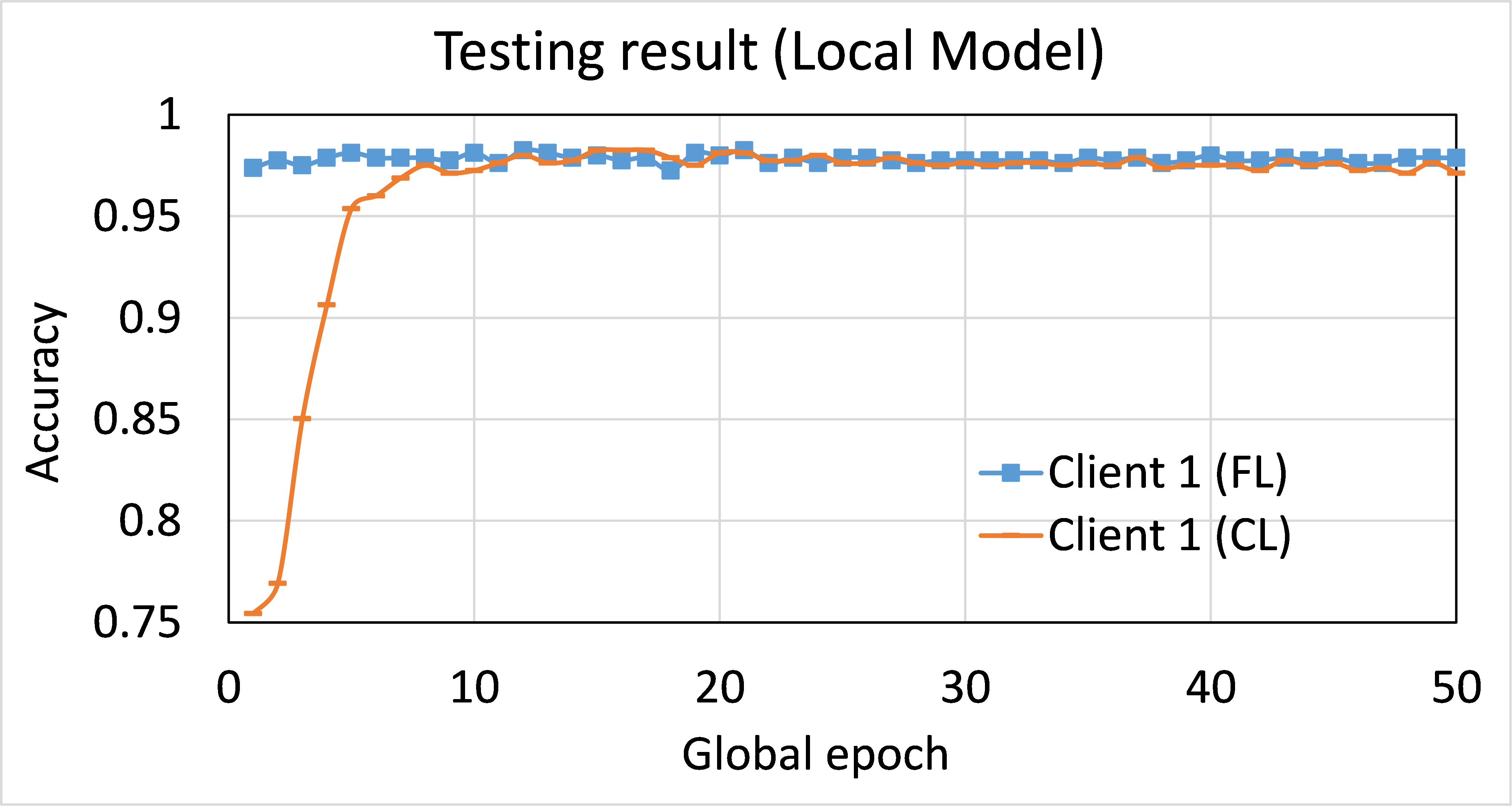}
			
		\caption{Experiment 3: Convergence curves of the (local) testing accuracy for the client~1 with and without leveraging FL. The dataset of client~1 is based on $\mathsf{var} = 0$ among five clients.}
		\label{fig:asyn2_var0_e3}
\end{figure}

\subsection{THEMIS performance over two clients with a data size variations in the dataset}
\label{appendix:2}

\subsection{THEMIS performance over ten clients with the phishing to legitimate email samples ratio variations in the dataset}
\label{appendix:4}
\begin{figure}[H]
	\centering
		\subfigure[]{
			\includegraphics[width=0.7\columnwidth, trim=2 2 2 2,clip]{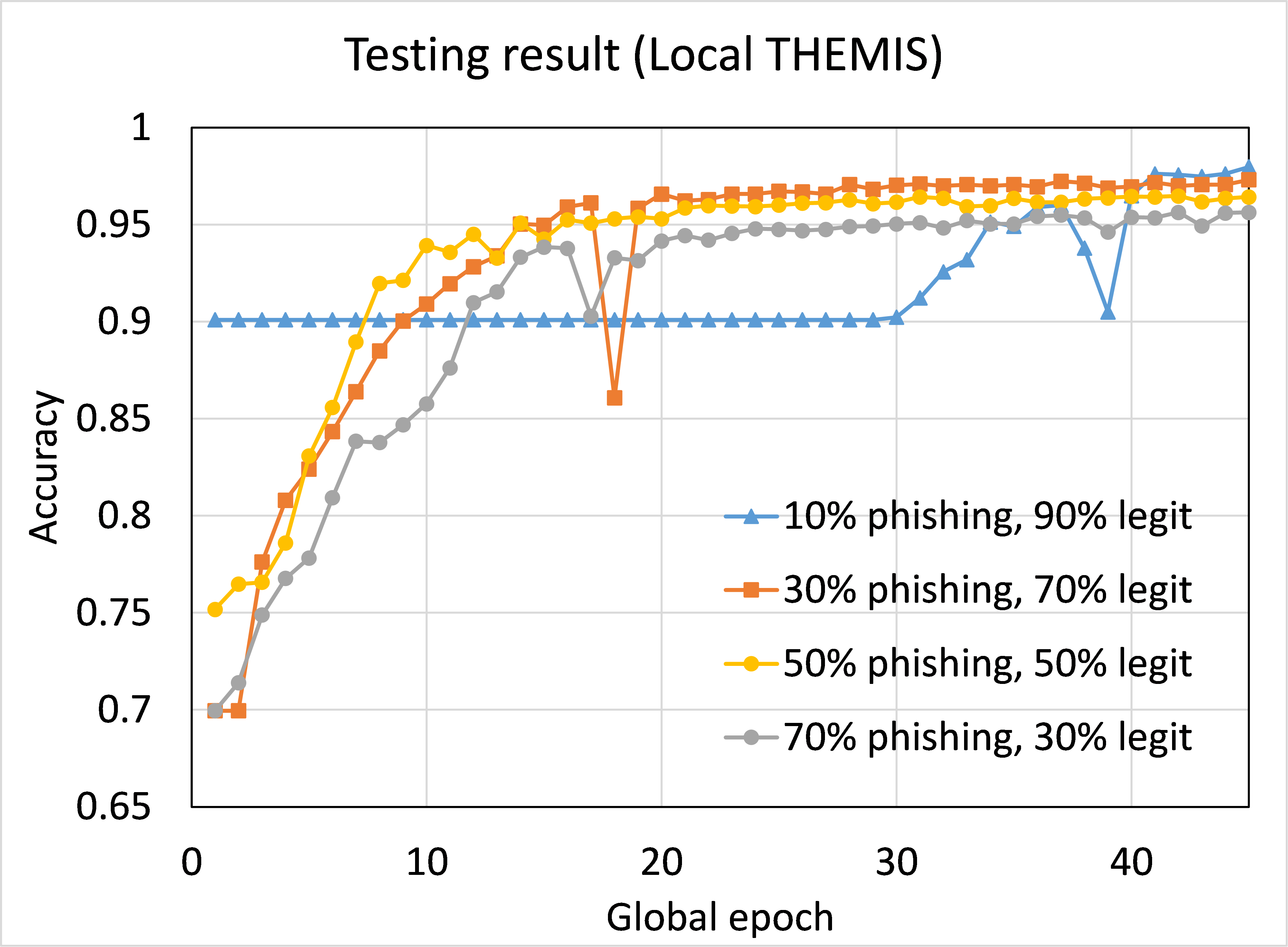}
			\label{fig:themis1e}
		}
		\vspace{1cm}
	\subfigure[]{
			\includegraphics[width=0.7\columnwidth, trim=2 2 2 2,clip]{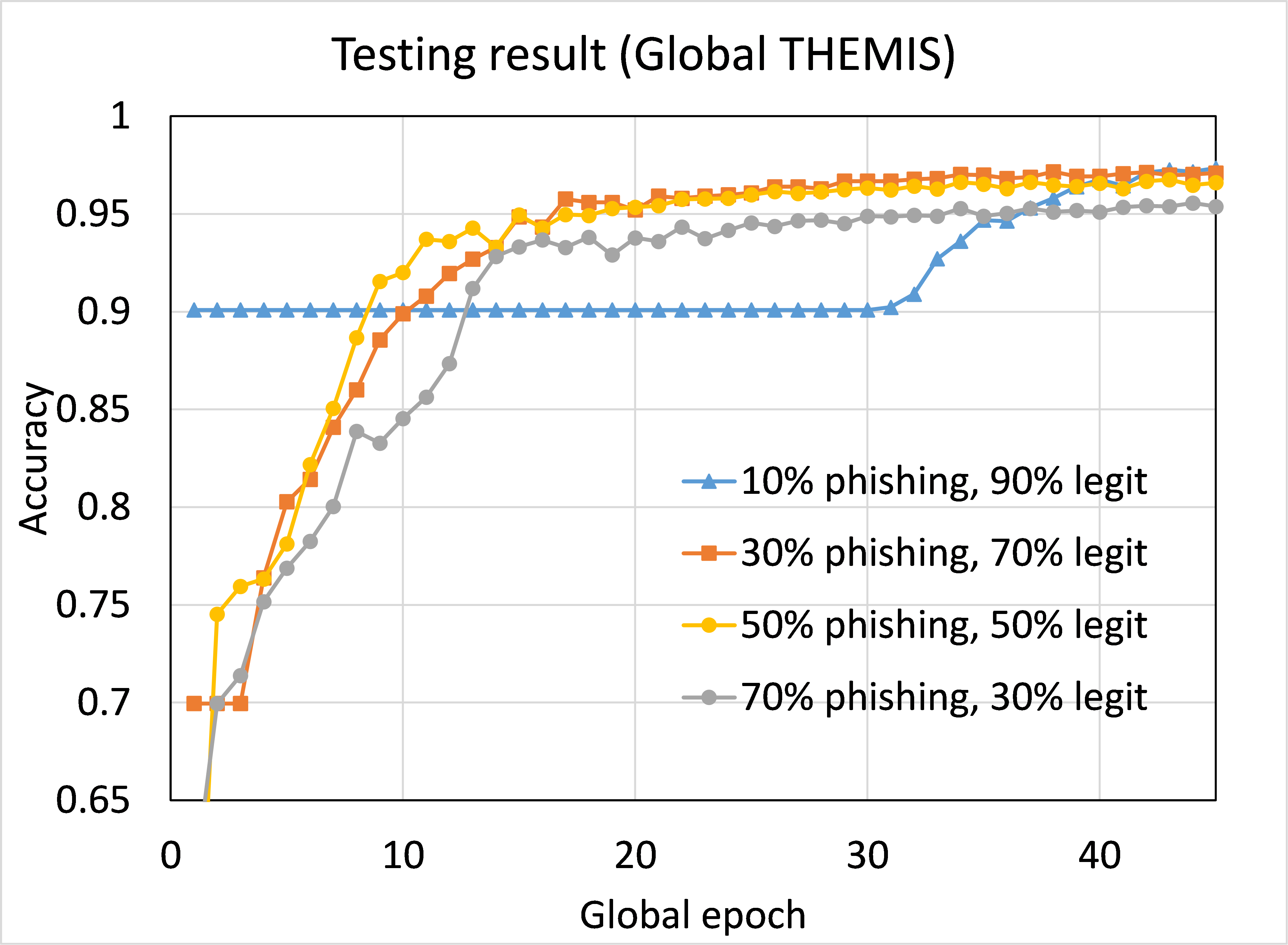}
			\label{fig:themis2e}
		}
		
		
		\subfigure[]{
		    \includegraphics[width=0.95\columnwidth, trim=2 1 1 1,clip]{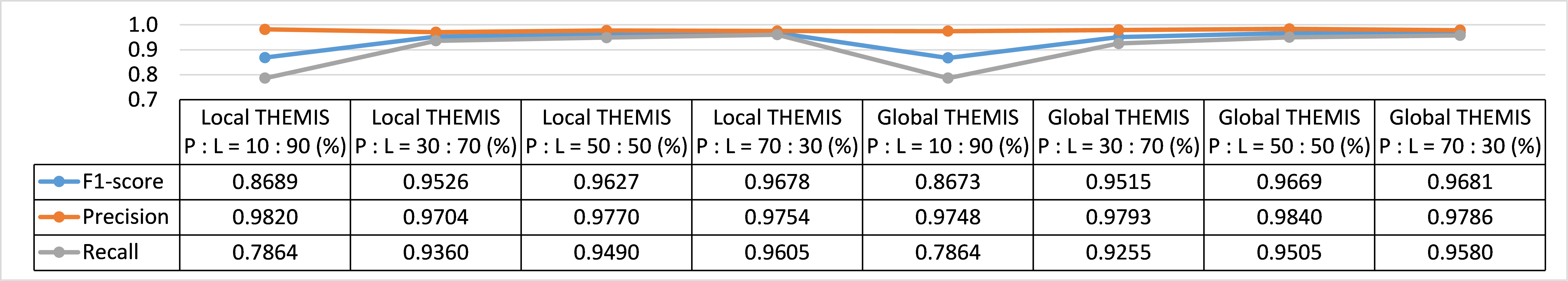}
		    \label{fig:themis3e}
		}
		
		\subfigure[]{
			\includegraphics[width=0.95\columnwidth, trim=2 2 2 2,clip]{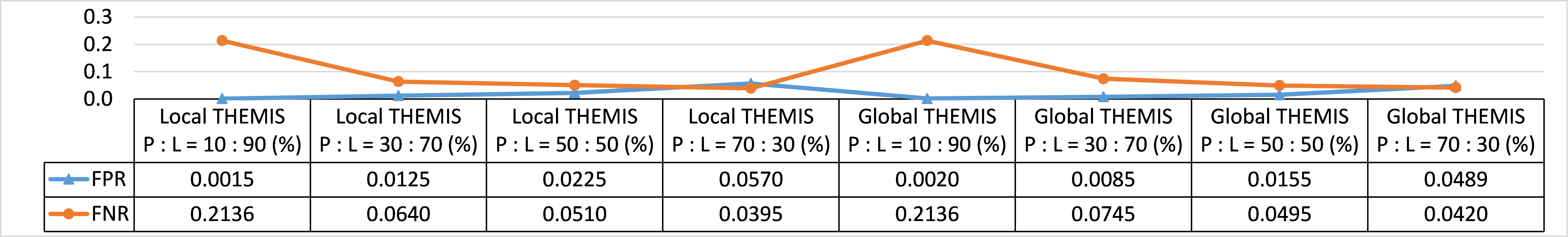}
			\label{fig:themis4e}
		}
			\caption{Showing the impact of different legit email to phishing email samples ratios in the local dataset to their convergence in FL with ten clients on testing accuracy curves of the (a) local model and (b) global model. Corresponding performance metrics of the testing results at the global epoch of 45 in centralized and federated learning (FL) with ten clients are depicted in (c) and (d).}
	\label{fig:p_l_ratio_appendix}
\end{figure}

\subsection{Distributed email learning under an extreme asymmetric data distribution with THEMIS (considering both email's header and body information)}

\begin{figure}[H]
	\centering
		\subfigure[]{
			\includegraphics[width=0.7\columnwidth, trim=2 2 2 2,clip]{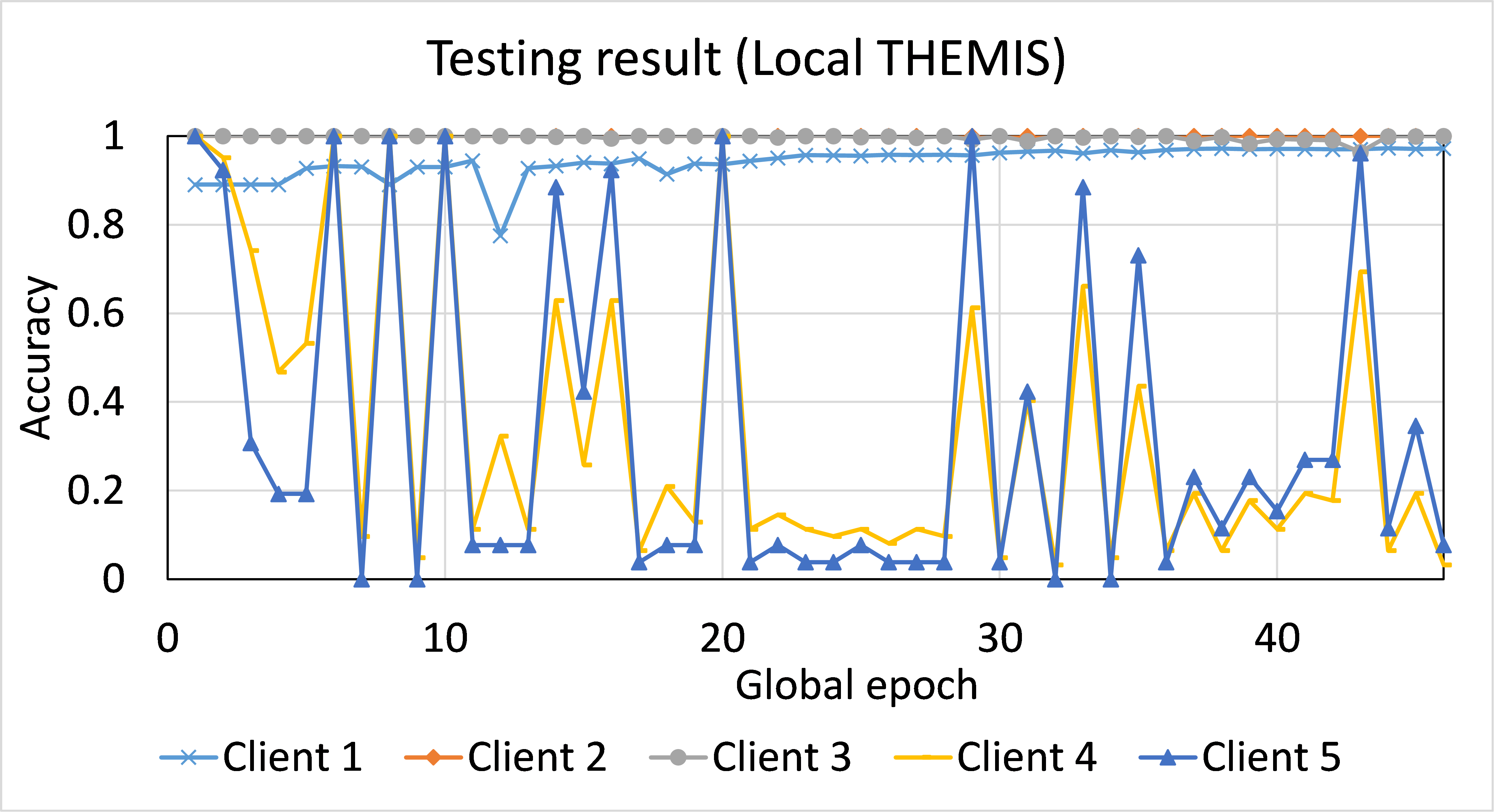}
		}
		
    \vskip5pt
	    \subfigure[]{
			\includegraphics[width=0.7\columnwidth, trim=2 2 2 2,clip]{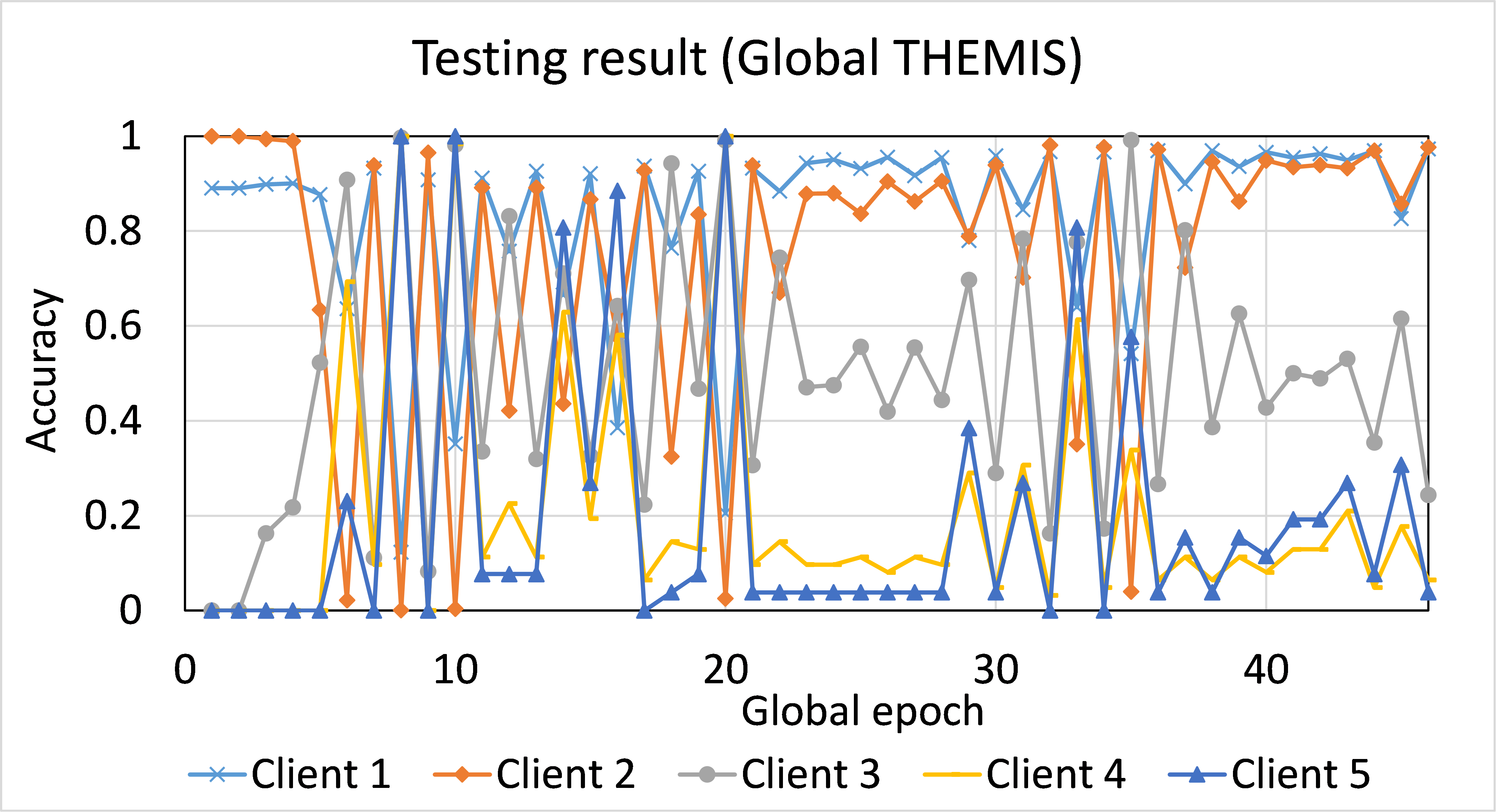}
		}
		\vskip5pt
		\subfigure[]{
			\includegraphics[width=0.7\columnwidth, trim=2 2 2 2,clip]{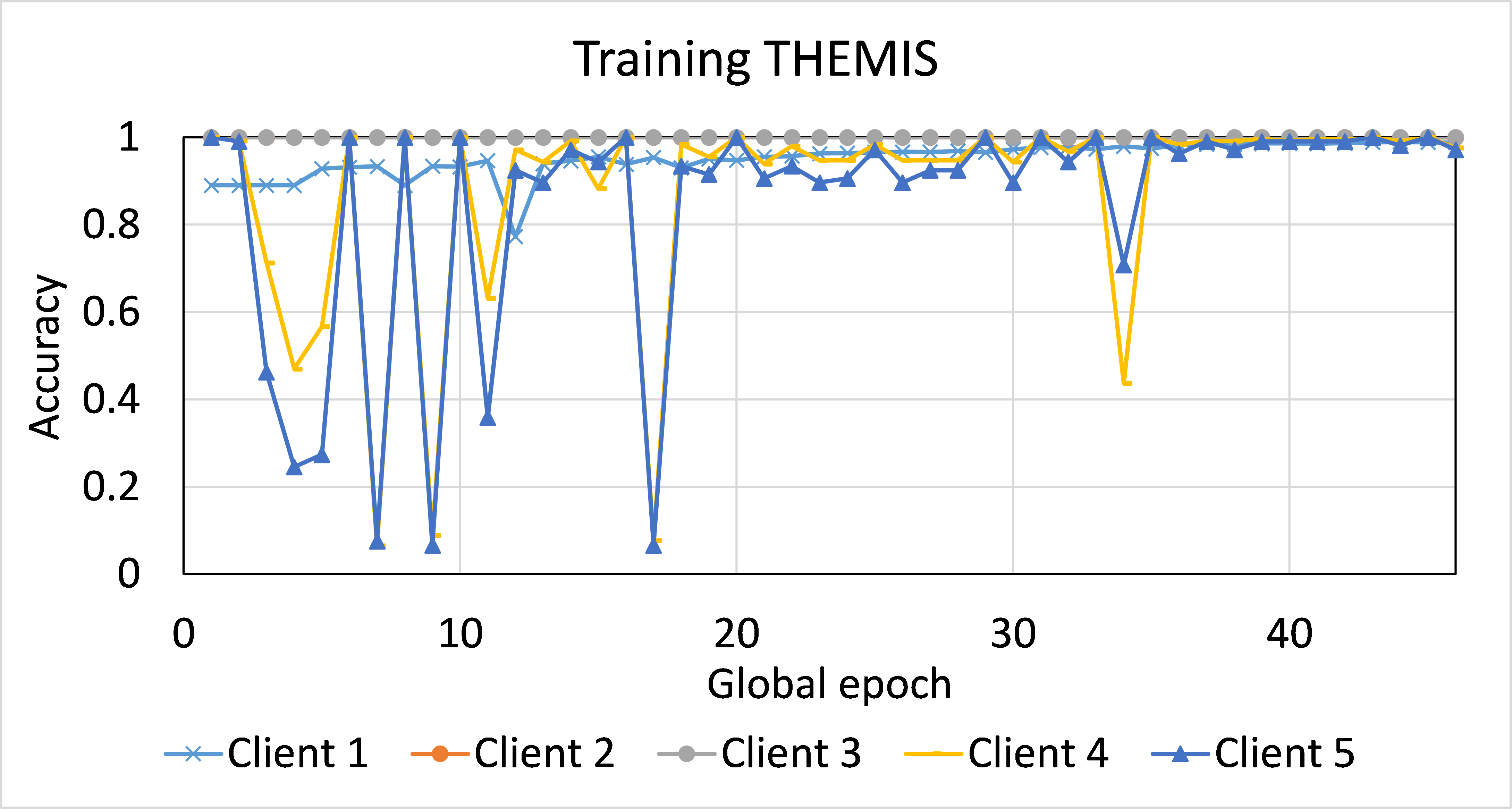}
			\label{figb:practical_themis}
		}
		\caption{THEMIS model with five clients: Client-level convergence curves of (a) testing accuracy of the local model, (b) testing accuracy of the global model, and (c) training accuracy.}
		\label{fig:practical_themis_appendix}
\end{figure}

\end{document}